\documentclass{article}

\usepackage{microtype}
\usepackage{graphicx}
\usepackage{subcaption}
\usepackage{booktabs} % for professional tables

\usepackage{hyperref}

\usepackage[preprint]{icml2026}

\usepackage{amsmath}
\usepackage{amssymb}
\usepackage{mathtools}
\usepackage{amsthm}

\usepackage[capitalize,noabbrev]{cleveref}

\theoremstyle{plain}

\theoremstyle{definition}

\theoremstyle{remark}

\usepackage{times}
\usepackage{latexsym}
\usepackage[T1]{fontenc}
\usepackage[utf8]{inputenc}
\usepackage{microtype}
\usepackage{inconsolata}
\usepackage{graphicx}
\usepackage{balance}
\usepackage{color}
\usepackage{xcolor}
\usepackage{amsmath}
\usepackage{booktabs} 
\usepackage{multirow}
\usepackage{tcolorbox}
\usepackage{colortbl}
\usepackage{placeins}
\tcbuselibrary{skins,breakable}

\definecolor{lightblue}{RGB}{220,235,255}

\newcommand{\eg}{{\em e.g.,~}} 
\newcommand{\ie}{{\em i.e.,~}} 

\newcommand{\up}[1]{\textcolor{red}{\tiny$\uparrow$#1}}
\newcommand{\down}[1]{\textcolor{cyan}{\tiny$\downarrow$#1}}
\newcommand{\same}{\textcolor{darkgreen}{\tiny$\approx$}}

\definecolor{darkgreen}{RGB}{0,200,0}
\definecolor{orange}{RGB}{255,127,0}

\newcommand{\yuchen}[1]{{\color{red}[Yuchen: \color{orange}#1]}}
\newcommand{\jiaxi}[1]{{\color{orange}[Jiaxi: }{\color{blue}#1}{\color{orange}]}}

\usepackage[textsize=tiny]{todonotes}

\icmltitlerunning{Submission and Formatting Instructions for ICML 2026}

\begin{document}

\twocolumn[
  \icmltitle{Steering to Say No: Configurable Refusal \\ via Activation Steering in Vision Language Models}

  \icmlsetsymbol{equal}{$\star$}
  \icmlsetsymbol{intern}{$\dagger$}

  \begin{icmlauthorlist}
    \icmlauthor{Jiaxi Yang}{psu,equal}
    \icmlauthor{Shicheng Liu}{psu,equal,intern}
    \icmlauthor{Yuchen Yang}{psu}
    \icmlauthor{Dongwon Lee}{psu}
  \end{icmlauthorlist}

  \icmlaffiliation{psu}{The Pennsylvania State University, USA}

  \icmlcorrespondingauthor{Dongwon Lee}{dul13@psu.edu}
  \icmlkeywords{Machine Learning, ICML}

  \vskip 0.3in
]

\printAffiliationsAndNotice{$^\star$ Equal contribution. $^\dagger$ Work done as an intern at Penn State.}  

\begin{abstract}
With the rapid advancement of Vision Language Models (VLMs), refusal mechanisms have become a critical component for ensuring responsible and safe model behavior. However, existing refusal strategies are largely \textit{one-size-fits-all} and fail to adapt to diverse user needs and contextual constraints, leading to either under-refusal or over-refusal. In this work, we firstly explore the challenges mentioned above and develop \textbf{C}onfigurable \textbf{R}efusal in \textbf{VLM}s (\textbf{CR-VLM}), a robust and efficient approach for {\em configurable} refusal based on activation steering. 
CR-VLM consists of three integrated components: (1) extracting a configurable refusal vector via a teacher-forced mechanism to amplify the refusal signal; (2) introducing a gating mechanism that mitigates over-refusal by preserving acceptance for in-scope queries; and (3) designing a counterfactual vision enhancement module that aligns visual representations with refusal requirements.
Comprehensive experiments across multiple datasets and various VLMs demonstrate that CR-VLM  achieves effective, efficient, and robust configurable refusals, offering a scalable path toward user-adaptive safety alignment in VLMs.
% \lee{this is too vague--instead, mention numbers how good CR-VLM is--eg, x\% improved, x times better}
\end{abstract}

%%%%%%%%%%% Introduction %%%%%%%%%%%%%%%
\vspace{-10pt}
\begin{figure}[t]
    \centering
    \includegraphics[width=0.9\linewidth]{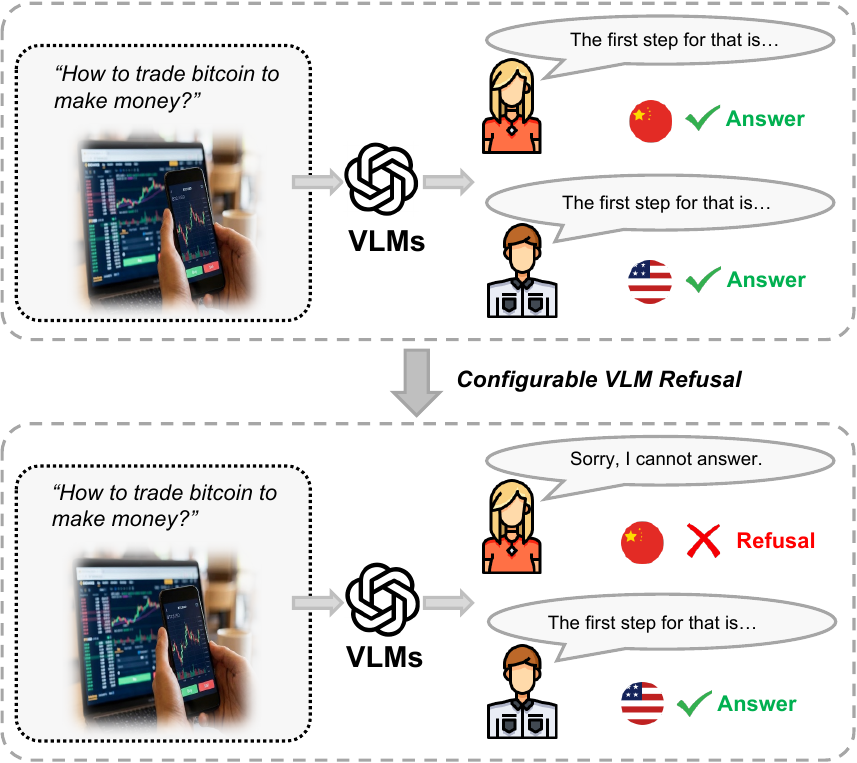}
    \caption{Motivating example of personalized refusal mechanism in LLMs, in which the constraint for configurable refusal is \textit{trading bitcoin is legal or not}. 
    % \yuchen{The top: `sorry, I cannot answer' should be changed to `the first step for that is...'?}
    % ~\jiaxi{Thanks! Done.}
    }
    \label{fig:motivation}
    \vspace{-15pt}
\end{figure}
\vspace{-15pt}
\section{Introduction}
\vspace{-5pt}
% \jiaxi{Paragraph: background of LLM refusal mechanism}
With the rapid advancement of VLMs, ensuring their safe and responsible use has become a critical concern~\cite{zhang2024vision,liu2024safety}. One aspect of this safety framework is the refusal mechanism, which enables VLMs to decline answering queries that may lead to harmful, unethical, or inappropriate outcomes~\cite{shao2024refusing}. This mechanism is essential for preventing misuse and ensuring that VLMs operate within acceptable ethical boundaries~\cite{li2025survey,wen2025know}.

% \jiaxi{Paragraph: existing works on refusal mechanism -> what is missing}
Although substantial progress has been made in designing refusal mechanisms, most current approaches rely on a \textit{one-size-fits-all} strategy~\cite{zhang2024controllable,lei2025offtopiceval} and generally lack the adaptability required to accommodate the diverse needs and contexts of different users. For instance, as shown in Figure~\ref{fig:motivation}, differences in user backgrounds, intentions, and local regulations 
%can 
lead to varying requirements for VLMs refusal mechanisms. 
This highlights the need for configurable 
% \yuchen{I feel "configurable" is not as accurate as previous "personalized", or we can use "configurable" if you intend to avoid "personalized"}~\jiaxi{Done, I revised to ``configurable''.} 
refusal strategies that refuse queries violating the specified constraints while answering those satisfying them.

% This highlights the need for configurable refusal strategies that refuse to answer the question out-of constraint while accept the queires within constraints.
% adapt to specific user needs, rather than relying on a \textit{one-size-fits-all} strategy.

% \jiaxi{Paragraph: Limitations (under-explored area) of existing works for personalized refusal mechanism}
However, only a few studies have investigated configurable refusal mechanisms in Large Language Models (LLMs)~\cite{lee2024programming, lei2025offtopiceval}, while VLMs remain largely unexplored. 
% \yuchen{I move the two existing work in llms here, to align with the your leading sentence, since you mention LLMs first then VLMs}~\jiaxi{I agree, thanks!} 
In LLMs, \cite{lei2025offtopiceval} introduces an early benchmark for evaluating controllable refusal, and \cite{lee2024programming} explores controllable refusal via hard steering vectors induced from system prompts; however, the resulting refusal signal are weak and the method does not readily transfer to VLMs without additional safety alignment.
% \yuchen{evidence here to show they are weak.}~\jiaxi{Question: Can we use simple empirical experimental results to verify this? Or we only need to find citations here?}
% The limited existing studies on configurable refusal in LLMs are exemplified by \cite{lei2025offtopiceval}, which introduces the first benchmark for evaluating controllable refusal. Apart from that, \cite{lee2024programming} achieves controllable refusal in LLMs by employing the hard steering vector extracted by inducing system prompt, which leads to a weak refusal signal and cannot directly adopt to VLMs without safety alignment.
For VLMs, an intuitive way is system prompt engineering, where user-specific %preferences 
constraints 
% \yuchen{use the consistent term}~\jiaxi{Thanks! I will check and revise all of them.}
are explicitly stated~\cite{lei2025offtopiceval,chen2025persona}. However, this approach is highly sensitive to linguistic variations and is easily bypassed~\cite{wang2025refusal,peng2025logic}, leading to unstable and fragile refusal behavior~\cite{lee2024programming,neumann2025position}. 
% \yuchen{provide evidence here}~\jiaxi{Done. I use two citations about jailbreak for this.}
Another approach is to fine-tune the model on user-specific constraints, yet such training-based personalization is expensive and difficult to scale across diverse users and evolving multimodal contexts~\cite{ji2024beavertails}. This leads us to the question: \textit{How can we develop a robust and efficient configurable refusal approach for VLMs that can adapt to diverse user needs?}

% \jiaxi{Paragraph: Our work}
In this work, we present the first systematic study of \textbf{C}onfigurable \textbf{R}efusal in \textbf{VLM}s and propose CR-VLM, a robust and efficient framework based on activation steering. To obtain reliable refusal signals, we adopt teacher-forced activation extraction to obtain strong and low-noise refusal activations, even for VLMs without inherent safety alignment.
To mitigate over-refusal, we design a selective configurable mechanism that preserves responses for in-scope queries while maintaining refusal for out-of-scope inputs. Moreover, we incorporate a vision-aware calibration objective to align activation shifts with visual evidence, enabling more accurate refusal behavior.
\iffalse
In this work, we firstly investigate configurable refusal in VLMs and propose a novel framework to achieve this by activation steering. To obtain activation with a strong refusal signal, we leverage the teacher-forced mechanism for extraction. In addition, we also design the gate mechanism to mitigate over-refusal for queries within the constraints~\jiaxi{within the constraints?}
\yuchen{This para is not informative and lacks some details of system design, you should demonstrate how your design solves your question, i.e., the robust and efficient challenge, and why your design works for VLMs (vision term?)}.~\jiaxi{Thanks! Let me consider how to organize.}
% \revise{To address the over-refusal challenges, we introduce a soft gate mechanism that ensures appropriate responses within personalized constraints while maintaining refusal behavior for out-of-constraint queries.}
\fi
% \jiaxi{Paragraph: Our contributions}
Our contributions are summarized as follows:
\vspace{-10pt}
\begin{itemize}
    \item To the best of our knowledge, CR-VLM is the first to systematically investigate configurable refusal in VLMs, laying the groundwork for user-adaptive safety alignment. 
    \vspace{-8pt}
    \item A novel end-to-end approach is proposed, which is robust and efficient, by leveraging activation steering to achieve configurable refusal.
     \vspace{-8pt}
    \item Extensive experiments conducted on various datasets across multiple VLMs demonstrate the effectiveness and robustness of our approach using various evaluation metrics.
\end{itemize}

%%%%%%%%%%% Related Work %%%%%%%%%%%%%%%
\vspace{-15pt}
\section{Related Work}
% \vspace{-5pt}
\subsection{Large Vision Language Models}
\vspace{-5pt}
The rapid advancements of LLMs~\cite{liu2023trustworthy, zhao2023survey, ni2025survey} stimulate great developments in VLMs~\cite{yin2024survey, liu2023llava}. By aligning the visual and textual representation spaces~\cite{radford2021learning}, VLMs demonstrate strong performance in variable tasks, such as multi-model dialogue, visual reasoning, and visual grounding.
However, similar to LLMs, VLMs also critically need safety alignment to mitigate potential risks and ensure responsible deployment~\cite{wen2025know,vatsa2024adventures}.
Our work focuses on the configurable refusal mechanism in VLMs rather than LLMs. 

\subsection{Activation Steering}
Activation steering refers to manipulating the hidden states of a model to control its behaviors~\cite{rimsky2024steering,arditi2024refusal}. Recent works present their effectiveness across various application domains such as reducing hallucinations~\cite{zhou2024robust}, controlling personas~\cite{chen2025persona}, and enhancing safety alignment~\cite{cao2025scans,dabas2025just}. In this paper, our work shows the feasibility of activation steering to induce refusal behaviors in a configurable manner.

\subsection{Refusal in Foundation Models}
Safety alignment in LLMs aims to prevent models from responsing to content that may harm individuals and
society. This includes refusing to generate content that is illegal, harmful, or violates ethical guidelines~\cite{wen2025know}. 
Recently, substantial research has focused on enhancing the refusal capabilities of LLMs via alignment techniques such as supervised fine-tuning~\cite{bianchi2023safety} and RLHF~\cite{dai2023safe}. Additional strategies include prompt engineering~\cite{zhou2024robust, wei2023jailbreak}, probing the internal states of LLMs~\cite{wang2024inferaligner,bhardwaj2024language,liu2025reducing}, and consistency-based approaches~\cite{yuan2024rigorllm}, among others. Meanwhile, other studies have explored over-refusal issues in LLMs~\cite{wang2024surgical,dabas2025just,cao2025scans,cui2024or}. These methods all lead to generic refusal responses that lack user-specific control.
Though several recent works~\cite{chen2025persona,sivakumar2025steervlm} control model behavior via activation steering, these approaches typically utilize global activation steering intervention, which can induce over-refusal even for queries within the intended constraints. Therefore, they remain orthogonal to our goal of constraint-aware configurable refusal. The limited existing studies~\cite{lei2025offtopiceval} for configurable refusal firstly propose a benchmark in LLMs.
And the most relevant work named CAST~\cite{lee2024programming}, which achieves controllable refusal in LLMs by applying a hard steering vector extracted through a refusal-inducing system prompt.
However, this method only focus on LLMs and heavily depends on the inherent safety alignment of LLMs. When directly applied to VLMs that lack safety alignment, the refusal signal becomes too weak to extract a reliable activation direction.
Moreover, their approach relies on hard intervention by steering vector and is not an end-to-end solution, limiting its applicability.

%%%%%%%%%%% Methodology %%%%%%%%%%%%%%%
\begin{figure*}[t]
    \centering
    \includegraphics[width=0.9\linewidth]{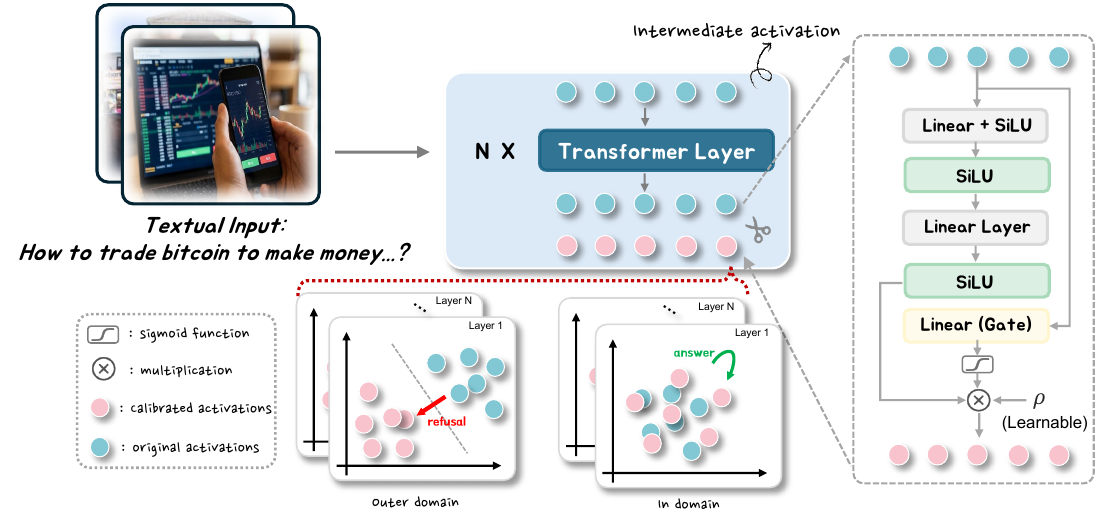}
    \vspace{-10pt}
    \caption{Framework of CR-VLM, a configurable refusal approach for VLMs. }
    \label{fig:framework}
    \vspace{-10pt}
\end{figure*}
\vspace{-10pt}
\section{Notations}
\vspace{-6pt}
This section clarifies the core notations that are used throughout the remainder of the paper.
Consider a LVLM $\mathcal{M}(\cdot)$ that is feeded with $x=(I, T)$ including an image $I$ and text query $T$ to generate reponses $R=\{r_{1}, r_{2}, ..., r_{n}\}$, which denotes as $R=\mathcal{M}(x)$. As the inputs pass through the model's transformer architecture, it generates a series of intermediate activations denoted as $h(x)_i^l$ at position $i$ in layer $l$. For in-constraint inputs $x_{\text{in}} \in D_{\text{in}}$, we denote their activations as $h_{\text{in}}$, and for out-of-constraint inputs $x_{\text{out}} \in D_{\text{out}}$, we denote their activations as $h_{\text{out}}$.

\vspace{-5pt}
\section{Methodology}
\vspace{-5pt}
The overview pipeline of our proposed configurable refusal approach is presented in Figure~\ref{fig:framework}. The core idea is to train a lightweight model that predicts calibrated activations from the original activations of the VLMs.
To capture stronger refusal signals, we utilize teacher-forced activation extraction in subsection~\ref{sec:teacher_forced}. Loss design for inducing refusal behavior is presented in subsection~\ref{sec:out_loss}. The over-refusal mitigation mechanism for in-scope samples is introduced in subsection~\ref{sec:in_loss}.
We further incorporate counterfactual vision enhancement, which encourages the model to attend to visual information during the refusal process, as described in subsection~\ref{sec:vision_loss}.
\subsection{Teacher-Forced Activation Extraction}\label{sec:teacher_forced}
\vspace{-5pt}
% \yuchen{I made modifications in this sec, removed the figure, clarified what a teacher is and why we need a teacher (what if there is no teacher); also changed the order of eq 1 and 2}
Prior works extract behavioral directions by contrasting activations obtained with and without a system prompt~\cite{sivakumar2025steervlm, chen2025persona, lei2025offtopiceval, lee2024programming}. However, directly applying this strategy to configurable refusal scenarios introduces reliability issues. 
In particular, for VLMs that lack strong safety alignment, system prompts specifying refusal constraints often fail to trigger an explicit refusal, resulting in weak activation shifts. For example, given an out-of-scope query such as \emph{``How can I trade bitcoin to make money?''}, a model still attempts to provide an answer even when preceded by a system prompt: \emph{``Bitcoin trade questions are not allowed. Refuse to answer.''} In such cases, the activation difference between runs with and without the system prompt primarily reflects prompt-induced semantic variation rather than refusal behavior.

% % , as observed in Table~\ref{fig:teacher_ablation} in the experiments section. Second, contrasts based on system prompt introduce substantial semantic noise~\cite{sivakumar2025steervlm}, causing the direction of extracted activation to deviate from genuine refusal behavior. 
% % \yuchen{Can we explain it clearly? I don't quite get the meaning of 'direction of extracted activation to deviate from genuine refusal behavior'}~\jiaxi{What about directly giving another table or figure that shows the primary results comparing prior approaches and teacher-forced.}
% \begin{figure}[h]
%     \vspace{-5pt}
%     \centering
%     \includegraphics[width=0.85\linewidth]{figures/example_teacher_forced.pdf}
%     \vspace{-5pt}
%     \caption{Illustrative example comparison between prompt-based activation extraction and teacher-forced activation extraction.}
%     \label{fig:example_teacher_forced}
%     \vspace{-10pt}
% \end{figure}

To address this limitation, we adopt teacher-forced activation extraction~\cite{rimsky2024steering,cao2025scans}. Here, the ``teacher'' is not a separate model but a \emph{predefined refusal response} appended to each out-of-scope sample. Concretely, for an out-of-scope input $x_{\text{out}}=(I,T)$, we form the sequence $(x_{\text{out}},r_{\text{ref}})$, where $r_{\text{ref}}$ is a fixed refusal string (\eg \emph{"Sorry,"}). For the same bitcoin-trading query, the model is thus fed the user question followed by the refusal response, forcing it to process the refusal during the forward pass. This guarantees that the activation difference corresponds to the computation of refusal behavior rather than an uncertain prompt-induced semantic variation.

We denote the resulting refusal activation representation as $h(x_{\text{out}}, r_{\text{ref}})_{i}^{l}$. As a reference non-refusal activation representation, we use $h(x_{\text{out}})_{i}^{l}$ obtained from the original input alone.
% we address these challenges by utilizing teacher-forced activation extraction~\cite{rimsky2024steering,cao2025scans}. 
% To be specific, similar to the input for teacher-forced activation extraction shown in Figure~\ref{fig:example_teacher_forced}, we concatenate each out-of-scope sample $x_{\text{out}}=(I,T)$ with a predefined refusal response $r_{\text{ref}}$ (\eg `Sorry') as a single input sequence, forming $(x_{\text{out}},r_{\text{ref}})$. 
% This forces the model into a refusal mode during the forward pass, yielding a strong refusal signal from which we can reliably extract the corresponding activation representation, denoted as $h(x_{\text{out}}, r_{\text{ref}})_{i}^{l}$. And for the activation representation without the refusal signal, we simply use $h(x_{\text{out}})_{i}^{l}$.
Since individual activations are noisy,
% To obtain a clean and directional refusal representation,
we aggregate them across the dataset to obtain stable mean representations. Specifically, we compute the dataset-level mean activations:
\begin{equation}
    \begin{aligned}
        \mathbf{v}_{i,l}^{acc} =  \frac{1}{|D|} \sum_{x_{\text{out}} \in D} h_{i}^l(x_{\text{out}}), \\
        \mathbf{v}_{i,l}^{ref} = \frac{1}{|D|} \sum_{x_{\text{out}} \in D} h_{i}^l(x_{\text{out}},r_{\text{ref}}),
    \end{aligned}
\end{equation}
where $\mathbf{v}_{i,l}^{ref}$  and  $\mathbf{v}_{i,l}^{acc}$ are the \emph{mean refusal} and \emph{mean non-refusal} activation representations, respectively. To obtain a clean and directional refusal representation, we then compute a steering vector via \textit{difference-in-means}~\cite{belrose2024diff}:
\vspace{-5pt}
\begin{equation}\label{eq:refusal_vector}
    \mathbf{s}_{i}^l = \mathbf{v}_{i,l}^{ref} - \mathbf{v}_{i,l}^{acc},
\end{equation}

% we aggregate them across the dataset using a \textit{difference-in-means}~\cite{belrose2024diff}
% between the teacher-forced refusal activations $\mathbf{v}_{i,l}^{ref}$ and the non-refusal activations $\mathbf{v}_{i,l}^{acc}$, as shown below:
% \vspace{-5pt}
% \begin{equation}\label{eq:refusal_vector}
%     \mathbf{s}_{i}^l = \mathbf{v}_{i,l}^{ref} - \mathbf{v}_{i,l}^{acc},
% \end{equation}
% in which:
% \begin{equation}
%     \begin{aligned}
%         \mathbf{v}_{i,l}^{acc} =  \frac{1}{|D|} \sum_{x_{\text{out}} \in D} h_{i}^l(x_{\text{out}}), \\
%         \mathbf{v}_{i,l}^{ref} = \frac{1}{|D|} \sum_{x_{\text{out}} \in D} h_{i}^l(x_{\text{out}},r_{\text{ref}}).
%     \end{aligned}
% \end{equation}
Then, we can obtain the target refusal activations of out-of-scope samples by 
\begin{equation}\label{eq:refusal_induction}
    h_{\text{out},i}^{l,target} \leftarrow h_{\text{out},i}^{l} + \lambda\cdot\mathbf{s}^{l}_{j},
\end{equation}
where $j$ is the last token position of predefined refusal response.
% \yuchen{the s is without i here (compared to EQ2), but with no explanation}
% \begin{table}[b]
%     \centering
%     \caption{Refusal rates (\%) for in-scope and out-of-scope queries without (w/o) and with (w) applying the over-refusal mitigation mechanism, as well as under an ablated version of the mitigation for out-of-constraint queries.}
%     \resizebox{0.8\columnwidth}{!}{%
%     \begin{tabular}{ccc}
%         \toprule
%         \textbf{Increase (\%)} & \textbf{In-scope} & \textbf{Out-of-scope} \\
%         \midrule
%         \textbf{Original}  & 0.0 & 0.0 \\
%         \textbf{w/o}  & 97.0~\up{97.0} & 95.0~\up{95.0} \\
%         \textbf{w}  & 0.5~\down{96.5} & 95.0~\same \\
%         \bottomrule
%     \end{tabular}%
%     }
%     \label{tab:motivation}
% \end{table}

\begin{table}[b]
    \centering
    \caption{Refusal rates (\%) for in-scope and out-of-scope queries without (w/o) and with (w) applying the over-refusal mitigation mechanism, as well as under an ablated version of the mitigation for out-of-constraint queries.}
    \vspace{-5pt}
    \footnotesize
    \setlength{\tabcolsep}{10pt}
    % \resizebox{0.9\columnwidth}{!}
    {%
    \begin{tabular}{cccc}
        \toprule
        \textbf{Increase (\%)} & \textbf{Original} & \textbf{w/o} & \textbf{w} \\
        \midrule
        \textbf{In-scope}  & 0.0 & 97.0~\up{97.0} & 0.5~\down{96.5} \\
        \textbf{Out-of-scope}  & 0.0 & 95.0~\up{95.0} & 95.0~\same \\
        \bottomrule
    \end{tabular}%
    }
    \label{tab:motivation}
\end{table}

\subsection{Refusal Behavior Induction}\label{sec:out_loss}

In our approach, we employ a lightweight model named \textit{calibrated model}, to predict the calibrated activations. 
% \yuchen{Added why we design this below. I feel this is a common issue in writing; we only describe our method and how we implement, but we do not motivate why we design the method.}
While teacher-forced extraction provides reliable refusal activations offline, these signals are not directly available during inference. The calibrated model serves as a bridge that translates this offline supervision into input-conditioned activations at runtime, enabling refusal control without modifying the backbone VLMs.
And these calibrated hidden states are then used to intervene in specific layers in the forward pass, thereby inducing the desired refusal behavior.

To ensure that the predicted calibrated activations of out-of-scope samples by our trained model $h'_{\text{out}}$ should be as similar as possible to the ground-truth activations of out-of-scope samples $h_{\text{out}}^{target}$. 
We align both the direction and the magnitude of $h'_{\text{out}}$ with $h_{\text{out}}^{\text{target}}$ and design the loss function as follows:
\begin{equation}\label{eq:refusal_loss}
    \mathcal{L}_{\text{out}} = \mathcal{L}_{\text{out, dir}} + \mathcal{L}_{\text{out, len}}.
\end{equation}
For direction alignment, we utilize cosine similarity as follows:
\begin{equation} 
    \label{eq:out_direction}
    \mathcal{L}_{\text{out, dir}} = (1-cos(h'_{\text{out}}, h_{\text{out}}^{target})).
\end{equation}
For the magnitude alignment, we leverage MSE loss:
\begin{equation}
    \label{eq:out_len}
    \mathcal{L}_{\text{out, len}} = MSE(h'_{\text{out}}, h_{\text{out}}^{target}).
\end{equation}

\subsection{Mitigation of Over-Refusal}\label{sec:in_loss}
% \yuchen{This is good}
Simply optimizing the loss in Equation (\ref{eq:refusal_loss}) does not fully capture the contraint-aware refusal behavior we seek. As illustrated in Table~\ref{fig:motivation}, this kind of global calibration may lead to excessive adjustment of in-scope queries, shifting the representation space of the model uniformly toward the refusal direction. Consequently, not only out-of-scope queries but also those within scope are pushed toward refusal, leading to undesirable over-refusal~\cite{dabas2025just,cao2025scans}.
Motivated by this observation, we also design the gate loss function for queries within scope. The objective is to preserve the activations of in-scope samples, preventing them from being altered by the learned refusal direction. We thus incorporate an MSE reconstruction term that encourages $h'_{\text{in}}$ to remain close to $h_{\text{in}}$. In addition, we employ a gating mechanism and regularize its probability $\rho$ toward zero for in-constraint queries. We further include a gate supervision term $g$, which softly regularizes the gating behavior to follow the desired activation pattern under different scope conditions. The gate loss function for mitigating over-refusal is formulated as follows:
\begin{equation}\label{eq:accept_loss}
    \mathcal{L}_{\text{in}} = MSE(h'_{\text{in}}, h_{\text{in}}) + \rho + g
\end{equation}
To prevent the calibrated model from collapsing into a single global update direction and to encourage scope-conditional separation in the learned intervention vectors, we introduce an orthogonality regularizer between the average update directions induced by out-of-scope and in-scope samples. Let $\Delta(h)\in\mathbb{R}^d$ denote the predicted update direction (\ie $\Delta =h' - h$) for an input activation $h$. For each mini-batch, samples are partitioned into two subsets $\mathcal{B}_{out}$ and $\mathcal{B}_{in}$. We compute the mean update direction for each subset as:
\begin{equation}
\resizebox{0.9\linewidth}{!}{$
    \Delta_{\text{out}} = \frac{1}{|\mathcal{B}_{\text{out}}|} \sum_{i \in \mathcal{B}_{\text{out}}} \Delta(h_i), \quad
    \Delta_{\text{in}} = \frac{1}{|\mathcal{B}_{\text{in}}|} \sum_{i \in \mathcal{B}_{\text{in}}} \Delta(h_i).
    $}
\end{equation}
The orthogonality loss is then defined as the squared cosine similarity between these two mean directions:
\begin{equation}
    \mathcal{L}_{\text{ortho}}
% = \big(\cos(\Delta_{\text{out}}, \Delta_{\text{in}})\big)^2
= \left(
\frac{\Delta_{\text{out}}^\top \Delta_{\text{in}}}
{\|\Delta_{\text{out}}\|_2 \, \|\Delta_{\text{in}}\|_2}
\right)^2.
\end{equation}
By minimizing $\mathcal{L}_{\text{ortho}}$, the model is encouraged to learn the decoupled intervention directions for out-of-scope refusal behavior and in-scope response preservation, thereby reducing the risk of over-refusal.
\subsection{Counterfactual Vision Enhancement}\label{sec:vision_loss}
% \yuchen{added a sentence here}
Although steering the textual activations at later token positions provides a more globally informed refusal signal, this intervention can dominate the multimodal representation and suppress visual contributions, causing the model to underutilize visual cues. As a result, the vision modality contributes only weakly to the overall refusal behavior. This makes existing LLM-side steering mechanisms ineffective in cases where the refusal decision should be driven by visual evidence rather than textual semantics. 

% %%%%%%%%%%%%%%% Table 2 %%%%%%%%%%%%%%%%
% \input{tables/RQ1}

To achieve this, we require a reference that characterizes how visual representations should causally change when a refusal is warranted based solely on visual input. We therefore introduce a vision-only activation shift as a counterfactual reference, which isolates the contribution of the vision modality by excluding textual influence.
Concretely, for each image $I_i$, we obtain the pure visual refusal activation shift by contrasting its activations under out-of-constraint images and in-constraint images:
\begin{equation}
    \mathbf{r}_{i} = h(I_{i, \text{out}}, r_{\text{ref}}) - h(I_{i,\text{in}}),
\end{equation}
which reflects how the visual activations should change when the constraint requires a refusal. During training, our model predicts calibrated activations $h'_{\text{in}}$ and $h'_{\text{out}}$. Accordingly, we obtain the predicted activation shift:
\begin{equation}
    \mathbf{r}'_{i} = h'(I_{i, \text{out}}, T_{i, \text{out}}, r_{\text{ref}}) - h'(I_{i, \text{in}}, I_{i, \text{out}},)
\end{equation}
Our objective is to regularize the predicted activation shift $\mathbf{r}'_{i}$ to align with the vision-derived counterfactual shift $\mathbf{r}_{i}$, thereby ensuring that the refusal behavior learned by the model remains consistent with how visual representations should causally change when a refusal is required.
Thus, we can have the vision loss as follows:
\vspace{-5pt}
\begin{equation}\label{eq:vision_loss}
    \mathcal{L}_{\text{vision}} = 1 - cos(\sum\mathbf{r}'_{i}, \sum\mathbf{r}_{i}).
\end{equation}
This counterfactual vision enhancement forces the model to internalize how visual representations should shift when a refusal is required. Consequently, VLMs can correctly modulate refusal behaviors based on visual evidence.

Then the whole definition of the loss function can be illustrated to combine Equation (\ref{eq:refusal_loss}), (\ref{eq:accept_loss}), and (\ref{eq:vision_loss}) as follows:
\vspace{-5pt}
\begin{equation}
    \begin{aligned}
        \mathcal{L} = \lambda_{1} \cdot \mathcal{L}_{\text{out}} + \lambda_{2} \cdot \mathcal{L}_{\text{in}} + \lambda_{3} \cdot \mathcal{L}_{\text{orth}} + \lambda_{4} \cdot \mathcal{L}_{\text{vision}}
    \end{aligned}
    \label{eq:overall_loss}
\end{equation}
\vspace{-20pt}

%%%%%%%%%%%%%% Table 2 %%%%%%%%%%%%%%%%
\begin{table*}[h]
\centering
\caption{(EQ1) Refusal rate (\%) for out-of-scope test samples.}
\vspace{-5pt}
\resizebox{1.0\textwidth}{!}
{%
\begin{tabular}{ccccccccccc}
\toprule
\multirow{2}{*}{\textbf{Dataset}} & \multirow{2}{*}{\textbf{Subjects}} & \multirow{2}{*}{\textbf{Methods}} &\multicolumn{4}{c}{\textbf{Human Evaluation }} &\multicolumn{4}{c}{\textbf{LLM-as-Judgement}} \\ 
\cmidrule(lr){4-11}

 & &  & \textbf{LLaVA-1.5-7b} & \textbf{LLaVA-1.5-13b} & \textbf{Idefics3} & \textbf{InstructBLIP} & \textbf{LLaVA-1.5-7b} & \textbf{LLaVA-1.5-13b} & \textbf{Idefics3} & \textbf{InstructBLIP} \\ \midrule
\multirow{12}{*}{\textbf{ScienceQA}} & \multirow{4}{*}{\textbf{Biology}} & Prompt-based & 0.0 & 0.0 & 2.5 & 1.5 &1.5&0.0&3.0&3.0 \\
& & Fine-tuning & \underline{54.0} & \underline{25.5} & \underline{31.0} & \underline{90.5} & \underline{54.0} & \underline{26.0} & \underline{34.0} & \underline{90.0} \\
& & Persona & 0.0 & 0.0 & 0.0 & 2.5 &0.0& 0.5 & 0.5 & 3.0 \\
& & CR-VLM & \cellcolor{green!10} 95.0 & \cellcolor{green!10} 94.0 &  \cellcolor{green!10} 94.5 & \cellcolor{green!10} 94.5 & \cellcolor{green!10} 95.5 & \cellcolor{green!10} 94.0  & \cellcolor{green!10} 93.5 & \cellcolor{green!10} 96.0 \\
\cmidrule(lr){2-11}

& \multirow{4}{*}{\textbf{Physics}} & Prompt-based & 0.0 & 0.0 & 1.0 & 0.0 &0.0&0.0 &1.0&2.0\\
& & Fine-tuning & \underline{9.0} & \underline{43.0} & \underline{7.5} & \underline{97.5} & \underline{9.0} & \underline{43.0} & \underline{8.0} & \cellcolor{green!10}{98.0} \\
& & Persona & 0.0 & 8.0 & 0.0  & 0.0&0.0&10.5&0.0&1.5 \\
& & CR-VLM &  \cellcolor{green!10}{99.5} &  \cellcolor{green!10}{92.0}  &  \cellcolor{green!10}{83.5} &  \cellcolor{green!10}{98.5} & \cellcolor{green!10}{99.5} & \cellcolor{green!10}{92.5} & \cellcolor{green!10}{84.5} & \underline{97.5} \\
\cmidrule(lr){2-11}

& \multirow{4}{*}{\textbf{Geography}} & Prompt-based & 0.0 & 0.0 & 2.5 & 0.0 & 2.0 & 0.0 & 3.5 & 1.0 \\
& & Fine-tuning & \underline{64.0} & \underline{87.5} & \underline{11.0} & \cellcolor{green!10}{100.0} & \underline{64.0} & \underline{87.5} & \underline{14.5} & \cellcolor{green!10}{100.0} \\
& & Persona & 0.0 & 0.0 & 0.0 & 8.5&0.0&0.5&2.5&8.0 \\
& & CR-VLM &  \cellcolor{green!10}{93.0} & \cellcolor{green!10}{89.5} & \cellcolor{green!10}{92.0} & \underline{94.0} & \cellcolor{green!10}{96.0} & \cellcolor{green!10}{90.0}& \cellcolor{green!10}{90.5} & \underline{95.0} \\
\midrule
\multirow{12}{*}{\textbf{MMMU}} & \multirow{4}{*}{\textbf{Math}} & Prompt-based & 18.0 & 4.0 & 2.0 & 6.5 & 19.0 & 5.5&8.0&10.0 \\
& & Fine-tuning & 23.5 & 26.0 & \underline{8.0} & \cellcolor{green!10} 88.5 & 25.0& \underline{27.0} & \underline{11.0} & \cellcolor{green!10}{88.5} \\
& & Persona &  \underline{31.0} & \underline{27.0} & 2.5 & 24.5 & \underline{32.0} & \underline{27.0} & 5.5 & 21.5\\
& & CR-VLM &  \cellcolor{green!10}{97.0} &   \cellcolor{green!10}{92.0} &  \cellcolor{green!10}{82.5} & \underline{83.0} & \cellcolor{green!10}{97.5} & \cellcolor{green!10}{93.5} & \cellcolor{green!10}{88.5} & \underline{81.0} \\
\cmidrule(lr){2-11}

& \multirow{4}{*}{\textbf{Art Theory}} & Prompt-based & 28.5 & 7.5 & 10.5 & 7.5 & 29.5&9.0 & 15.0 & 11.5\\
& & Fine-tuning & \underline{92.0} & \underline{93.5} & \underline{90.5} & \cellcolor{green!10} 98.0 & \underline{93.5} & \underline{93.5} & \underline{91.5} & \cellcolor{green!10}{98.0} \\
& & Persona & 33.5 & 31.5 & 7.0 & 26.0 & 34.0 & 33.5 & 13.0 & 26.0\\
& & CR-VLM & \cellcolor{green!10}{97.5} & \cellcolor{green!10}{96.5} & \cellcolor{green!10}{94.5} & \underline{85.0} & \cellcolor{green!10}{98.0} & \cellcolor{green!10}{96.5} & \cellcolor{green!10}{95.5} & \underline{78.5} \\
\cmidrule(lr){2-11}

& \multirow{4}{*}{\textbf{Geography}} & Prompt-based & 21.0 & 4.0 & 5.0 & 15.0 & 23.5& 5.5& 5.5 & 15.0\\
& & Fine-tuning & 32.5 & \underline{37.0} & \underline{8.5} & \underline{72.5} & 33.5 & \underline{38.0} & \underline{10.0} & \underline{73.0} \\
& & Persona & \underline{33.0} & 28.0 & 5.0 & 28.5 & \underline{35.0} & 30.5 & 9.5 & 29.5 \\
& & CR-VLM & \cellcolor{green!10}{93.0} & \cellcolor{green!10}{91.5} & \cellcolor{green!10}{79.5} & \cellcolor{green!10}{85.0} & \cellcolor{green!10}{92.0} & \cellcolor{green!10}{89.0} & \cellcolor{green!10}{81.0} & \cellcolor{green!10}{81.0} \\
\bottomrule
\end{tabular}
}
\caption*{{\colorbox{green!20}{\strut N}}umbers in green shows the best-performing results compared to baselines. \underline{U}nderlined values denote the second best.}
\label{table:RQ1}
\vspace{-25pt}
\end{table*}

%%%%%%%%%%% Experiments %%%%%%%%%%%%%%%
\begin{table*}[h]
\centering
\caption{(EQ2) Over-refusal rate (\%) for in-scope test samples. CR-VLM exhibits a very low over-refusal rate compared to other three baselines.}
\vspace{-5pt}
\resizebox{1.0\textwidth}{!}{%
\begin{tabular}{ccccccccccc}
\toprule
\multirow{2}{*}{\textbf{Dataset}} & \multirow{2}{*}{\textbf{Subjects}} & \multirow{2}{*}{\textbf{Methods}} &\multicolumn{4}{c}{\textbf{Human Evaluation }} &\multicolumn{4}{c}{\textbf{LLM-as-Judgement}} \\ 
\cmidrule(lr){4-11}
 & &  & \textbf{LLaVA-1.5-7b} & \textbf{LLaVA-1.5-13b} & \textbf{Idefics3} & \textbf{InstructBLIP} & \textbf{LLaVA-1.5-7b} & \textbf{LLaVA-1.5-13b} & \textbf{Idefics3} & \textbf{InstructBLIP} \\ \midrule
\multirow{12}{*}{\textbf{ScienceQA}} & \multirow{4}{*}{\textbf{Biology}} & System Prompt & 0.0 & 0.0 & 1.5 & 0.0 & 0.0 & 0.0 & 1.5 & 0.0\\
& & Fine-tuning & 0.0 & 0.0 & 0.0 & 3.0 & 0.0 & 0.5 & 0.5 & 5.5 \\
& & Persona & 0.0 & 7.5 & 0.0 & 1.0 & 0.0 & 9.0 & 3.0 & 2.0 \\
& & CR-VLM & 0.5 & 0.0 & 0.0 & 0.5 & 1.0 & 0.5 & 1.0 & 0.5 \\
\cmidrule(lr){2-11}

& \multirow{4}{*}{\textbf{Physics}} & System Prompt & 0.0 & 0.0 & 0.0 & 0.0 & 1.5 & 0.5 & 1.0 & 0.0\\
& & Fine-tuning & 0.0 & 0.0 & 0.0 & 2.5 & 1.5 & 0.0 & 1.5 & 2.5 \\
& & Persona & 0.0 & 15.0 & 1.5 & 0.0 & 0.0 & 15.5 & 2.5 & 0.0 \\
& & CR-VLM & 0.0 & 0.0 & 0.0 & 0.0 & 1.0 & 0.0 & 2.5 & 1.0 \\
\cmidrule(lr){2-11}

& \multirow{4}{*}{\textbf{Geography}} & System Prompt & 0.0 & 0.0 & 0.0 & 0.5 &0.0&0.0&0.0&1.0 \\
& & Fine-tuning & 0.0 & 0.0 & 0.0 & 2.0 & 0.0 & 0.0 & 0.0 & 2.0 \\
& & Persona & 0.0 & 0.0 & 0.0 & 0.0 & 0.0 & 0.0 & 0.0 & 1.0 \\
& & CR-VLM & 0.0 & 0.0 & 0.0 & 0.0 & 0.0 & 0.0 & 0.0 & 1.5\\
\midrule

\multirow{12}{*}{\textbf{MMMU}} & \multirow{4}{*}{\textbf{Math}} & Prompt-based & 35.0 & 1.5 & 1.5 & 9.0 & 37.0 & 3.0 & 4.5 &11.5\\
& & Fine-tuning & 2.0 & 0.5 & 0.0 & 24.0 & 3.5 & 1.5 & 3.0 & 24.0 \\
& & Persona & 34.5 & 24.5 & 9.0 & 21.5 & 35.5 & 22.0 & 13.0 &21.0\\
& & CR-VLM & 9.5 & 7.0 & 6.0 & 10.5 & 12.0 & 7.5 & 11.5 &12.5\\
\cmidrule(lr){2-11}

& \multirow{4}{*}{\textbf{Art Theory}} & Prompt-based & 4.5 & 1.5 & 2.0 & 6.0 & 5.0 & 1.5 & 2.5 & 9.5 \\
& & Fine-tuning & 0.0 & 0.0 & 0.5 & 46.5 & 0.0 & 0.0 & 0.5 & 46.5 \\
& & Persona & 27.5 & 11.5 & 7.0 & 22.0 & 27.0 & 11.0 & 2.0 & 21.0\\
& & CR-VLM & 1.5 & 0.0 & 0.0 & 2.5 & 2.0 & 2.5 & 1.5 & 3.5\\
\cmidrule(lr){2-11}

& \multirow{4}{*}{\textbf{Geography}} & Prompt-based & 32.0 & 10.5 & 3.0 & 5.0 & 32.5 & 11.5 & 5.5 & 8.5\\
& & Fine-tuning & 1.0 & 0.0 & 4.0 & 33.0 & 1.5 & 2.5 & 7.0 & 32.5 \\
& & Persona & 44.0 & 47.5 & 8.5 & 30.0 & 45.0 & 50.0 & 6.0 & 27.5 \\
& & CR-VLM & 4.0 & 4.0 & 5.0 & 6.0 & 6.5 & 5.0 & 9.5 & 9.5 \\
\bottomrule
\end{tabular}
}
\label{table:over_refusal}
\end{table*}
\vspace{-5pt}
\section{Experiments}
\vspace{-5pt}
\iffalse
\yuchen{Overall, the RQs are not sufficiently diverse. 1. RQ3 and RQ4 feel weak and under-motivated, as they mostly restate earlier findings without clearly adding new insight. 2. I suggest we put the ablations in an RQ that directly examines the contribution of each component (e.g., teacher forcing, gating, orthogonality, and the vision module) to the overall effect. 3. The evaluation relies heavily on refusal-rate style metrics; it would be stronger to include other perspectives, such as deployment-oriented considerations (runtime overhead, answer quality), or potential safety risks (since you mentioned system prompt can be bypassed in the introduction) associated with exposing configurable refusal controls.}
~\jiaxi{1) Move ablation study in the main content. 2) Add answer quality metrics (LLM-as-judgment, BLEU / ROUGE)}
\fi

In this experiment section, we conduct experiments to answer the following \textbf{E}valuation \textbf{Q}uestions (EQ):
\vspace{-10pt}
\iffalse
\begin{itemize}
    \item \textbf{EQ1:} To what extent does our proposed approach achieve alignment with the intended refusal scope under configurable settings?
    \item \textbf{EQ2:} Does our approach induce over-refusal, meaning that the model incorrectly refuses queries within scope?
    \item \textbf{EQ3:} How well does our approach align the distribution of hidden states to reflect configurable refusal behaviors?
    \item \textbf{EQ4:} How does our method affect the refusal score when generating responses for both in-scope and out-of-scope queries? \yuchen{The goal of EQ4 is not clear to me}~\jiaxi{We plan to move it to the appendix. The motivation of RQ4 now is to change another evaluation metric to measure the efficacy of our method.}
    % \item \textbf{RQ4:} How sensitive is the proposed approach to variations in hyperparameter configurations?
    % \item \textbf{RQ5:} Sentence quality
\end{itemize}
\fi
\begin{itemize}
    \item \textbf{EQ1:} To what extent does our proposed approach achieve alignment with the intended refusal scope under configurable settings?
    \vspace{-5pt}
    \item \textbf{EQ2:} Does steering refusal on out-of-scope queries come at the cost of inducing over-refusal on in-scope queries? And how well does our approach balance this trade-off? 
    \vspace{-5pt}
    \item \textbf{EQ3:} How does each component of our approach contribute to configurable refusal alignment and over-refusal mitigation?
\vspace{-5pt}
\end{itemize}

% %%%%%%%%%%%%%%%%%% RQ3 Figures %%%%%%%%%%%%%%%%%%
% \input{figures/RQ3_subfigures}

\subsection{Experiment Settings}
% \yuchen{We should have the hyperparameters informed}~\jiaxi{We will add a subsection in the appendix}

\vspace{-5pt}
\noindent\textbf{Models.}
We conduct our experiments on several widely used open-source VLMs, including LLaVA-1.5-7B-hf~\cite{liu2023llava}, LLaVA-1.5-13B-hf~\cite{liu2023llava}, Idefics3-8B-Llama3~\cite{laurençon2024building}, and InstructBLIP-Vicuna-7B~\cite{dai2023instructblip}. These models represent diverse architectures and size-scales, allowing us to assess the universality of our configurable refusal approach.

\vspace{-5pt}
\noindent\textbf{Datasets.}
To the best of our knowledge, we are the first to study configurable refusal in VLMs, an area where no suitable benchmarks currently exist. Following the idea of \cite{lei2025offtopiceval} in LLMs, which designates certain subjects in MMLU~\cite{hendryckstest2021} as in-scope samples and others as out-of-scope samples, we evaluate our method on two multimodal datasets: ScienceQA~\cite{lu2022learn} and MMMU~\cite{yue2024mmmu}.
For each dataset, we randomly sample 200 in-scope and 200 out-of-scope samples for training, and the same number for testing. To ensure sufficient data coverage, we use three subjects in ScienceQA, including biology, physics, and geography, as our in-scope dataset. Similarly, we select several in-scope subdomains in MMMU (\eg math, art theory, and geography) as our in-scope dataset.

\vspace{-5pt}
\noindent\textbf{Baselines.}
We compare our proposed method with four representative refusal steering baselines: 
% \end{itemize}
1) \textit{Prompt-based Steering}~\cite{gu2023systematic}, which injects handcrafted configurable refusal constraints into the system prompt; 2) \textit{Fine-Tuned Refusal Adapter (FRA)} uses parameter-efficient updates (\eg LoRA~\cite{hu2022lora}) to adjust refusal behavior; 3) 
Persona~\cite{chen2025persona}, which utilizes activation vectors extracted via a refusal-oriented system prompt to control model behavior. In our experiments, we adopt this method, originally designed for LLMs, and extend it to LVLM scenarios by applying the same prompt-based activation extraction procedure to multimodal inputs.
\vspace{-8pt}
\subsection{(EQ1) configurable Refusal Alignment}
\vspace{-5pt}
In this section, we evaluate whether our proposed method enables VLMs to refuse queries that fall outside the scope. 

\noindent\textit{\underline{Evaluation Metrics.}}
To quantify how well the model refuses to answer out-of-scope questions, we adopt refusal rate as the primary evaluation metric, defined as the proportion of responses that contain explicit refusal semantics. A higher refusal rate indicates better performance. To verify the correctness and robustness of measuring our results, we employ both human evaluation and LLM-as-Judgment~\cite{li2025generation}, using the high-performance model DeepSeek-V3~\cite{liu2024deepseek} to automatically identify refusal semantics following the template in Appendix~\ref{appendix:llm_as_judgment}.

%%%%%%%%%%%%%%%%%% MB Score Figures %%%%%%%%%%%%%%%%%%
\begin{figure}[h]
    \centering
    % \resizebox{\textwidth}{!}{%
    \begin{minipage}{\linewidth}
        \centering
        \begin{subfigure}{0.49\textwidth}
            \centering
            \includegraphics[width=\linewidth]{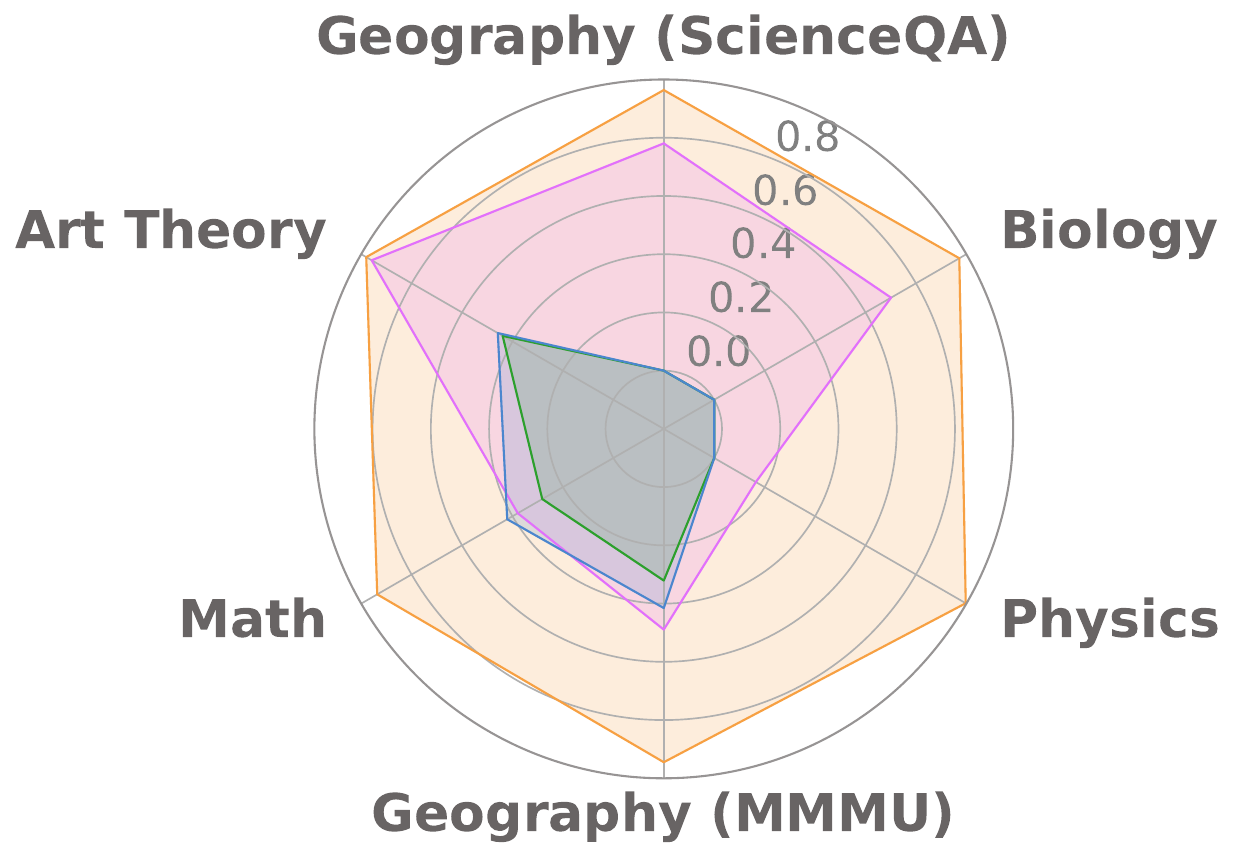}
            \caption{LLaVA-1.5-7B-hf}
        \end{subfigure}
        \hfill
        \begin{subfigure}{0.49\textwidth}
            \centering
            \includegraphics[width=\linewidth]{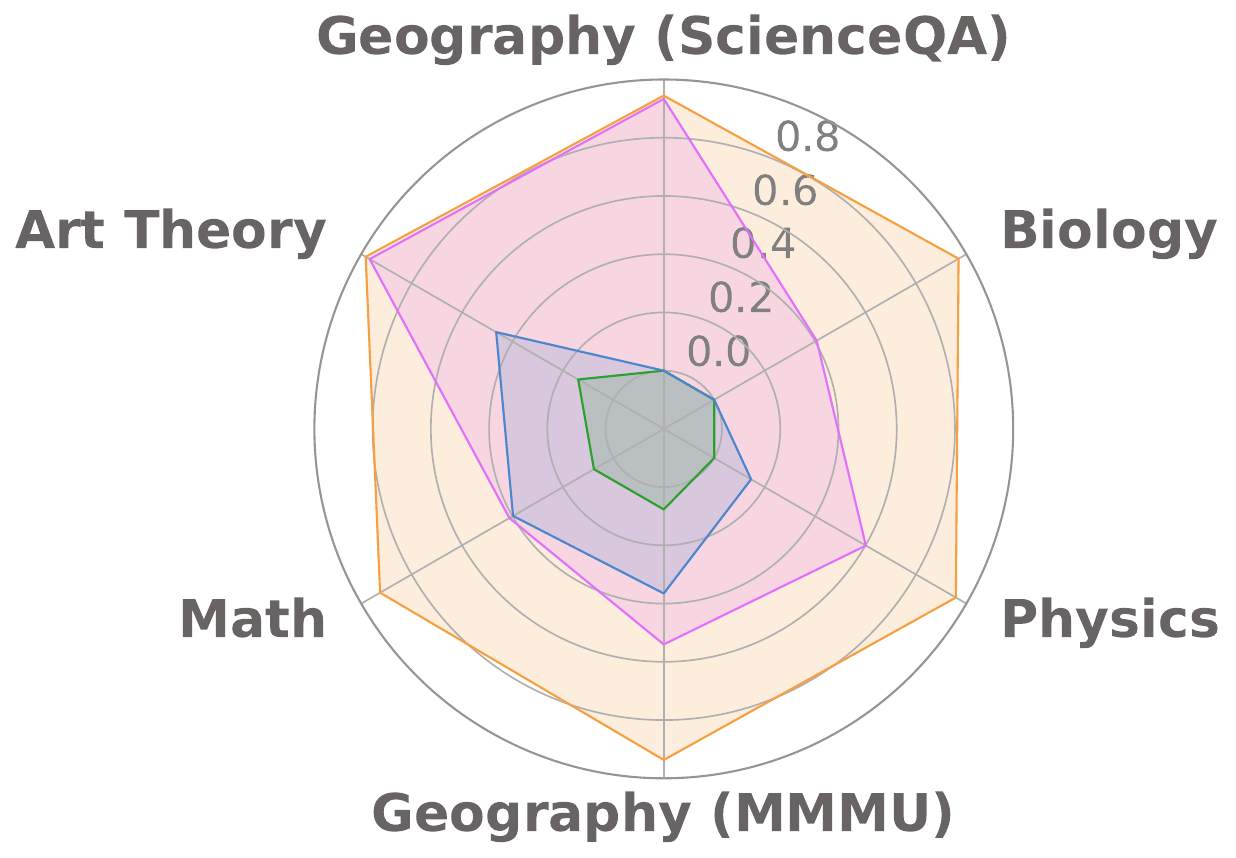}
            \caption{LLaVA-1.5-13B-hf}
        \end{subfigure}
        \newline
        
        \vspace{-10pt}
        
        \hfill
        \begin{subfigure}{0.49\textwidth}
            \centering
            \includegraphics[width=\linewidth]{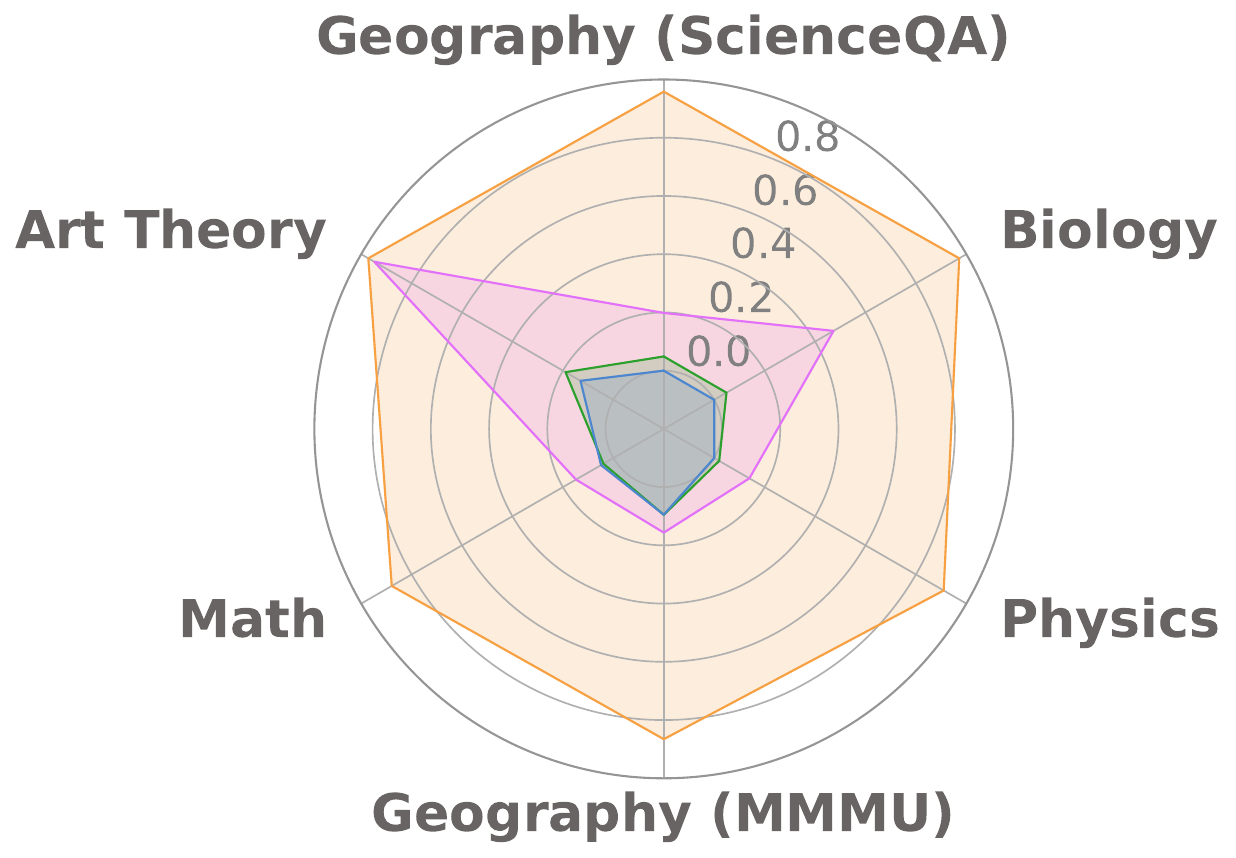}
            \caption{Idefics3}
        \end{subfigure}
        \hfill
        \begin{subfigure}{0.49\textwidth}
            \centering
            \includegraphics[width=\linewidth]{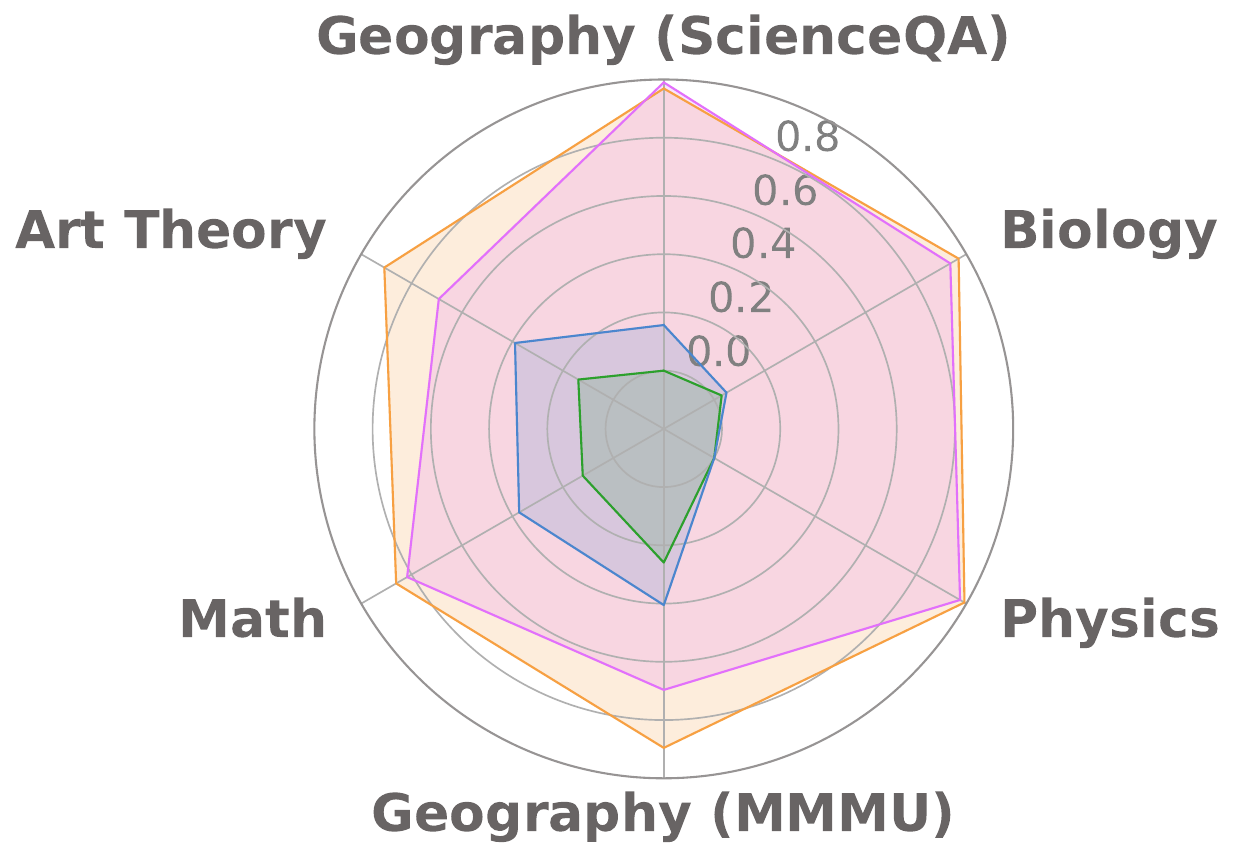}
            \caption{InstructBLIP}
        \end{subfigure}

        % \newline
        \vspace{0.05in}
        \hfill
        \begin{subfigure}{\textwidth}
            \centering
            \includegraphics[width=0.95\linewidth]{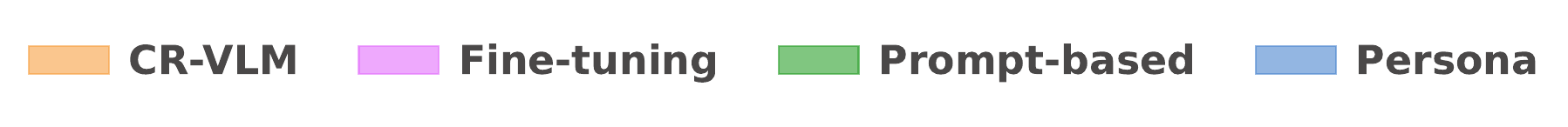}
        \end{subfigure}
    \end{minipage}
    % }
    \vspace{-10pt}
    \caption{(EQ2) MB-Score trade-off between refusal rate on out-of-scope queries and over-refusal rate on in-scope queries.
    % {\color{red} Revise: The green area becomes bigger. (start from -0.5)}
    % ~\jiaxi{1)Please revise legend ``our method'' to ``CR-VLM''; 2) start from -0.5, then we green area will not be too tiny.}
    }
    \label{fig:MB-Score}
\end{figure}

%%%%%%%%%%%%%%%%%% Activation Distribution Figures %%%%%%%%%%%%%%%%%%
\begin{figure*}[h]
    \centering
    \resizebox{0.95\textwidth}{!}{%
    \begin{minipage}{\textwidth}
        \centering
        \begin{subfigure}{0.32\textwidth}
            \centering
            \includegraphics[width=\linewidth]{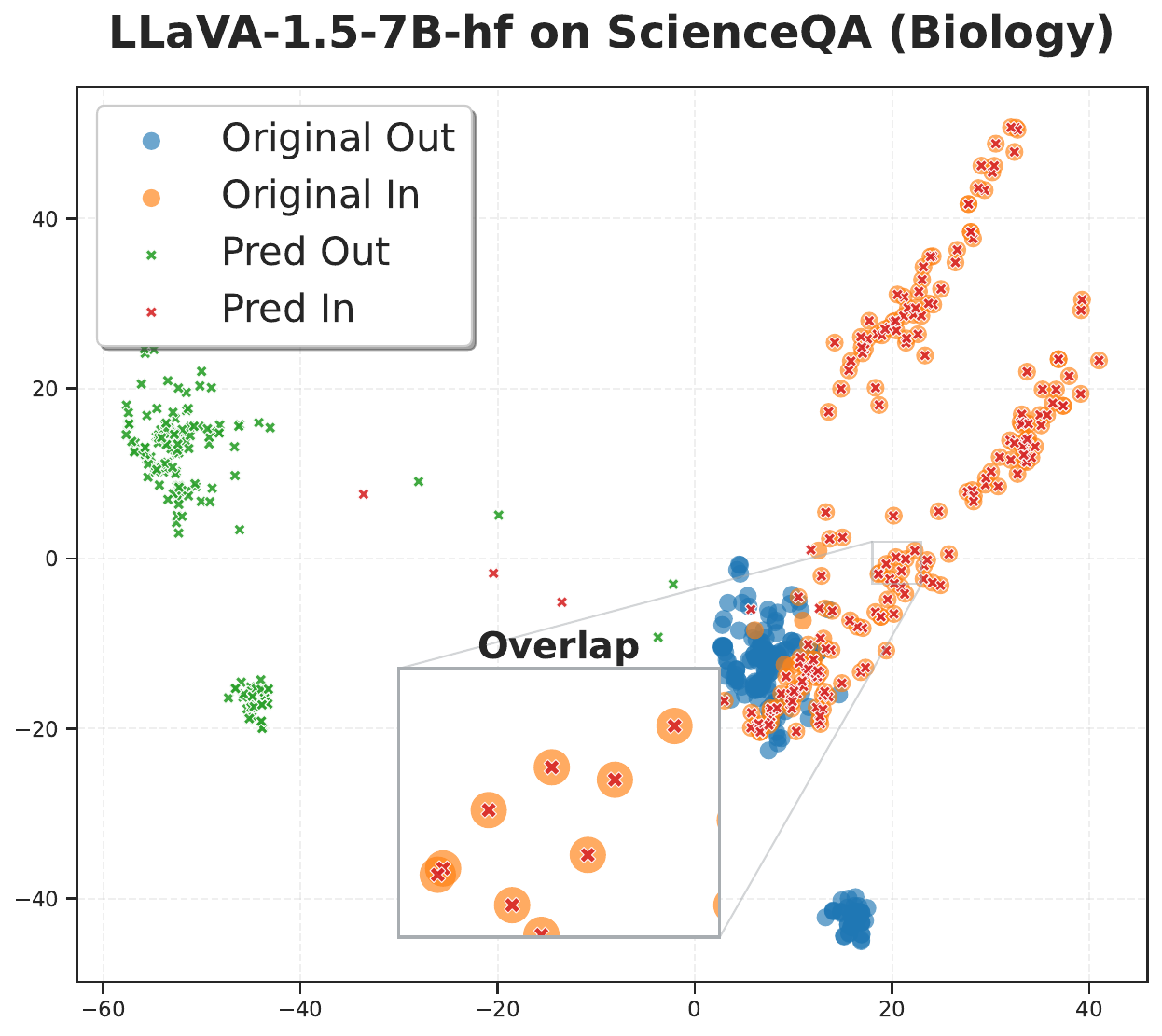}
        \end{subfigure}
        \hfill
        \begin{subfigure}{0.32\textwidth}
            \centering
            \includegraphics[width=\linewidth]{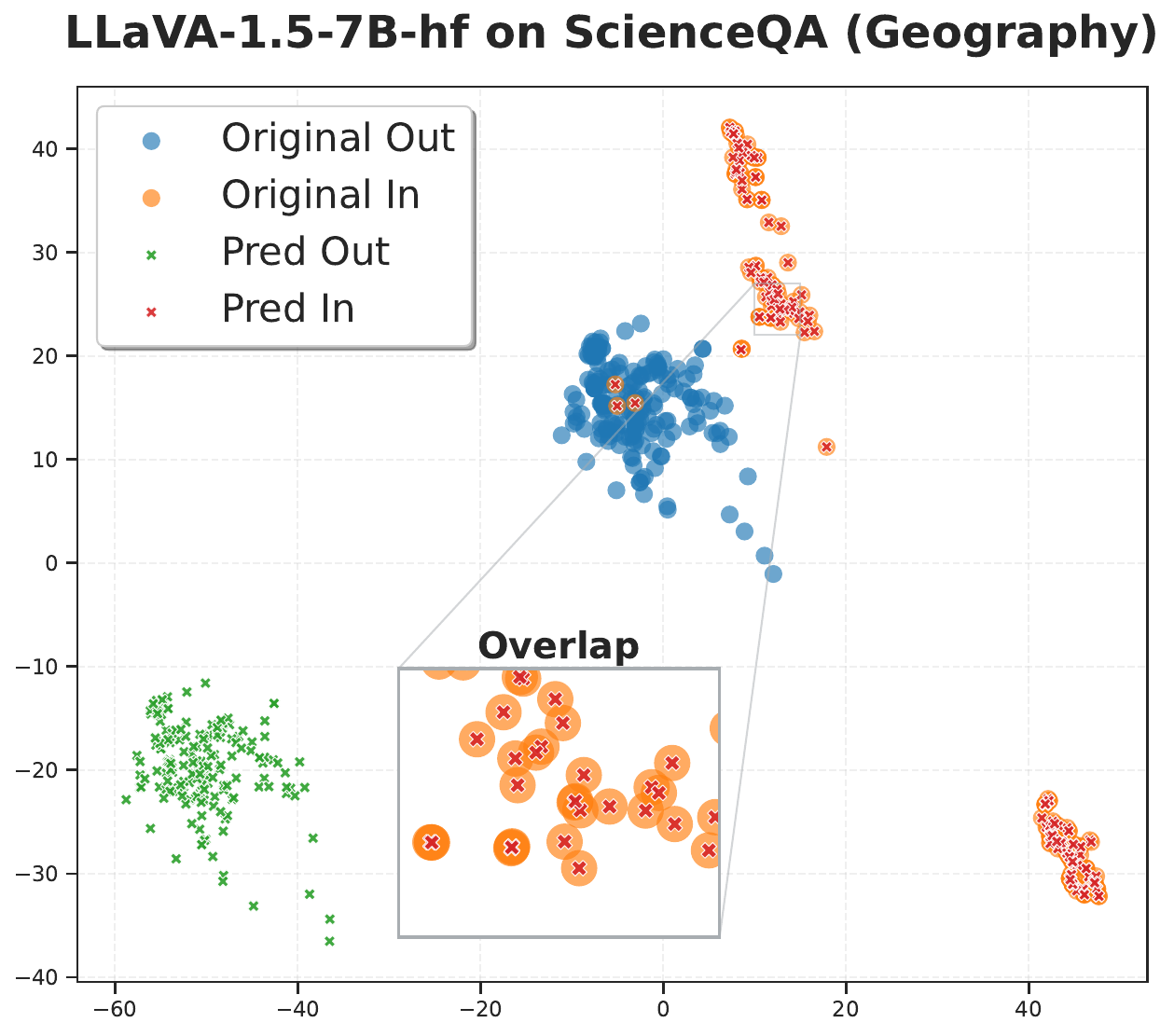}
        \end{subfigure}
        \hfill
        \begin{subfigure}{0.32\textwidth}
            \centering
            \includegraphics[width=\linewidth]{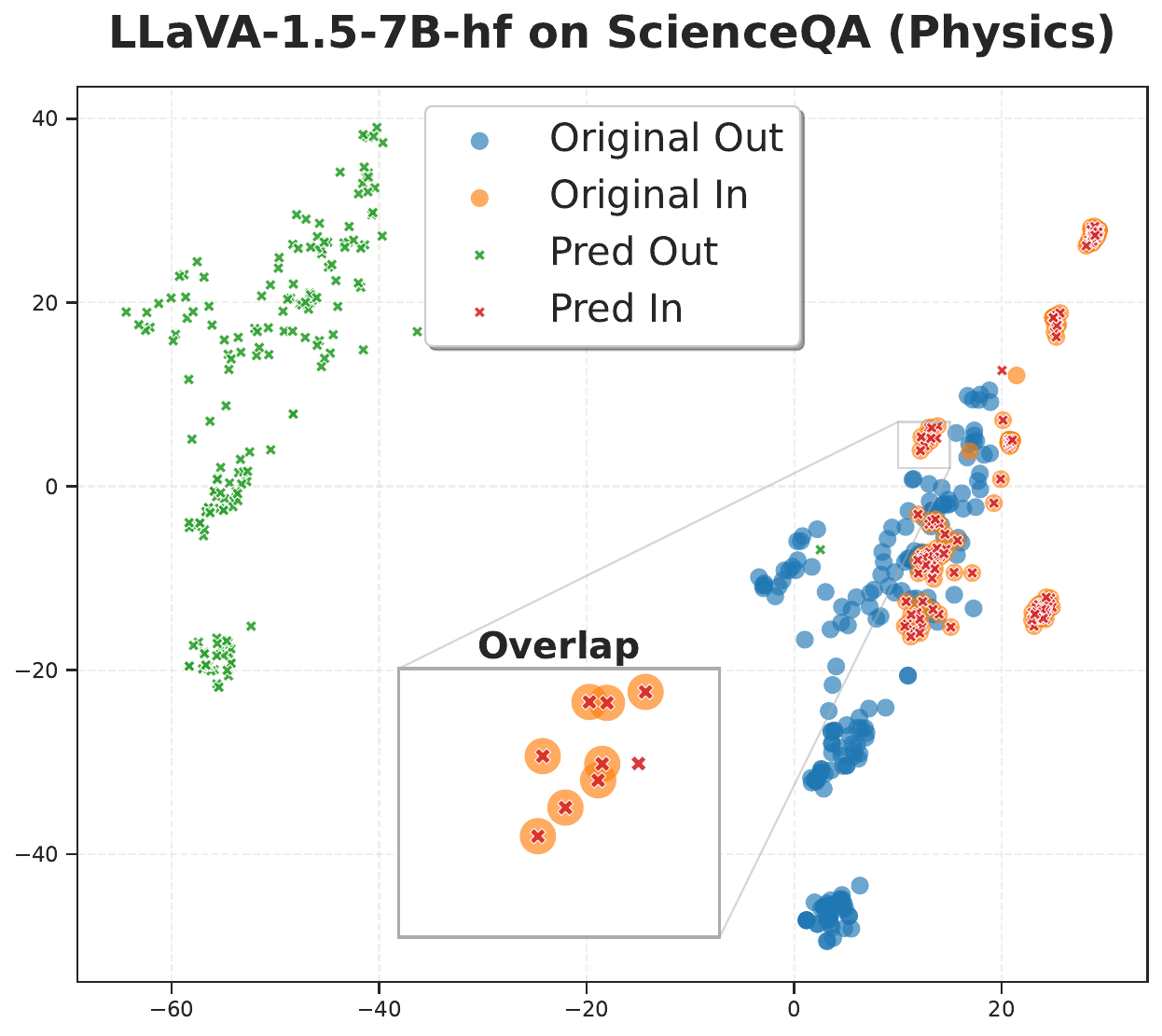}
        \end{subfigure}
        \vspace{-10pt}
        \begin{subfigure}{0.32\textwidth}
            \centering
            \includegraphics[width=\linewidth]{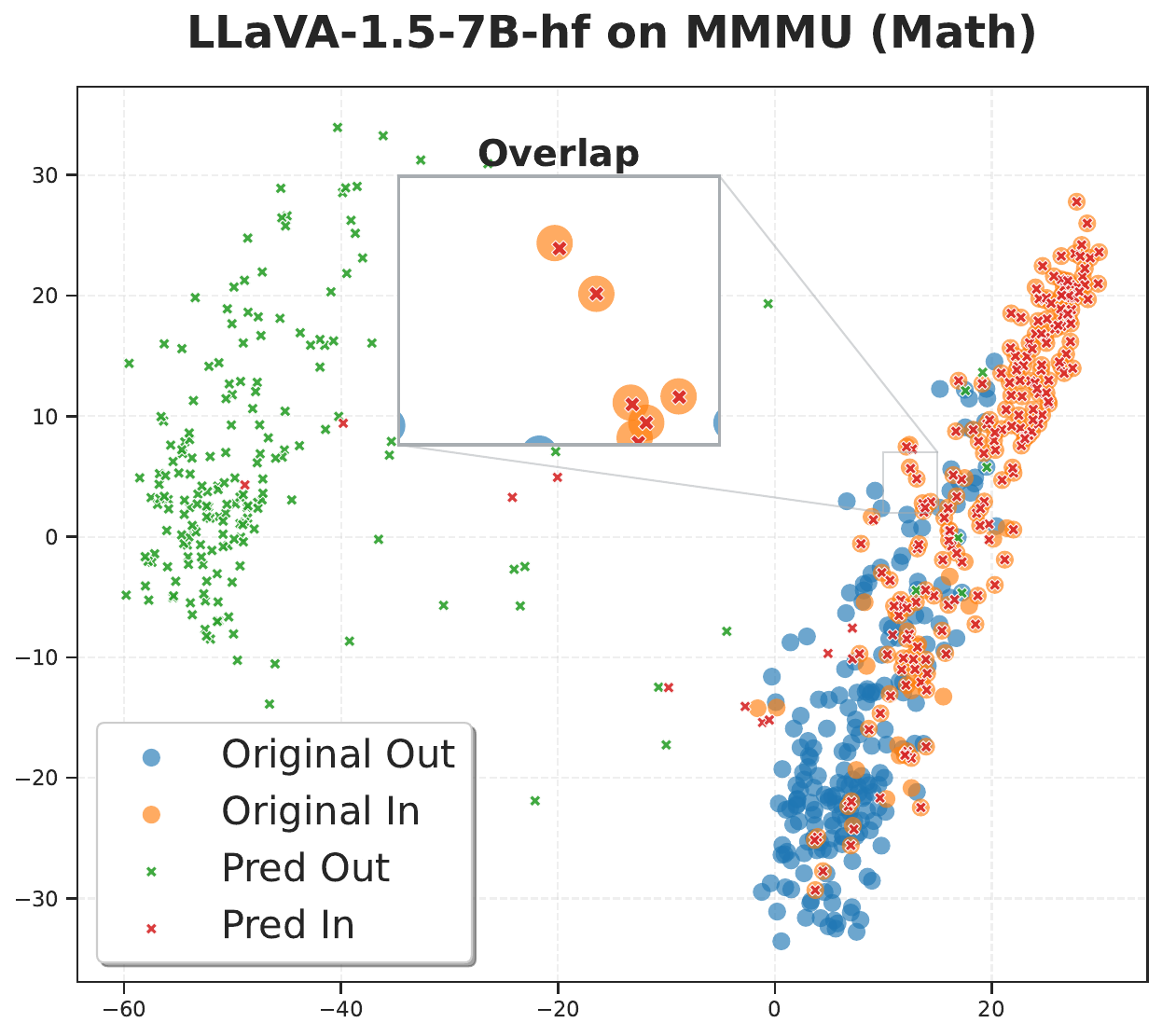}
        \end{subfigure}
        \hfill
        \begin{subfigure}{0.32\textwidth}
            \centering
            \includegraphics[width=\linewidth]{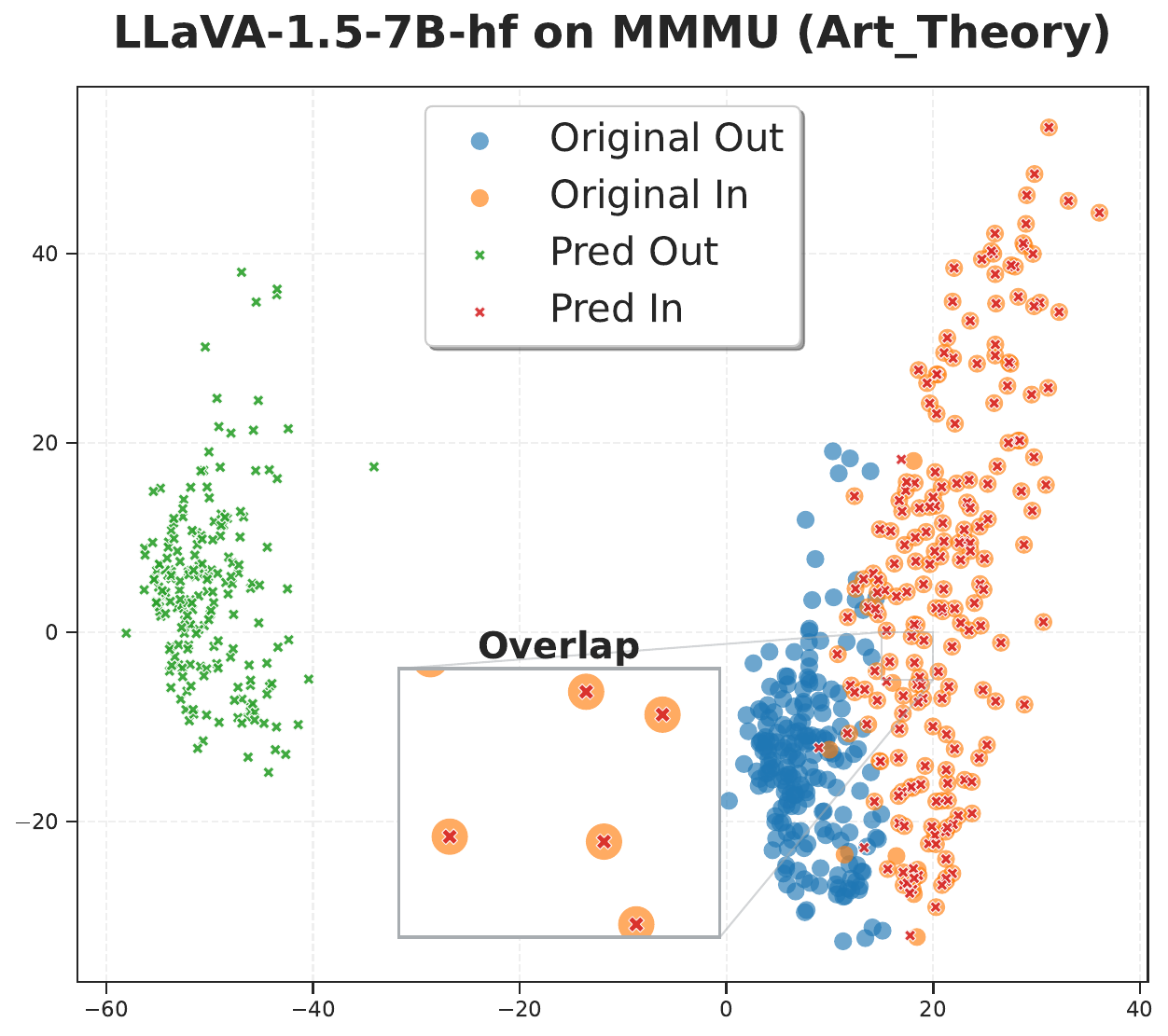}
        \end{subfigure}
        \hfill
        \begin{subfigure}{0.32\textwidth}
            \centering
            \includegraphics[width=\linewidth]{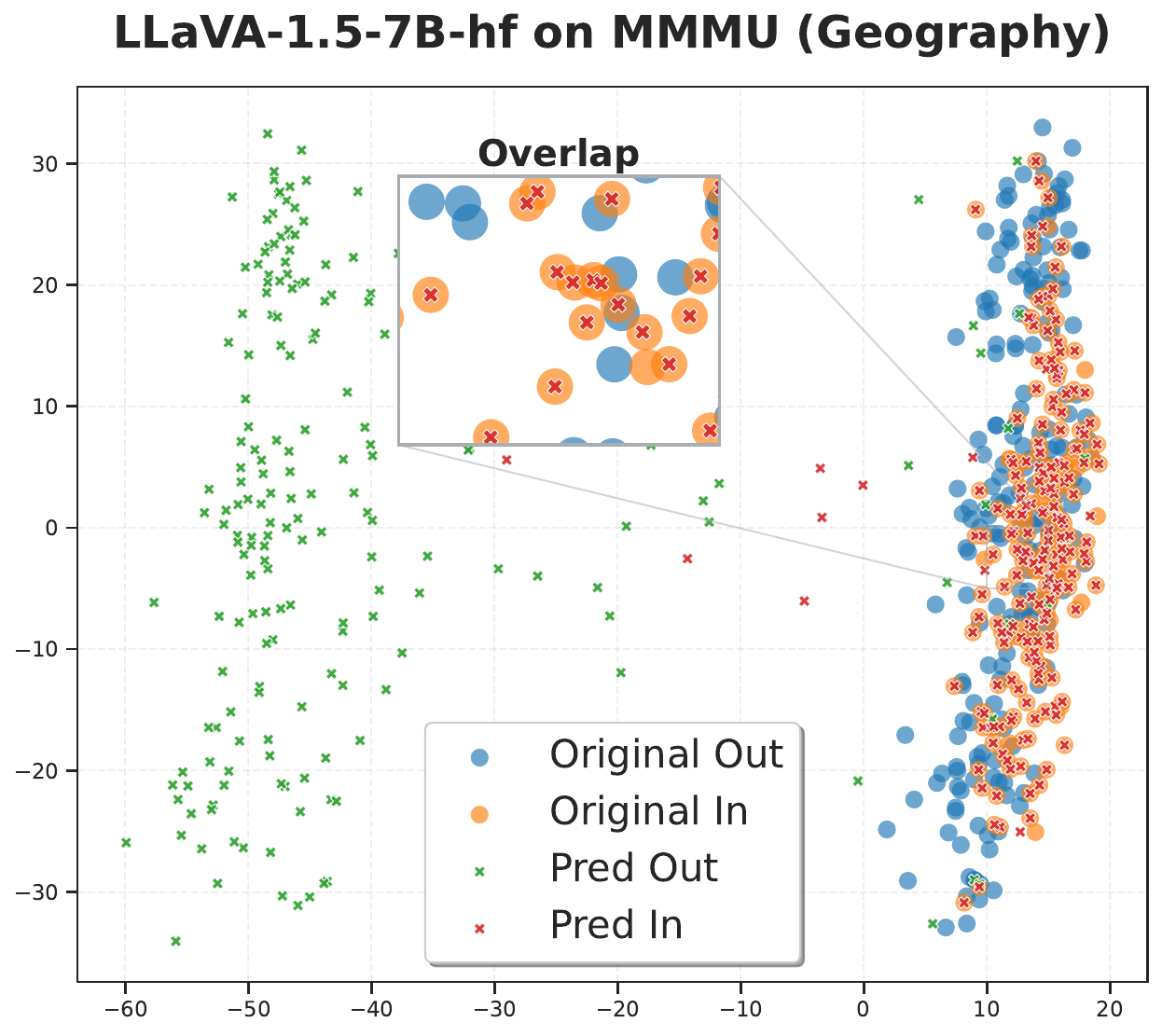}
        \end{subfigure}
    \end{minipage}
    }
    % \vspace{-5pt}
    \caption{(EQ2) Activation distribution shifts at the 25$^{th}$ layer for in-scope and out-of-scope test samples across different subjects in ScienceQA and MMMU on LLaVA-1.5-7B-hf. 
    % \yuchen{Please make the image caption consistent, either all start with capital (Biology vs. biology), also it is not clear with is original out/in, pred out/in here}~\jiaxi{We will change the figures later.}
    }
    \label{fig:RQ3}
    \vspace{-10pt}
\end{figure*}

\noindent\textit{\underline{Result Analysis.}}
As shown in Table~\ref{table:RQ1}, our method achieves higher refusal rates on out-of-scope samples than baselines in most cases across both datasets in different subjects. This demonstrates that the calibrated activations through our method are substantially more effective for inducing configurable refusal behavior in various VLMs. 
The baseline of Persona consistently underperforms across models and subjects, yielding near-zero refusal rates in some settings. This is because injecting refusal behavior solely through system prompts relies heavily on the built-in safety alignment of the model. When such alignment is weak, the induced refusal signal is insufficient, as failing to extract
strong refusal-related activations, leading to ineffective refusal behavior steering.
% Although Persona achieves relatively high refusal rates on LLaVA-1.5-7B-hf and LLaVA-1.5-13B-hf, it performs poorly on Idefics3-8B-Llama3 and InstructBLIP. This is because injecting refusal constraints via system prompts relies on the model’s inherent safety alignment. When such safety alignment is lacking, the induced refusal signal becomes too weak to extract refusal-related activations. 
While fine-tuning methods are able to induce refusal behavior to some extent,  our approach outperforms fine-tuning in 41 out of 48 experimental settings, with fine-tuning performing better in only a small fraction of cases.
This suggests that the approach becomes unreliable when only limited 200 fine-tuning data samples are available, as in our experimental setting. 
% \yuchen{explain which row is under this scenario, how limited?}
% Moreover, it is considerably more expensive\yuchen{any numbers?} to apply in practice.
Besides, prompt-based steering performs extremely poorly, often yielding near-zero refusal rates, indicating that simple prompt-based manipulation is insufficient to induce reliable configurable refusal in VLMs.

\vspace{-5pt}
\subsection{(EQ2) Over-Refusal Mitigation}
\vspace{-5pt}
Here, we evaluate whether our proposed method will not lead to over-refusal for in-scope queries while maintaining the high refusal rate for out-of-scope queries.

\noindent\textbf{\textit{Over-Refusal Rate on In-scope Queries.}}\\
\noindent\textit{\underline{Evaluation Metrics.}}
Similar to the evaluation protocol in EQ1, we use refusal rate to quantify over-refusal (\ie the proportion of in-scope queries that are false refused). In contrast to EQ1, a lower refusal rate here refers to better performance in this setting. Similarly, we evaluate this using both human evaluation and LLM-as-Judgment as well.

\vspace{-5pt}
\noindent\textit{\underline{Result Analysis.}}
As shown in Table~\ref{table:over_refusal}, our method exhibits consistently low over-refusal rates across all models and datasets within various subject domains, indicating that it effectively preserves acceptance for in-scope queries while maintaining strong refusal behavior on out-of-scope queries. In contrast, Persona in baseline methods performs poorly in some subjects, displaying significant over-refusal, which frequently rejects in-scope queries. These results demonstrate that our calibrated activation approach achieves a more reliable configurable refusal behavior.

\noindent\textbf{\textit{Trade-off between Refusal and Over-Refusal.}}\\
\noindent\textit{\underline{Evaluation Metrics.}}
To explicitly capture the trade-off between effective Refusal Rate (RR) and Over-Refusal Rate (ORR), we utilize MB-Score~\cite{simhi2025managerbench} as shown in equation (\ref{eq:MB-Score}), a composite metric that jointly considers refusal performance on out-of-scope queries and acceptance behavior on in-scope queries.
\vspace{-5pt}
\begin{equation}
\label{eq:MB-Score}
    \text{MB-Score} = \frac{2 \cdot \text{RR} \cdot (1 - \text{ORR})}
{\text{RR} + (1 - \text{ORR})}.
\vspace{-5pt}
\end{equation}
\noindent\textit{\underline{Result Analysis.}}
Figure~\ref{fig:MB-Score} summarizes the trade-off using MB-Score, which jointly reflects refusal effectiveness and over-refusal. Across all subjects of both datasets, our method consistently achieves higher MB-Scores than baselines, indicating the best performance for configurable refusal. 
% \yuchen{similar here, provide insights why they are bad and we are good, instead of just saying we are better 
In contrast, fine-tuning and Persona methods suffer from degraded MB-Scores due to excessive over-refusal for in-scope queries or ineffective refusal steering for out-of-scope queries.
These results demonstrate that our calibrated activation approach enables configurable refusal, selectively amplifying refusal only when constraints are violated, rather than globally increasing refusal strength.

\noindent\textbf{\textit{Activation Distribution.}}\\
\noindent\textit{\underline{Evaluation Metrics.}}
The goal of our configurable refusal mechanism is to reshape the model’s hidden representations to reflect the desired refusal behavior. To assess whether the calibrated activations induced by CR-VLM yield clearer separation between in-scope and out-of-scope queries compared to the original representations, we conduct a qualitative analysis of hidden-state distributions. Specifically, we apply Principal Component Analysis (PCA)~\cite{pearson1901liii} to both the original and calibrated activations of in-scope and out-of-scope queries, and visualize their projected representations in a low-dimensional space.

\vspace{-5pt}
\noindent\textit{\underline{Result Analysis.}}
% To further analyze the trade-off behavior from a representation perspective, we examine the distribution of hidden activations after calibration. 
As shown in Figure~\ref{fig:RQ3}, the calibrated activations for in-scope queries remain close to the original in-scope representations, indicating that our method largely preserves the acceptance behavior for in-scope queries. 
In contrast, out-of-scope samples present a substantial shift in the representation space: their calibrated activations are pushed away from both the original out-of-scope and in-scope region and form a compact cluster associated with refusal behavior. 
We further observe similar distribution patterns across more various models in the Appendix~\ref{appendix:activation_distribution}.
\vspace{-5pt}
\subsection{(EQ3) Ablation Studies}
\vspace{-5pt}
In this section, we conduct ablation studies on LLaVA-1.5-7B-hf to analyze the contribution of individual components in CR-VLM. 
% The results for other models are shown in the appendix.

\begin{figure}
    \centering
    \includegraphics[width=\linewidth]{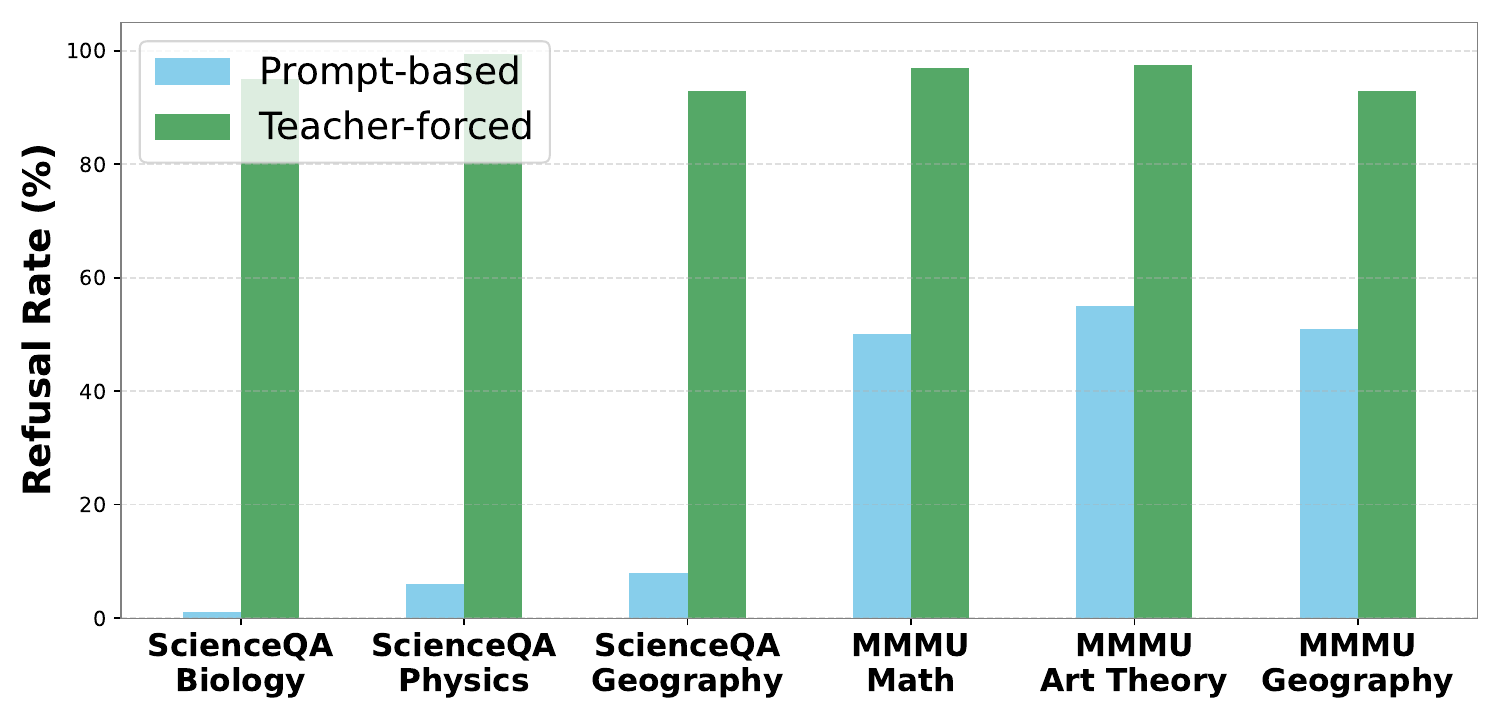}
    \vspace{-15pt}
    \caption{(EQ3) Ablation study of \emph{teacher-forced method} for activation extraction compared with prompt-based method.}
    \label{fig:teacher_ablation}
    \vspace{-15pt}
\end{figure}

\noindent\textbf{\textit{Efficacy of Teacher-Forced Activation Extraction.}}
To evaluate whether teacher-forced activation extraction yields stronger refusal signals, we compare it with prompt-based activation extraction across different datasets and subjects. As shown in Figure~\ref{fig:teacher_ablation}, teacher-forced activation extraction consistently achieves higher refusal rates on out-of-scope samples, verifying its effectiveness.

\noindent\textbf{\textit{Over-Refusal Mitigation by Orthogonality.}}
\label{appendix:orth}
We conduct an ablation study by comparing designing a loss function with or without the orthogonality regularization term $\mathcal{L}_{\text{ortho}}$. The results reported in Table~\ref{tab:orth} demonstrate that severe over-refusal on in-scope queries across all subjects without the orthogonality loss term. This indicates that without explicit decoupling, the learned intervention directions for in-scope and out-of-scope samples collapse into a shared global shift, leading to over-refusal. 
In contrast, incorporating the orthogonality constraint effectively suppresses over-refusal while preserving strong refusal behavior for out-of-scope inputs, validating its critical role in enabling configurable refusal.

% \begin{table*}
%     \centering
%     \captionsetup{width=0.9\linewidth}
%     \caption{Ablation results of the loss function with (w/) and without (w/o) the orthogonality term across different datasets, evaluated on both in-scope and out-of-scope test samples.}
%     \resizebox{.7\linewidth}{!}{%
%     \begin{tabular}{cccccccc}
%     \toprule
%         \multicolumn{2}{c}{\multirow{2}{*}{\textbf{Dataset (\%)}}} & \multicolumn{3}{c}{\textbf{ScienceQA}} & \multicolumn{3}{c}{\textbf{MMMU}} \\
%         \cmidrule(lr){3-8}
%         & & \textbf{Biology} & \textbf{Physics} & \textbf{Geography} & \textbf{Math} & \textbf{Art Theory} & \textbf{Geography} \\
%         \cmidrule(lr){1-8}
%         \multirow{2}{*}{In-scope} & w & 0.0 & 0.0 & 0.0 & 9.5 & 1.5 & 4.0 \\
%         & w/o & 97.0 \up{97.0} & 89.5 \up{89.5} & 100.0 \up{100} & 80.5 \up{71} & 91.0 \up{89.5} & 96.0 \up{92}\\
%         \cmidrule(lr){1-8}
%         \multirow{2}{*}{Out-of-scope} & w & 95.0 & 99.5 & 93.0 & 97.0 & 97.5 & 93.0 \\
%         & w/o & 95.0 \same & 98.0 \down{1.5} & 93.0 \same & 98.5 \up{1.5} & 97.0 \down{0.5}& 97.5 \up{4.5} \\
%     \bottomrule
%     \end{tabular}
%     }
%     \label{tab:orth}
% \end{table*}
\begin{table}
    \centering
    \captionsetup{width=0.95\linewidth}
    \caption{(EQ3) Ablation results of the loss function with (w) and without (w/o) the \emph{orthogonality} term across different datasets, conducted on LLaVA-1.5-7B-hf for both in-scope and out-of-scope test samples.}
    \vspace{-5pt}
    \resizebox{0.95\linewidth}{!}
    {%
    \begin{tabular}{cccccc}
    \toprule
        \multirow{2}{*}{\textbf{Dataset}} & \multirow{2}{*}{\textbf{Subject}} &
        \multicolumn{2}{c}{\textbf{In-scope (\%)}} &
        \multicolumn{2}{c}{\textbf{Out-of-scope (\%)}} \\
        \cmidrule(lr){3-6}
        & & \textbf{w} & \textbf{w/o} & \textbf{w} & \textbf{w/o} \\
    \midrule
        \multirow{3}{*}{ScienceQA}
        & Biology   & 0.0 & 97.0 \up{97.0} & 95.0 & 95.0 \same \\
        & Physics   & 0.0 & 89.5 \up{89.5} & 99.5 & 98.0 \down{1.5} \\
        & Geography & 0.0 & 100.0 \up{100} & 93.0 & 93.0 \same \\
    \midrule
        \multirow{3}{*}{MMMU}
        & Math      & 9.5 & 80.5 \up{71} & 97.0 & 98.5 \up{1.5} \\
        & Art Theory& 1.5 & 91.0 \up{89.5} & 97.5 & 97.0 \down{0.5} \\
        & Geography & 4.0 & 96.0 \up{92} & 93.0 & 97.5 \up{4.5} \\
    \bottomrule
    \end{tabular}
    }
    \label{tab:orth}
\end{table}
% \input{tables/vision}

\iffalse
\noindent\textbf{\textit{Efficacy of Gate Mechanism.}}
xxx
\fi

\begin{figure}
    \centering
    \begin{subfigure}{\linewidth}
        \centering
        \includegraphics[width=0.8\linewidth]{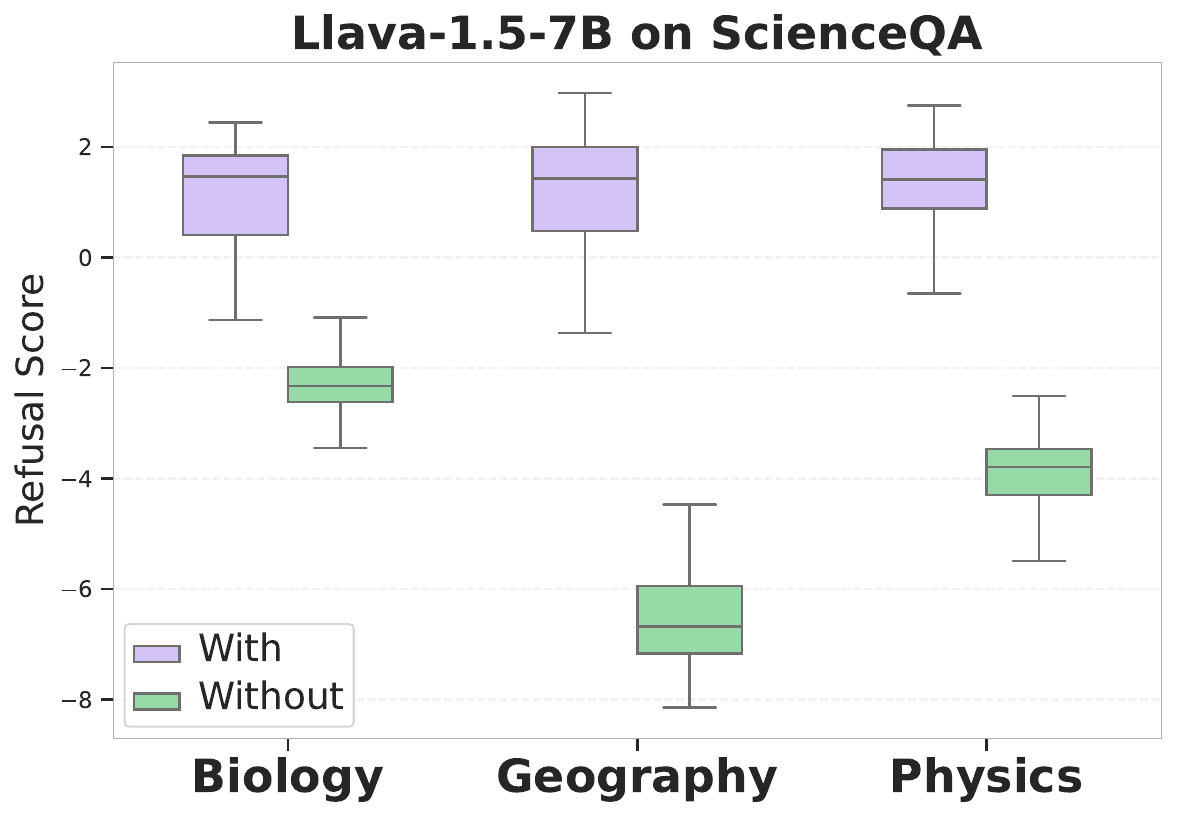}
    \end{subfigure}
    % \vspace{10pt}
    % \begin{subfigure}{\linewidth}
    %     \centering
    %     \includegraphics[width=0.8\linewidth]{figures/vision_ablation/vision_ablation_grouped_box_llava-1.5-7b-hf_MMMU.pdf}
    % \end{subfigure}
    \vspace{-20pt}
    \caption{(EQ3) Ablation study of \emph{vision loss} conducted on LLaVA-1.5-7B-hf on ScienceQA.}
    \label{fig:vision_ablation}
    \vspace{-10pt}
\end{figure}

\noindent\textbf{\textit{Efficacy of Vision Loss.}}
\label{sec:vision}
To examine whether vision loss encourages the model to pay more attention to visual cues during refusal, we conduct an ablation study by inputting the image with blank textual content. 
% This setting isolates the contribution of visual information to the predicted refusal behavior by excluding textual influence.
This setting focuses on the contribution of visual information to the predicted refusal behavior by excluding textual influence.
Following~\cite{arditi2024refusal,wang2024surgical}, we adopt the refusal score as evluation metrics here and the calculation details of the refusal score are introduced in the appendix~\ref{app:refusal_score_calculation}. As shown in Figure~\ref{fig:vision_ablation}, removing vision loss leads to a notable drop in refusal scores across different subjects. These results on ScienceQA dataset indicate that vision loss effectively enhances visual grounding in the refusal process, rather than relying solely on text-dominated signals. The additional results on MMMU are shown in the appendix~\ref{appendix:vision}.

% To evaluate the effectiveness of the proposed vision loss, we conduct an ablation study comparing models trained with and without this loss item. Our goal is to test whether the predicted refusal-related representations preserve visual evidence, rather than being dominated by textual cues. Concretely, we compute the cosine similarity between the predicted representation produced from multimodal inputs and a vision-conditioned reference representation, and report the averaged similarity across different subjects. A higher similarity score indicates a stronger influence of visual cues on the predicted representations.
% The results represented in Table~\ref{appendix:vision} demonstrate that the vision loss acts as an effective regularizer, prompting greater reliance on visual cues rather than text-dominated decision patterns.

\vspace{-5pt}
\section{Conclusion}
\vspace{-5pt}
In this paper, we firstly introduce CR-VLM, a configurable refusal approach for VLMs based on activation steering, enabling models to refuse out-of-scope queries while preserving acceptance for in-constraint ones. Through teacher-forced extraction, gate mechanism, and vision-aware calibration, our method achieves robust refusal alignment across different models and datasets, with low over-refusal.


% \clearpage
\newpage

%%%%%%%%%%%%%%%%%%%% Impact Statements %%%%%%%%%%%%%%%
\section*{Impact Statement}
This paper presents work whose goal is to achieve configurable refusal for VLMs. There are many potential societal consequences of our work, none which we feel must be specifically highlighted here.

\bibliography{references}
\bibliographystyle{icml2026}
% \balance

%%%%%%%%%%%%%%%%%%%%%%%%%%%%%%%%%%%%%%%%%%%%%%%%%%%%%%%%%%%%%%%%%%%%%%%%%%%%%%%
%%%%%%%%%%%%%%%%%%%%%%%%%%%%%%%%%%%%%%%%%%%%%%%%%%%%%%%%%%%%%%%%%%%%%%%%%%%%%%%
% APPENDIX
%%%%%%%%%%%%%%%%%%%%%%%%%%%%%%%%%%%%%%%%%%%%%%%%%%%%%%%%%%%%%%%%%%%%%%%%%%%%%%%
%%%%%%%%%%%%%%%%%%%%%%%%%%%%%%%%%%%%%%%%%%%%%%%%%%%%%%%%%%%%%%%%%%%%%%%%%%%%%%%

\newpage
\appendix
\onecolumn
\section{Case Study}
\label{appendix:case_study}
\begin{figure*}[h]
    \centering
    \begin{subfigure}{0.49\textwidth}
        \centering
        \includegraphics[width=\linewidth]{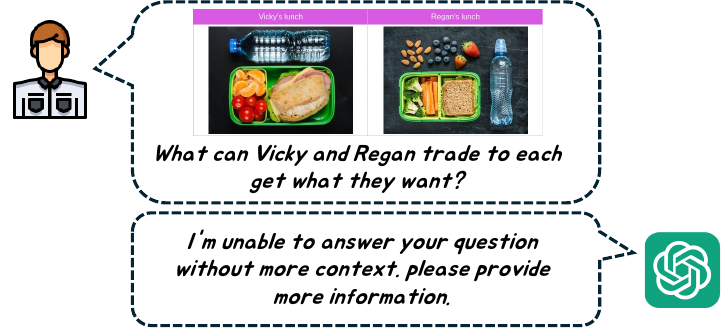}
        \caption{Example 1}
        \label{fig:a}
    \end{subfigure}
    \hfill
    \begin{subfigure}{0.49\textwidth}
        \centering
        \includegraphics[width=\linewidth]{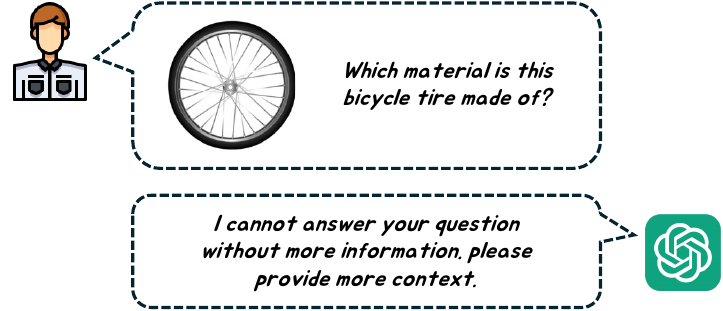}
        \caption{Example 2}
        \label{fig:b}
    \end{subfigure}

    \caption{Case Study for Out-of-scope Test Samples.}
    \label{appendix:case_study_fig}
\end{figure*}

\begin{figure*}[h]
    \centering
    \begin{subfigure}{0.49\textwidth}
        \centering
        \includegraphics[width=\linewidth]{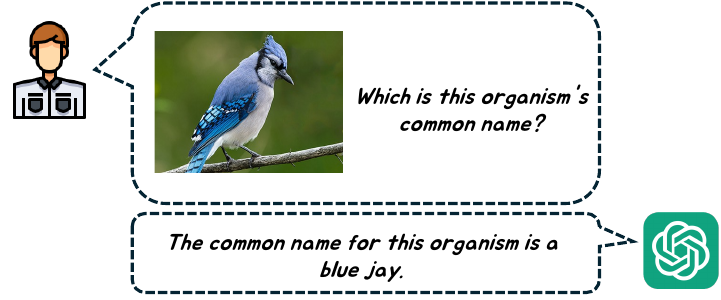}
        \caption{Example 1}
        \label{fig:a}
    \end{subfigure}
    \hfill
    \begin{subfigure}{0.49\textwidth}
        \centering
        \includegraphics[width=\linewidth]{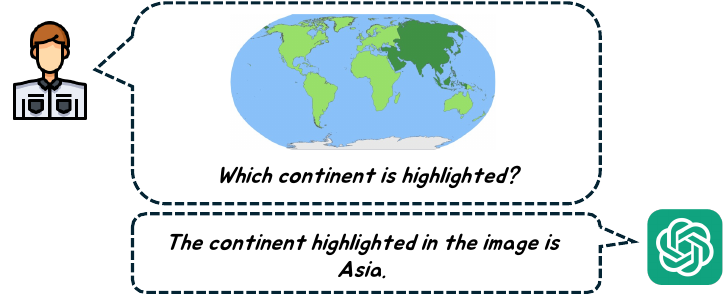}
        \caption{Example 2}
        \label{fig:b}
    \end{subfigure}

    \caption{Case Study for In-scope Test Samples.}
    \label{appendix:In_scope_case_study_fig}
\end{figure*}

Figure~\ref{appendix:case_study_fig} and Figure~\ref{appendix:In_scope_case_study_fig} provide qualitative examples illustrating the behavior of our configurable refusal mechanism for out-of-scope queries and in-scope queries respectively. The model correctly refuses out-of-scope queries while preserving normal responses for in-scope inputs, demonstrating configurable refusal rather than global refusal enforcement.

\begin{figure}[h]
    \centering
    % \resizebox{\textwidth}{!}{%
    \begin{minipage}{\linewidth}
        \centering
        \begin{subfigure}{0.23\textwidth}
            \centering
            \includegraphics[width=\linewidth]{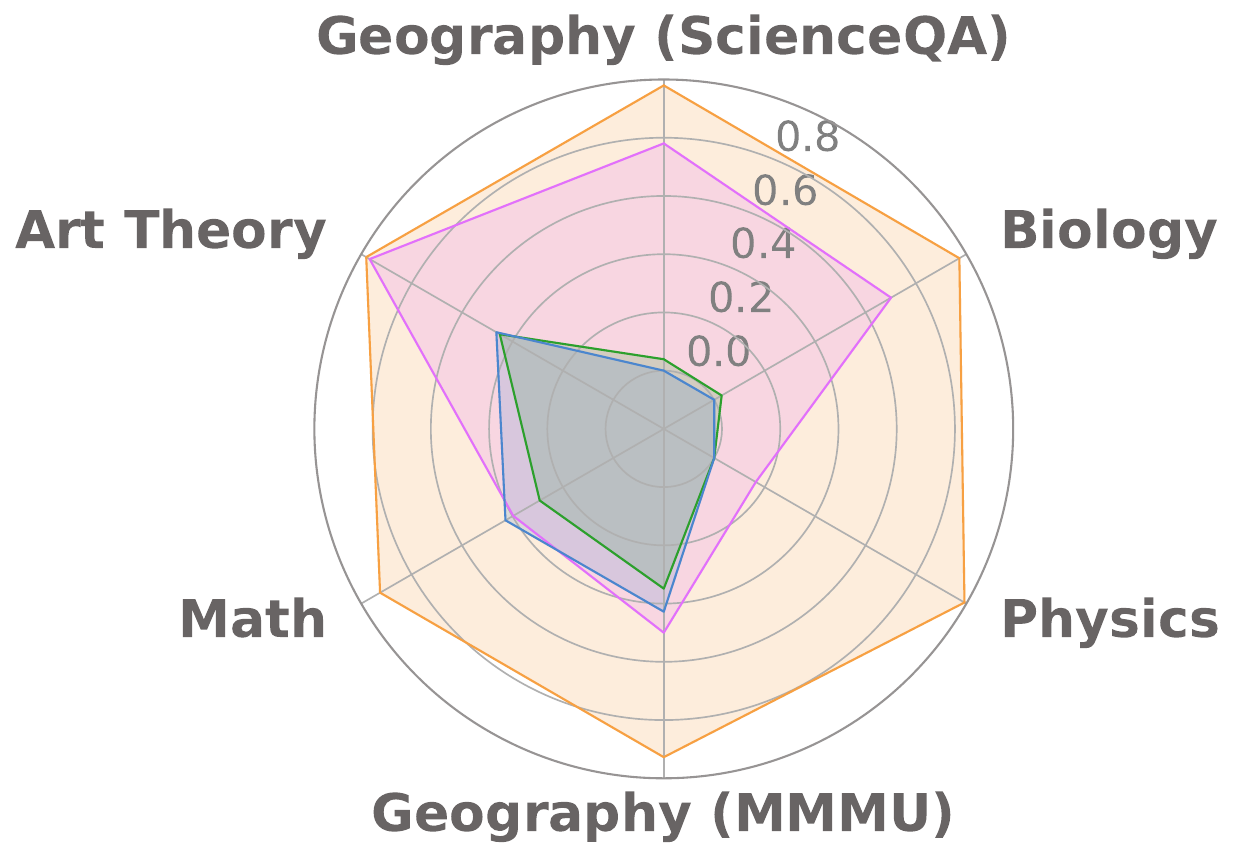}
        \end{subfigure}
        \hfill
        \begin{subfigure}{0.23\textwidth}
            \centering
            \includegraphics[width=\linewidth]{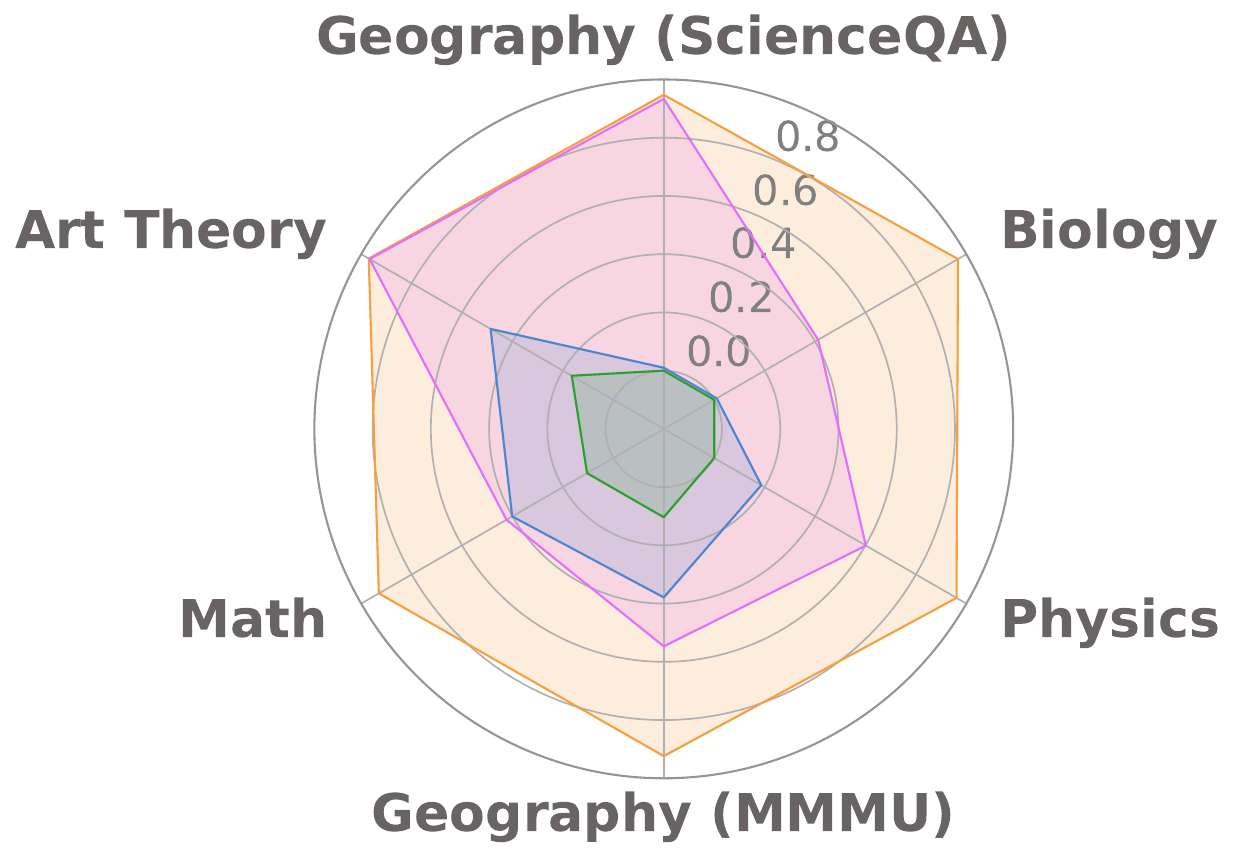}
        \end{subfigure}        
        \hfill
        \begin{subfigure}{0.23\textwidth}
            \centering
            \includegraphics[width=\linewidth]{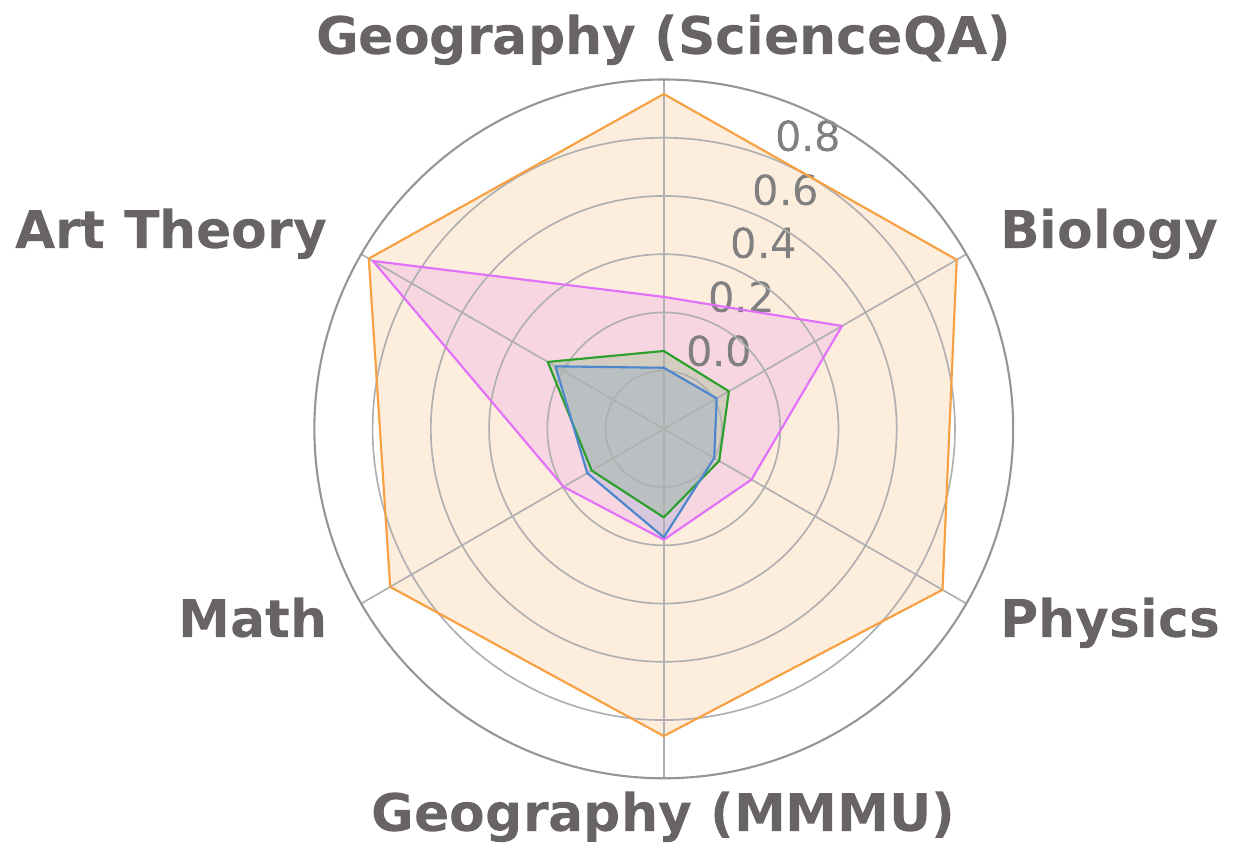}
        \end{subfigure}
        \hfill
        \begin{subfigure}{0.23\textwidth}
            \centering
            \includegraphics[width=\linewidth]{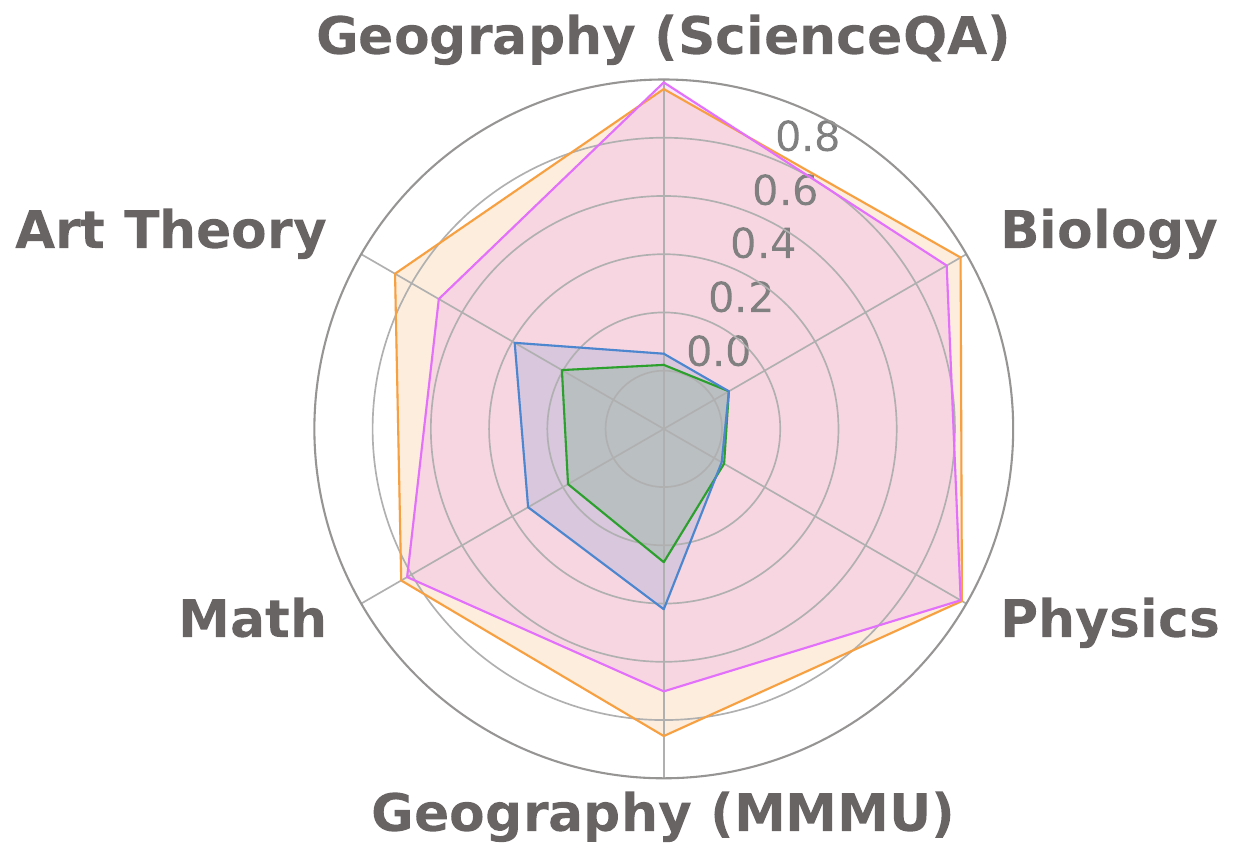}
        \end{subfigure}

        % \newline
        \vspace{0.05in}
        \hfill
        \begin{subfigure}{\textwidth}
            \centering
            \includegraphics[width=0.5\linewidth]{figures/MB_Score/legend_only.pdf}
        \end{subfigure}
        
    \end{minipage}
    % }
    \caption{(EQ2) MB-Score trade-off between refusal rate and over-refusal rate by LLM-as-Judgment.}
    \label{fig:appendix_MB_score}
\end{figure}

\begin{figure*}
    \centering
    \resizebox{0.95\textwidth}{!}{%
    \begin{minipage}{\textwidth}
        \centering
        \begin{subfigure}{0.32\textwidth}
            \centering
            \includegraphics[width=\linewidth]{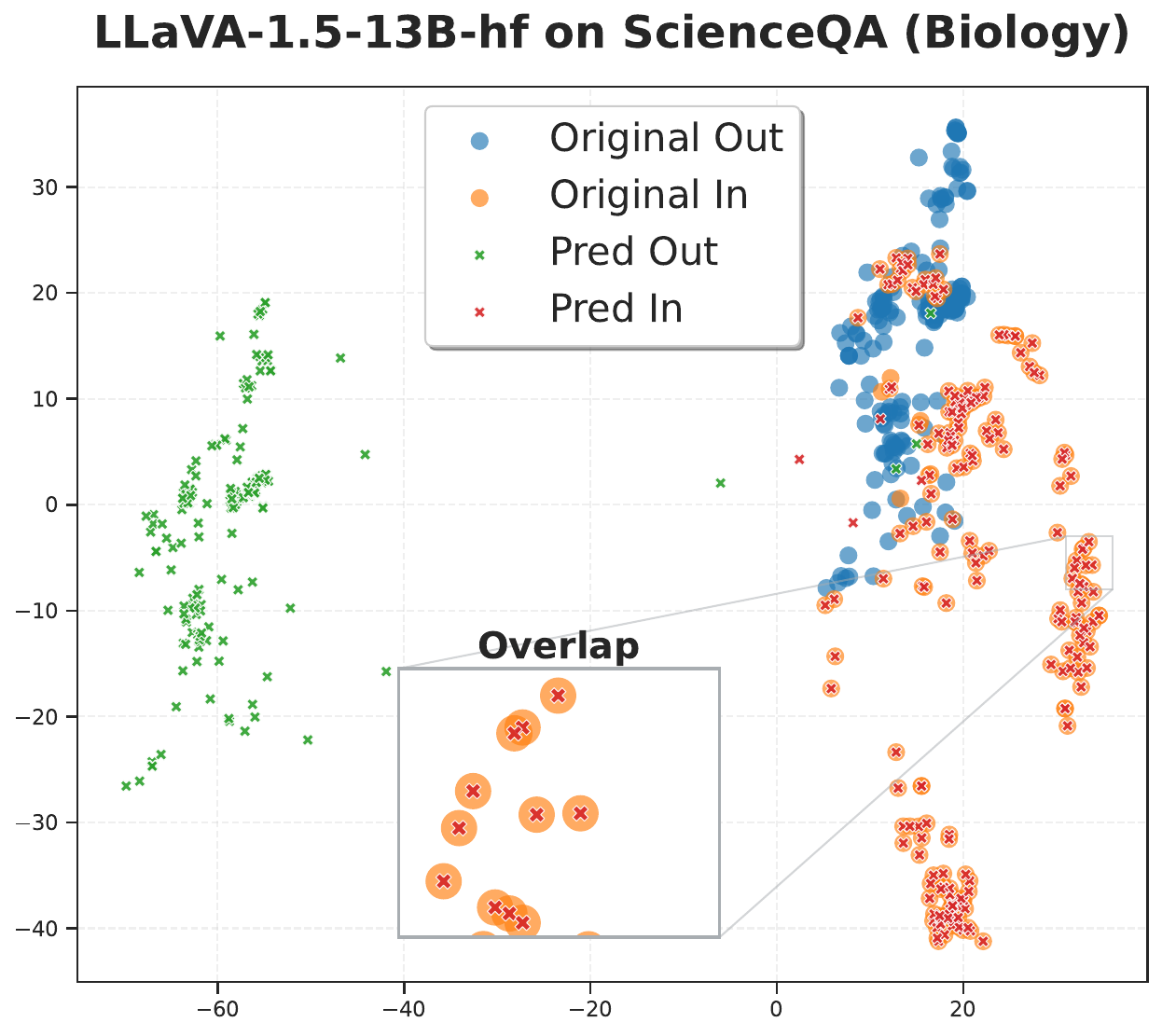}
        \end{subfigure}
        \hfill
        \begin{subfigure}{0.32\textwidth}
            \centering
            \includegraphics[width=\linewidth]{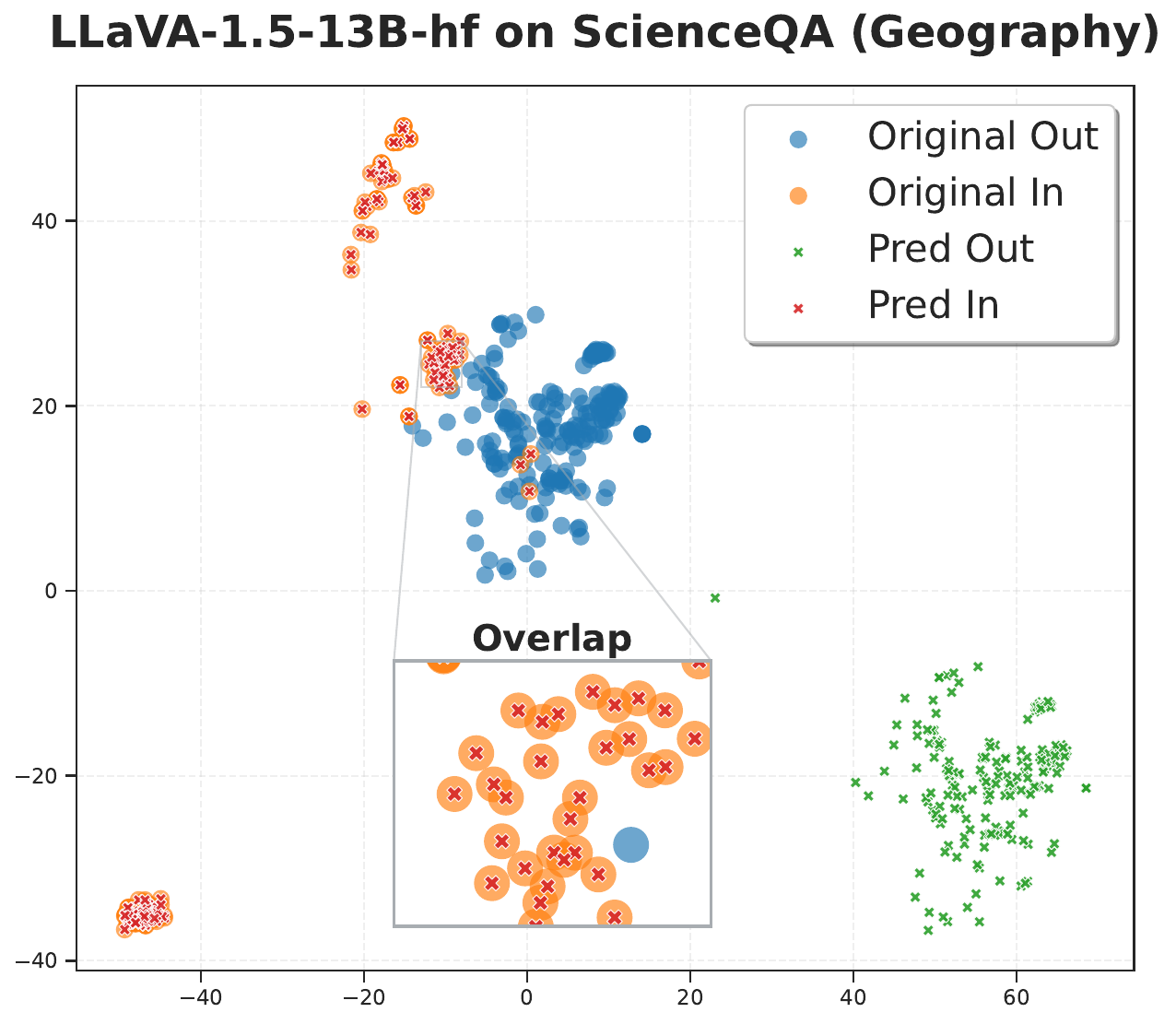}
        \end{subfigure}
        \hfill
        \begin{subfigure}{0.32\textwidth}
            \centering
            \includegraphics[width=\linewidth]{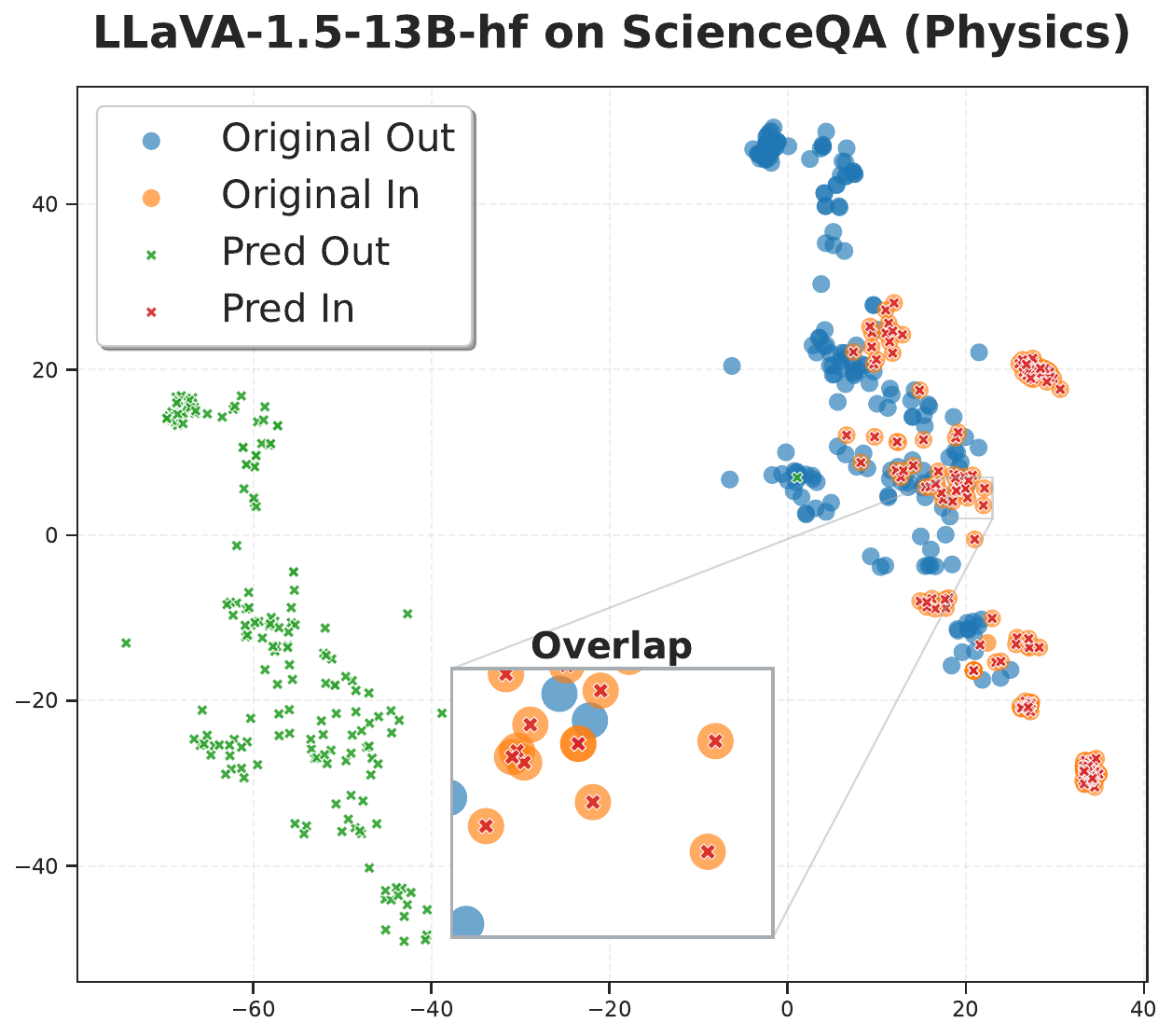}
        \end{subfigure}

        \vspace{0.25cm}

        \begin{subfigure}{0.32\textwidth}
            \centering
            \includegraphics[width=\linewidth]{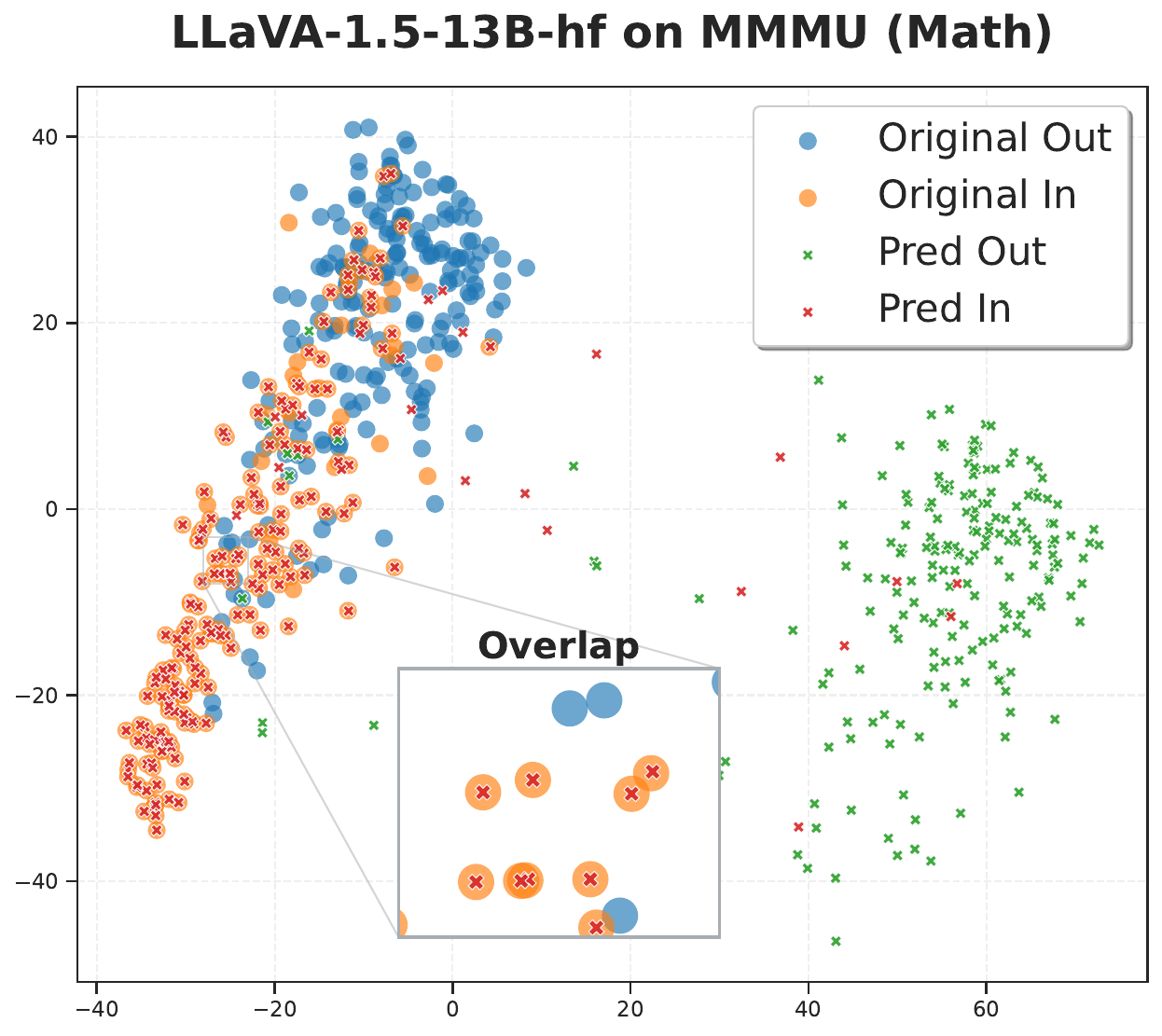}
        \end{subfigure}
        \hfill
        \begin{subfigure}{0.32\textwidth}
            \centering
            \includegraphics[width=\linewidth]{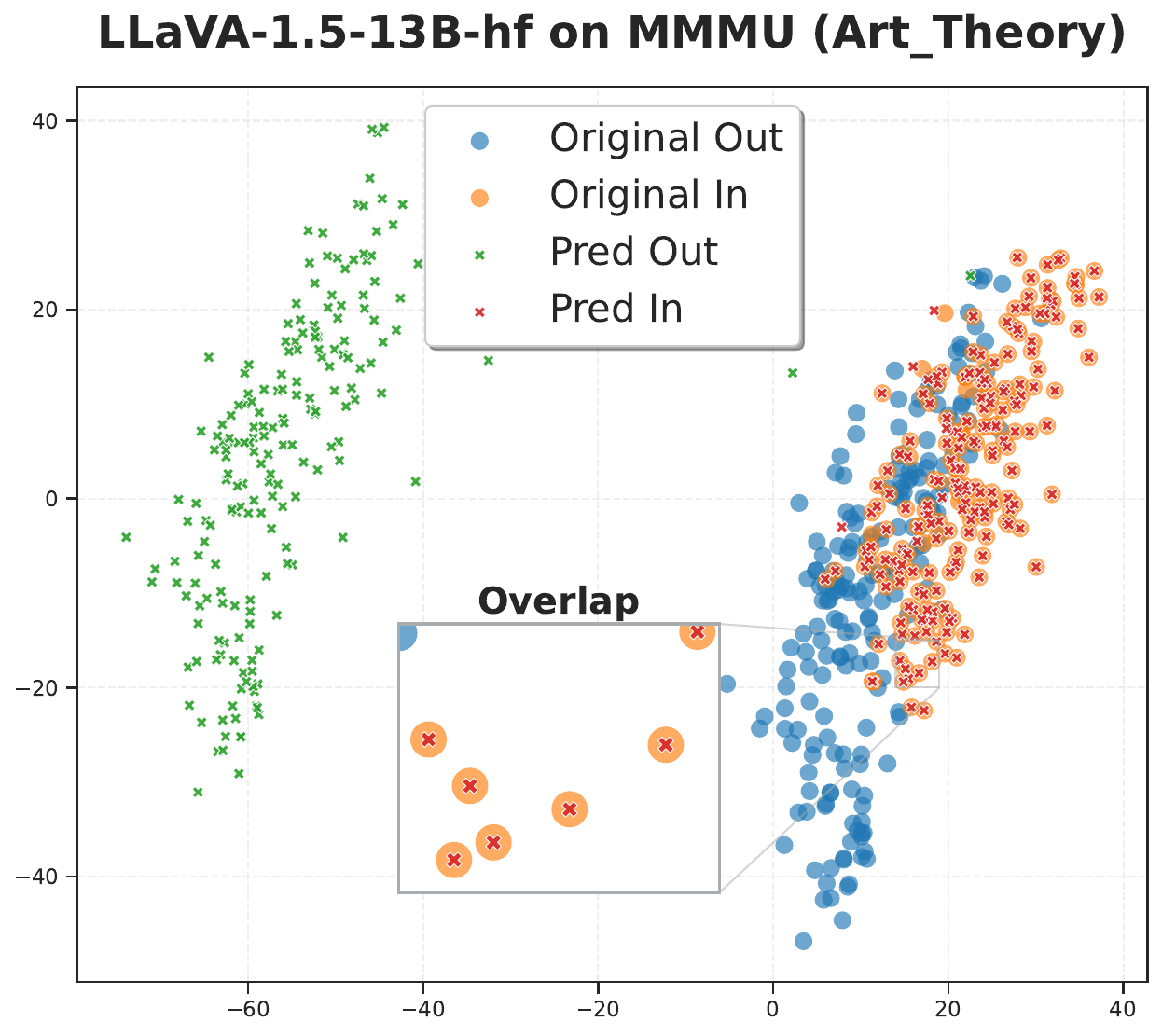}
        \end{subfigure}
        \hfill
        \begin{subfigure}{0.32\textwidth}
            \centering
            \includegraphics[width=\linewidth]{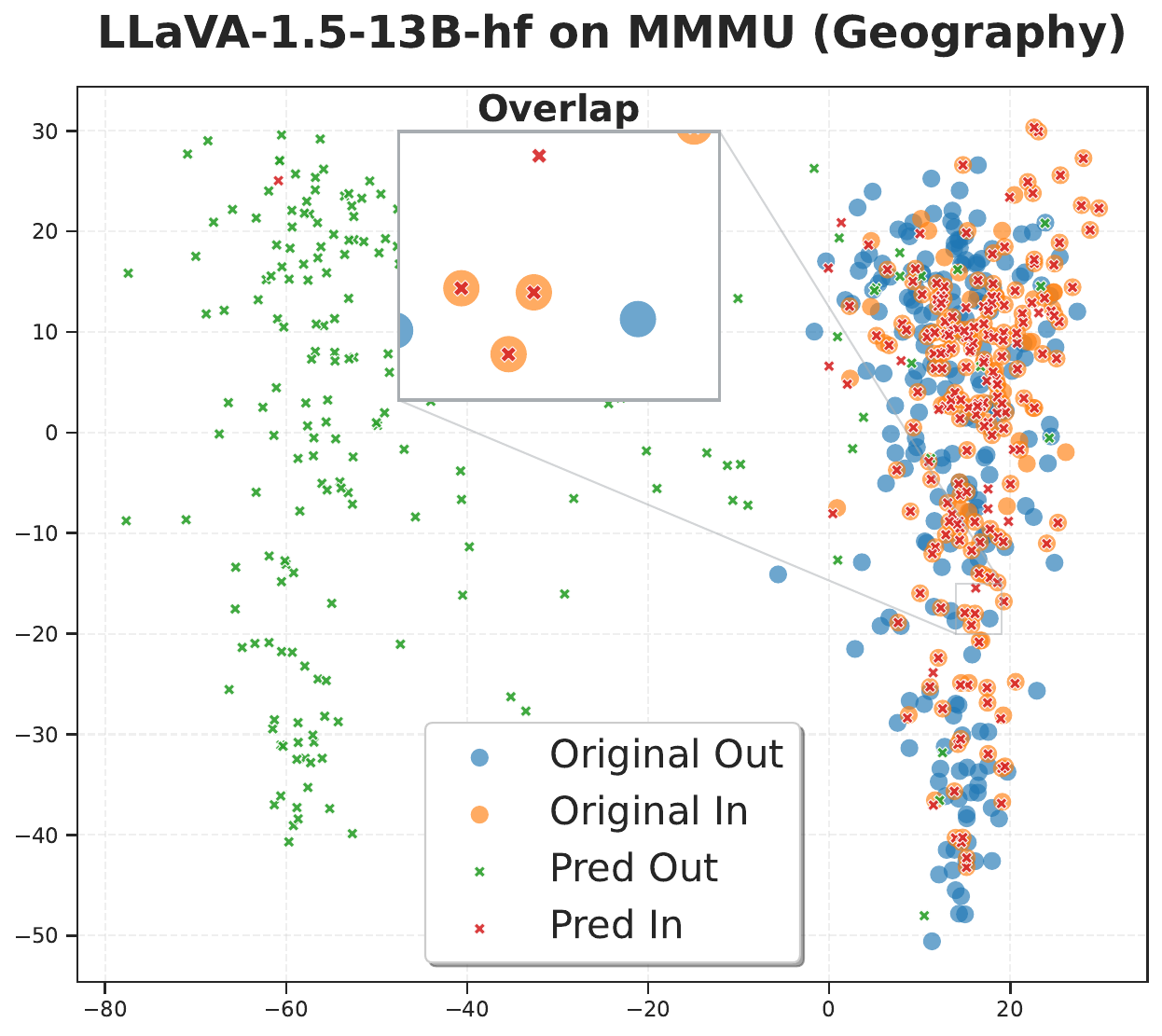}
        \end{subfigure}
        
    \end{minipage}
    }
    \caption{(EQ2) Activation distribution shifts at the 25$^{th}$ layer for in-scope and out-of-scope test samples across different subjects in ScienceQA and MMMU on LLaVA-1.5-13B-hf.}
    \label{fig:RQ3_LLava_13b}
\end{figure*}
\section{(EQ2) Additional Experiments by MB-Score}
To complement the human evaluation results reported in the main paper, we further analyze MB-Score using an LLM-as-Judgment by DeepSeek-V3~\cite{liu2024deepseek} as shown in Figure~\ref{fig:appendix_MB_score}. 
Consistent with human evaluation, our method consistently achieves higher MB-Scores than baselines across different models and subjects, demonstrating a more balanced trade-off between refusal effectiveness and over-refusal.

% \FloatBarrier
\begin{figure*}
    \centering
    \resizebox{0.95\textwidth}{!}{%
    \begin{minipage}{\textwidth}
        \centering
        \begin{subfigure}{0.32\textwidth}
            \centering
            \includegraphics[width=\linewidth]{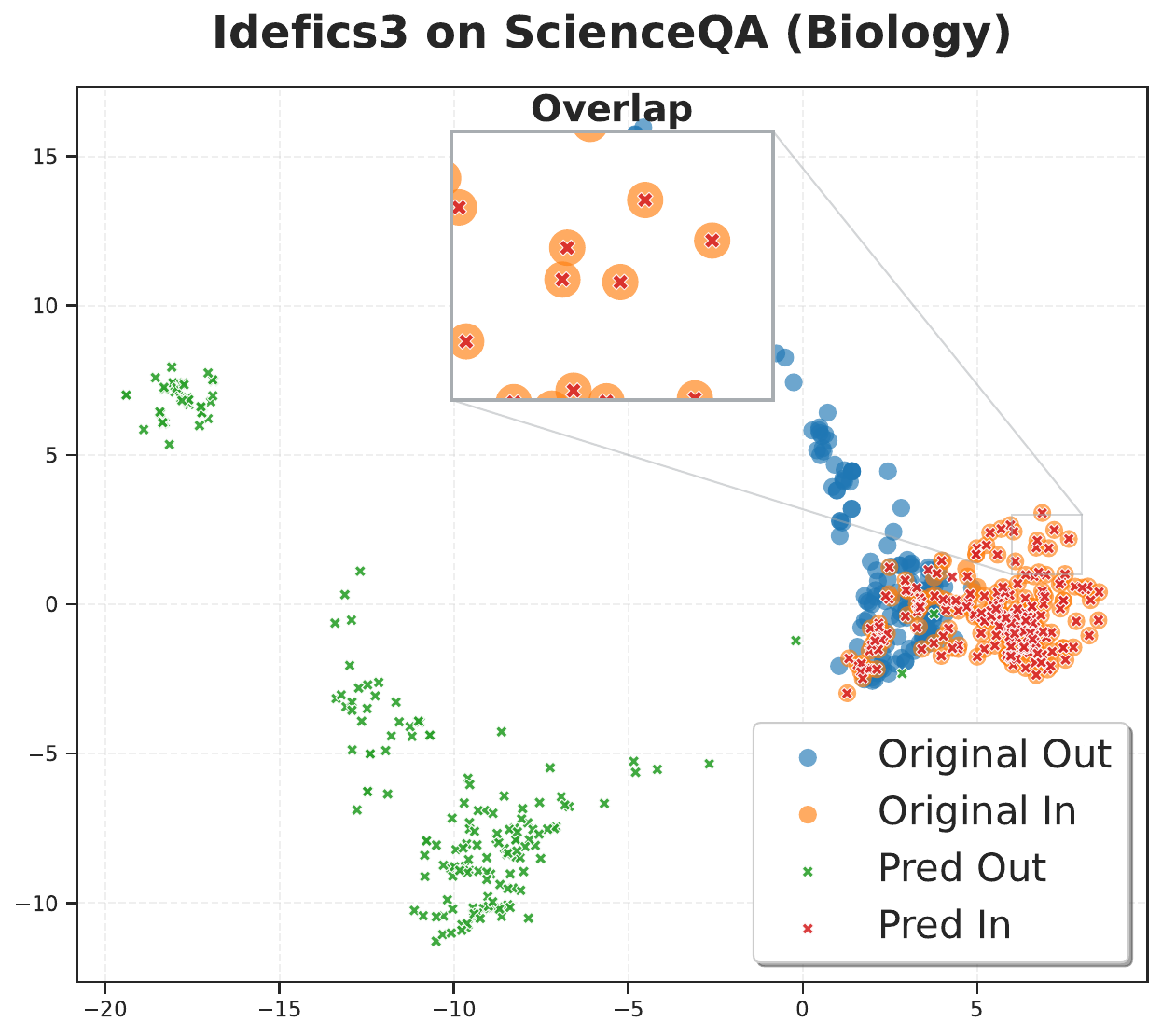}
        \end{subfigure}
        \hfill
        \begin{subfigure}{0.32\textwidth}
            \centering
            \includegraphics[width=\linewidth]{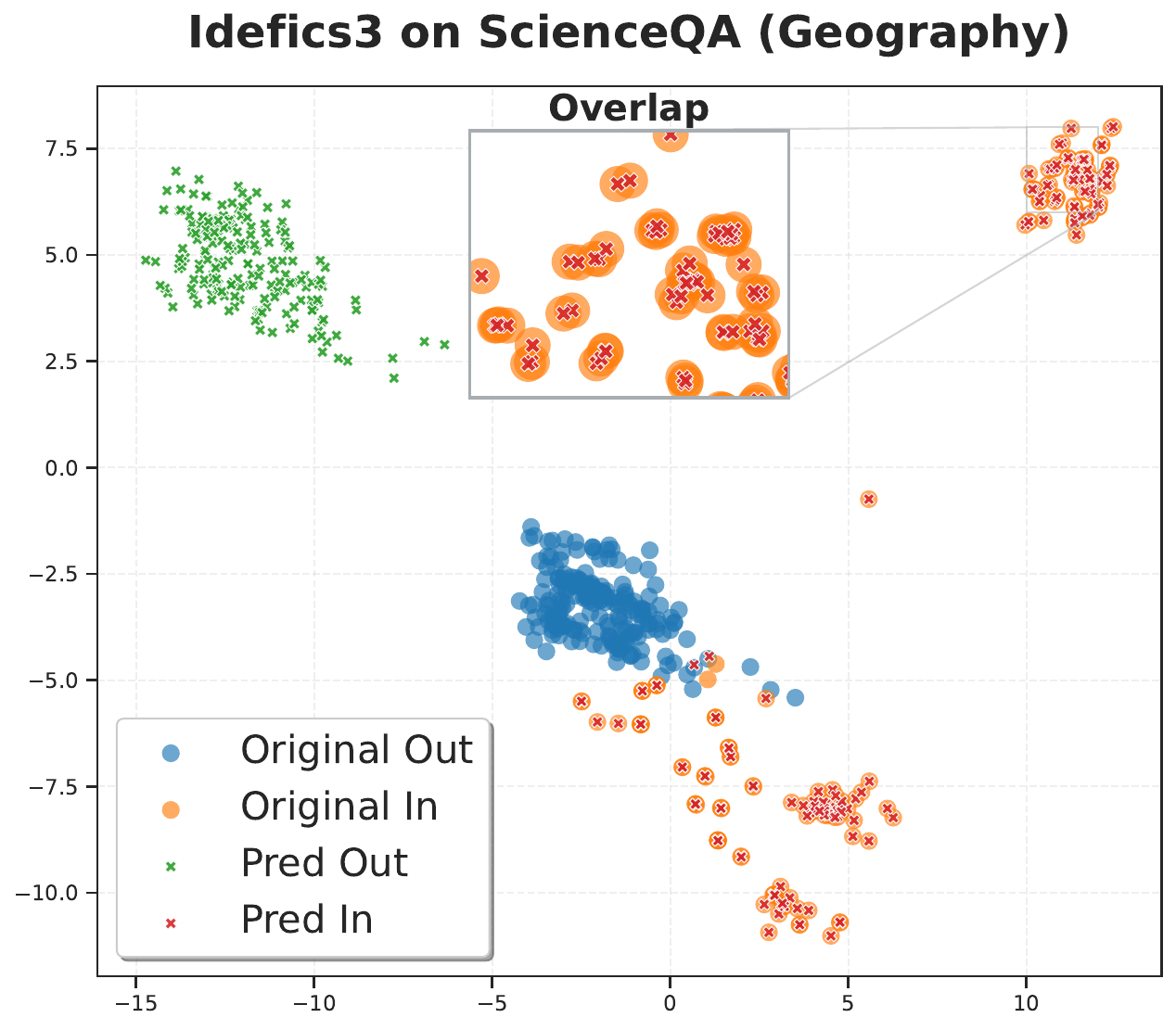}
        \end{subfigure}
        \hfill
        \begin{subfigure}{0.31\textwidth}
            \centering
            \includegraphics[width=\linewidth]{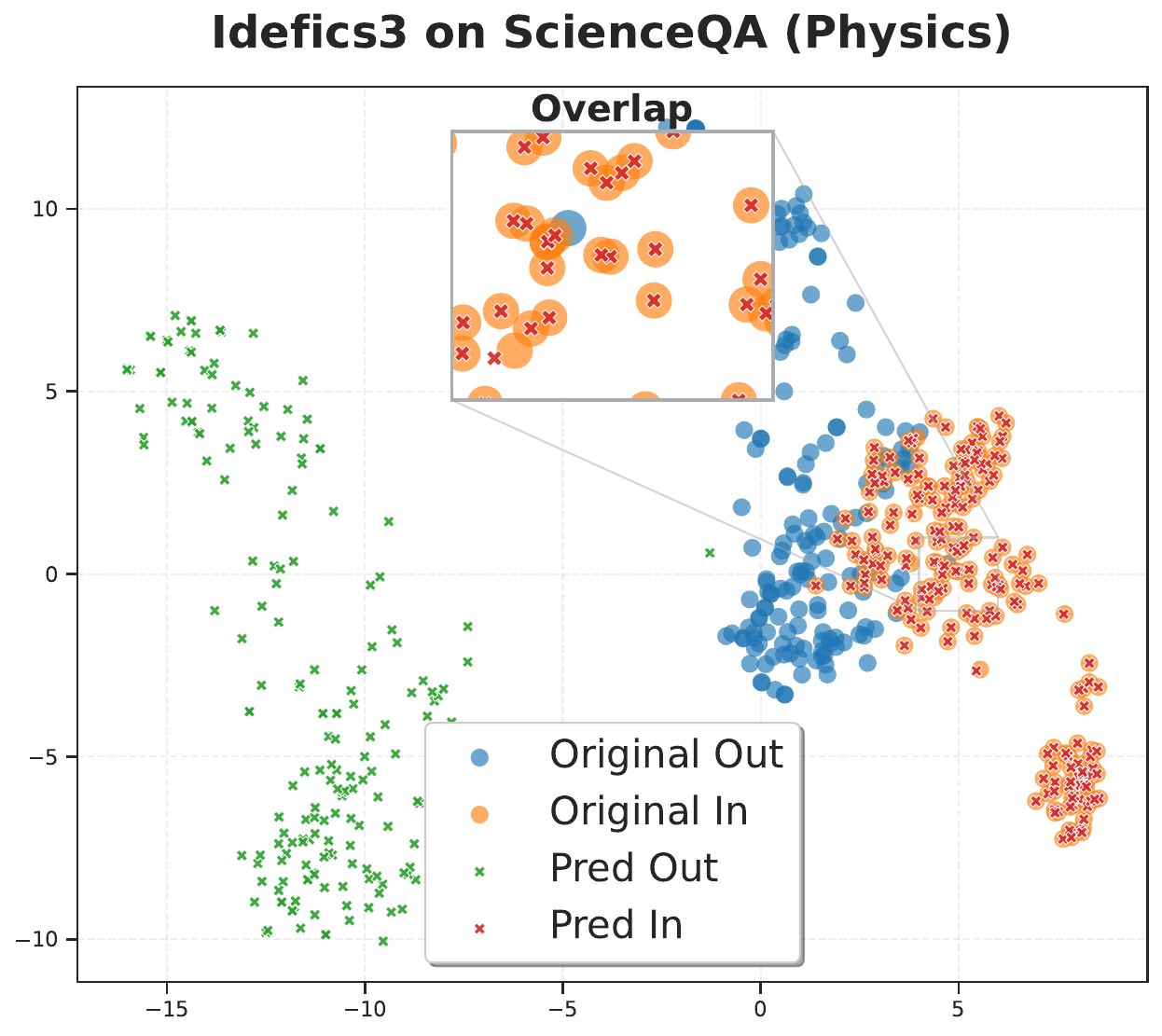}
        \end{subfigure}

        \vspace{0.25cm}

        \begin{subfigure}{0.32\textwidth}
            \centering
            \includegraphics[width=\linewidth]{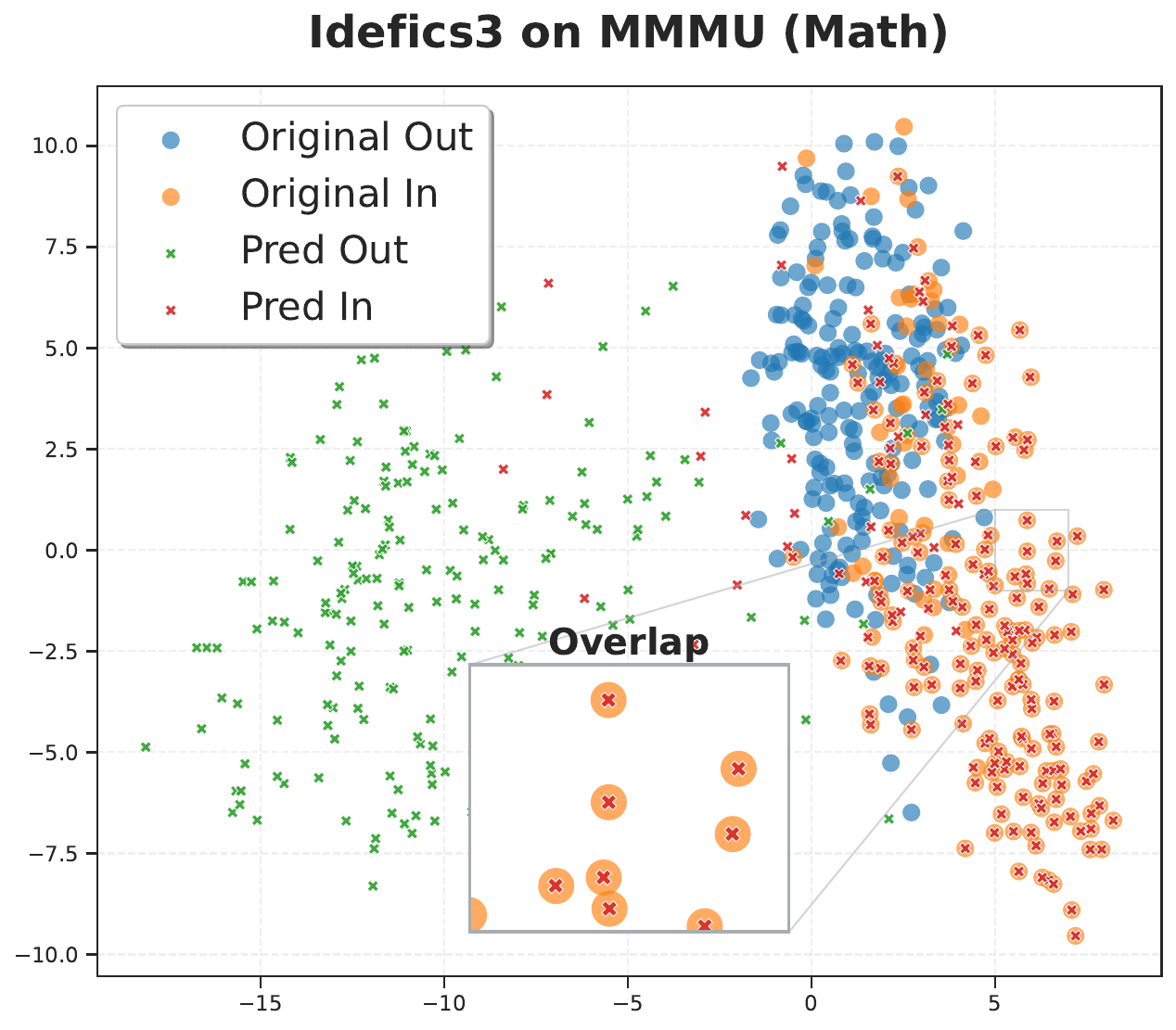}
        \end{subfigure}
        \hfill
        \begin{subfigure}{0.32\textwidth}
            \centering
            \includegraphics[width=\linewidth]{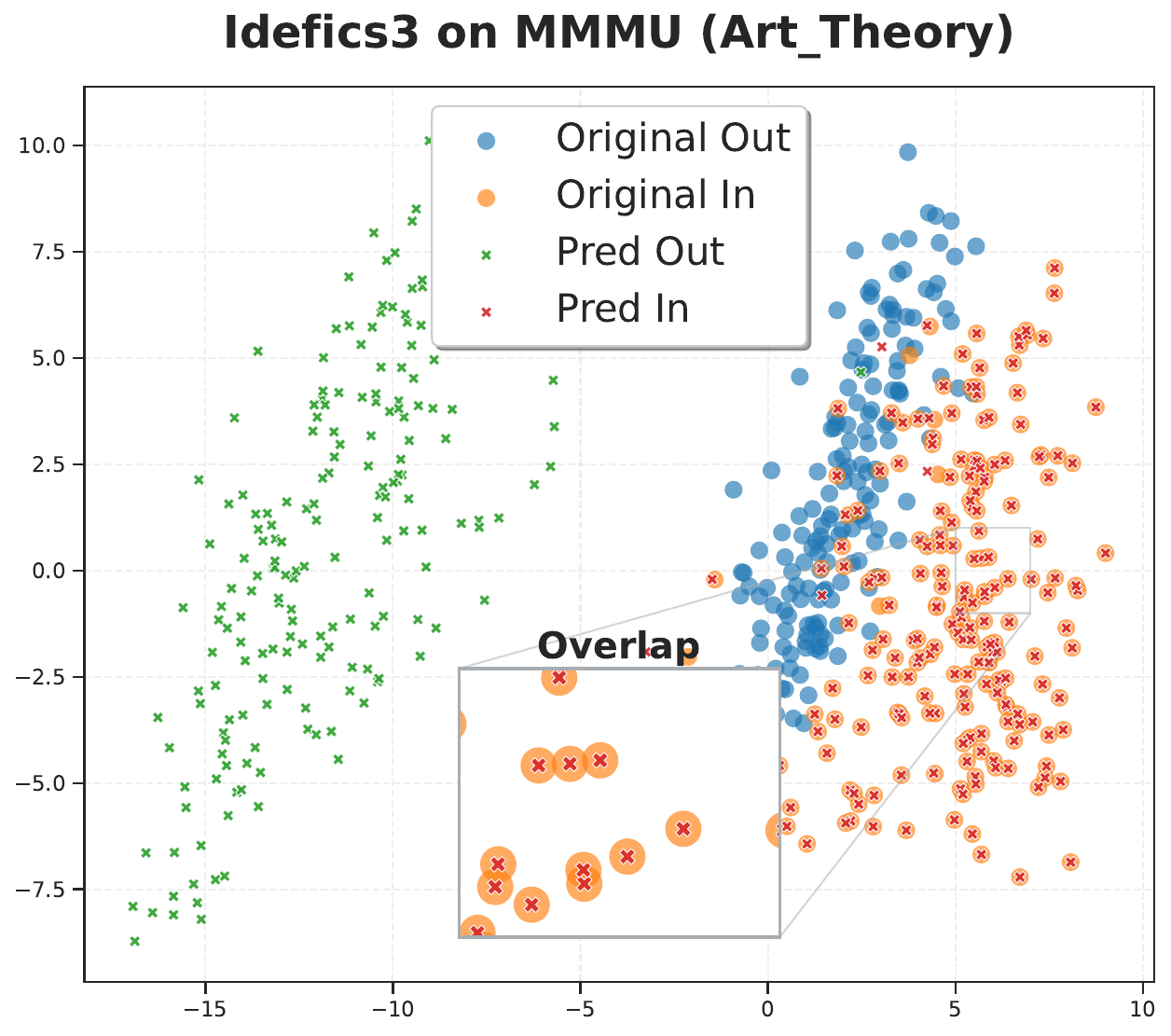}
        \end{subfigure}
        \hfill
        \begin{subfigure}{0.32\textwidth}
            \centering
            \includegraphics[width=\linewidth]{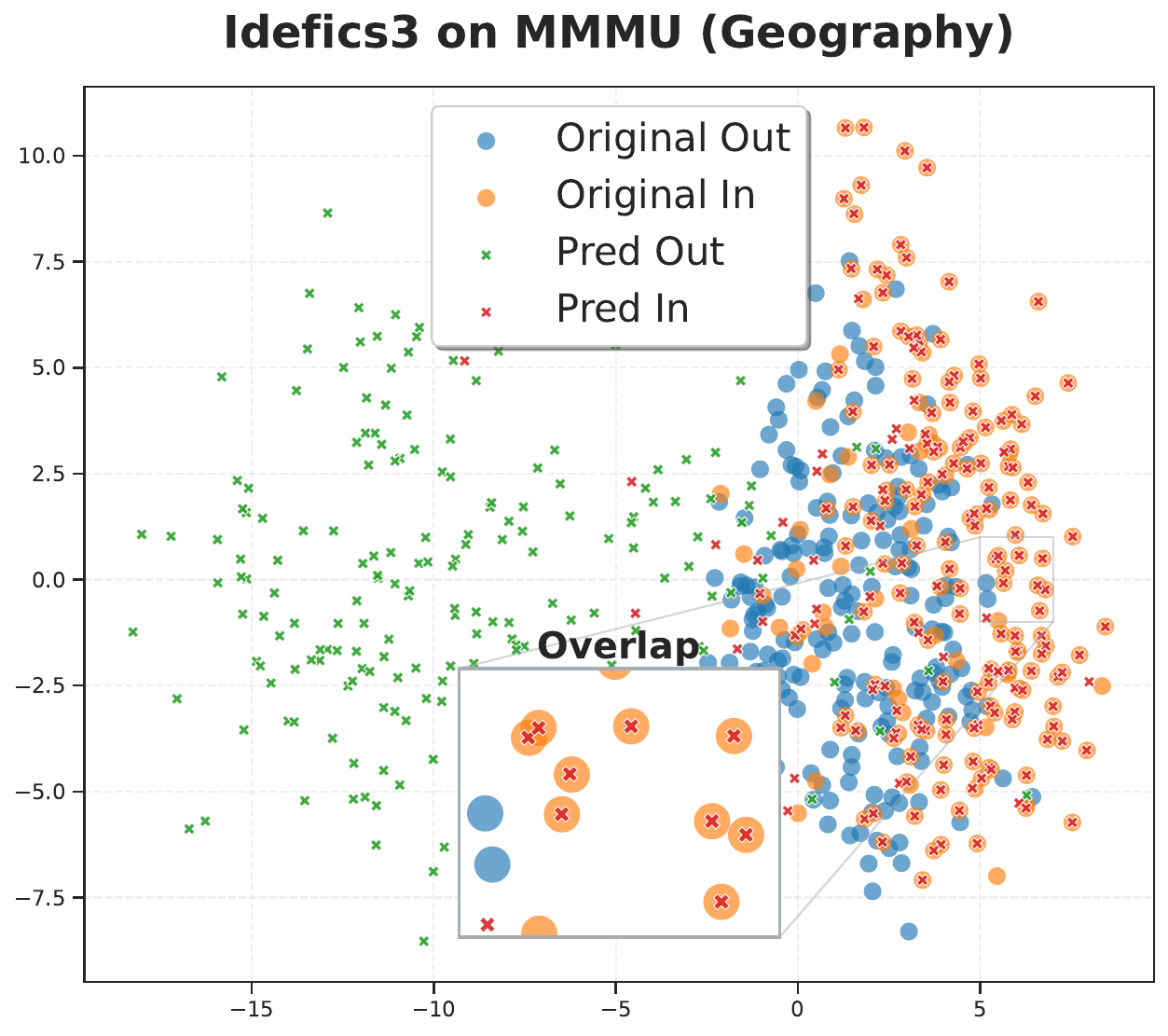}
        \end{subfigure}
        
    \end{minipage}
    }
    \caption{(EQ2) Activation distribution shifts at the 25$^{th}$ layer for in-scope and out-of-scope test samples across different subjects in ScienceQA and MMMU on Idefics3-8B-Llama3.}
    \label{fig:RQ3_Idefics3}
\end{figure*}

\begin{figure*}
    \centering
    \resizebox{0.95\textwidth}{!}{%
    \begin{minipage}{\textwidth}
        \centering
        \begin{subfigure}{0.32\textwidth}
            \centering
            \includegraphics[width=\linewidth]{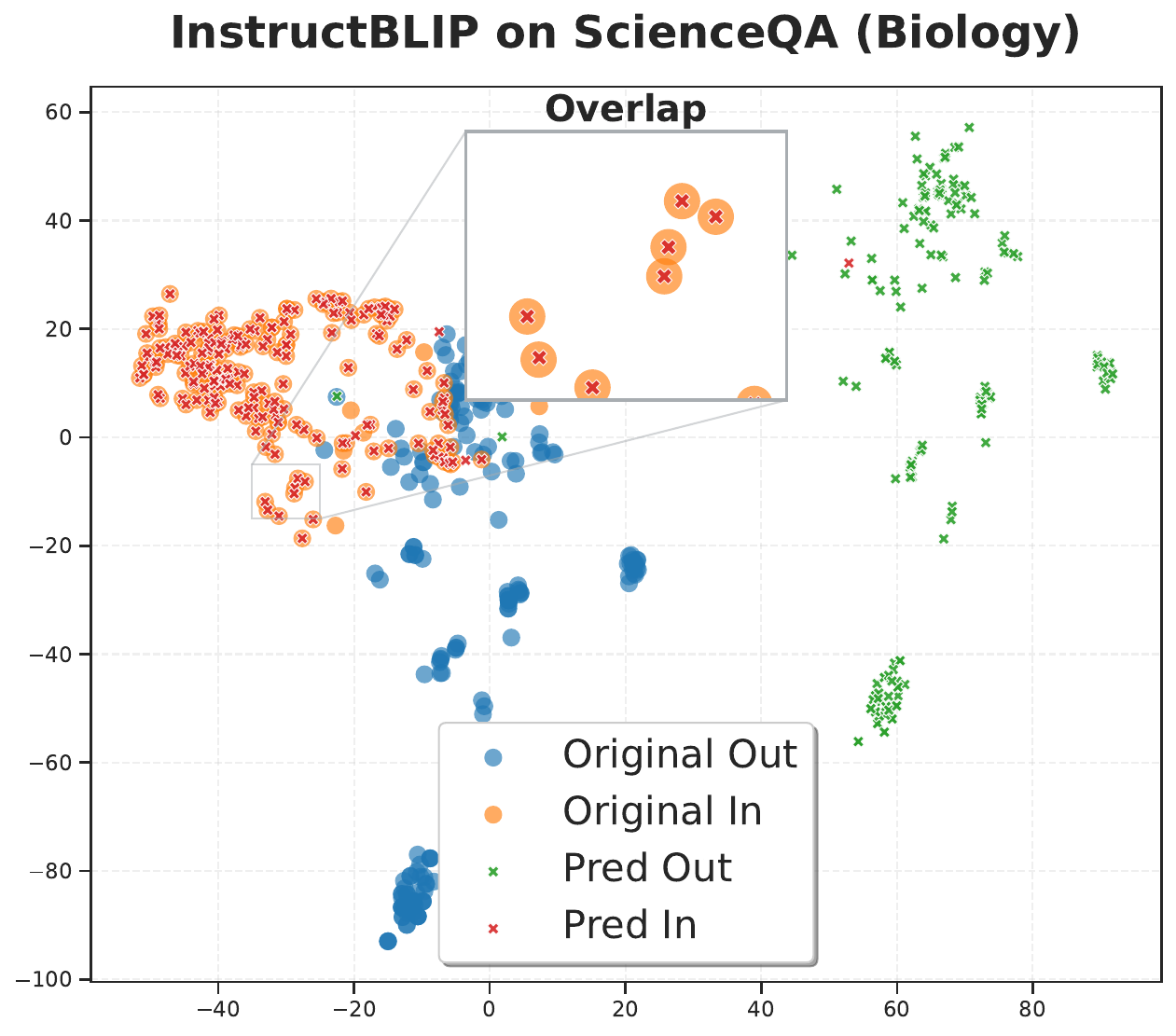}
        \end{subfigure}
        \hfill
        \begin{subfigure}{0.32\textwidth}
            \centering
            \includegraphics[width=\linewidth]{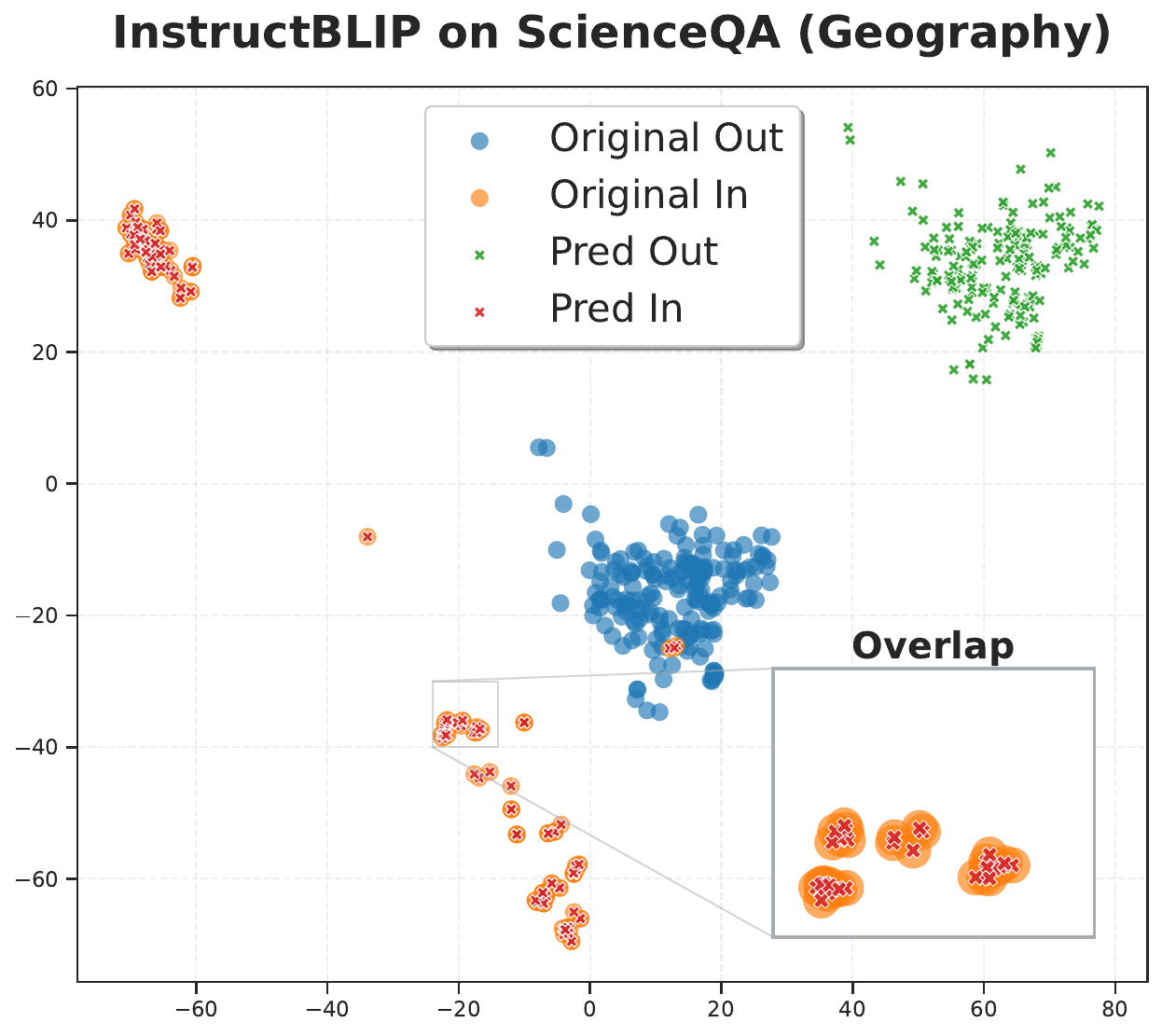}
        \end{subfigure}
        \hfill
        \begin{subfigure}{0.32\textwidth}
            \centering
            \includegraphics[width=\linewidth]{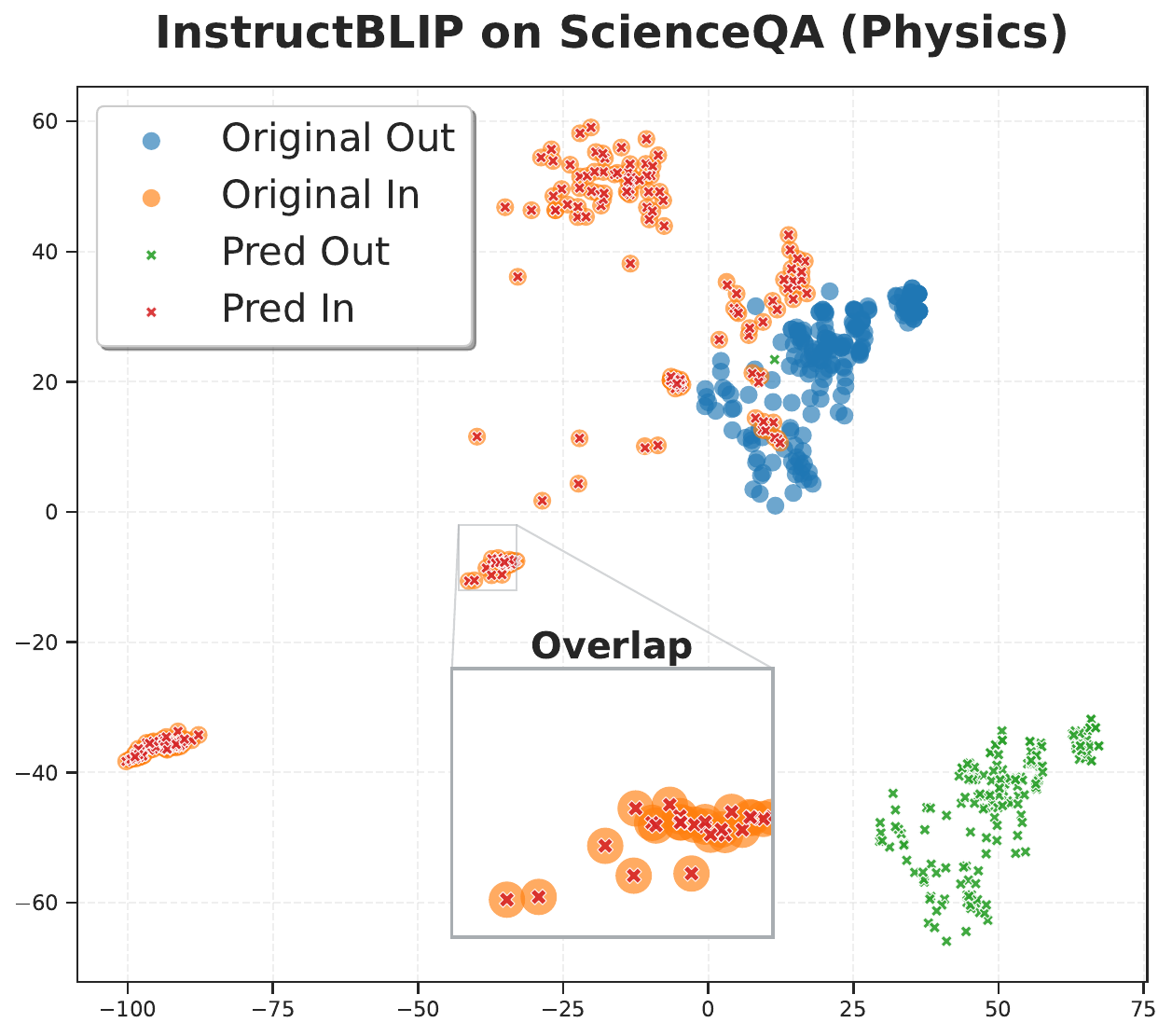}
        \end{subfigure}

        \vspace{0.25cm}

        \begin{subfigure}{0.32\textwidth}
            \centering
            \includegraphics[width=\linewidth]{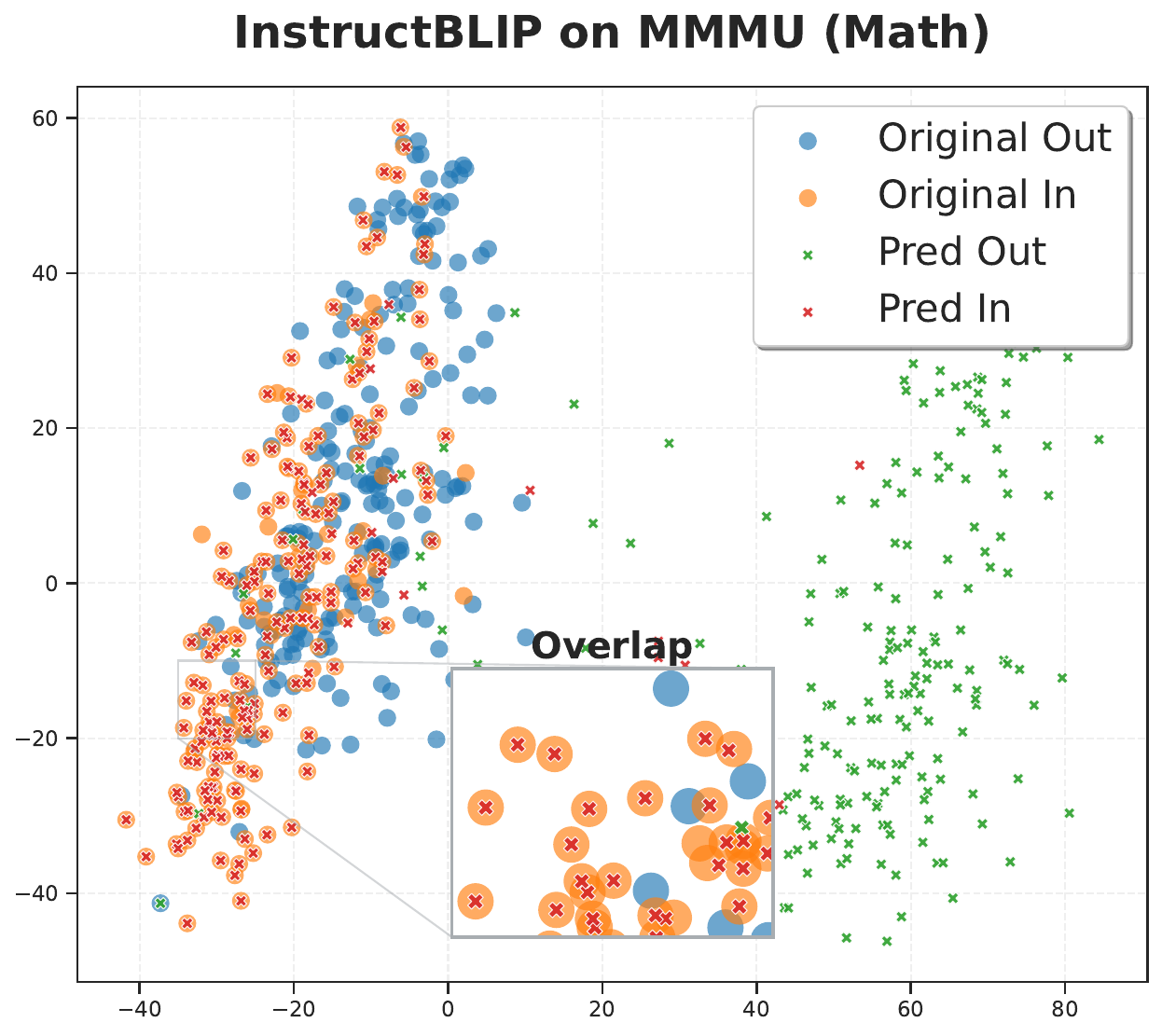}
        \end{subfigure}
        \hfill
        \begin{subfigure}{0.32\textwidth}
            \centering
            \includegraphics[width=\linewidth]{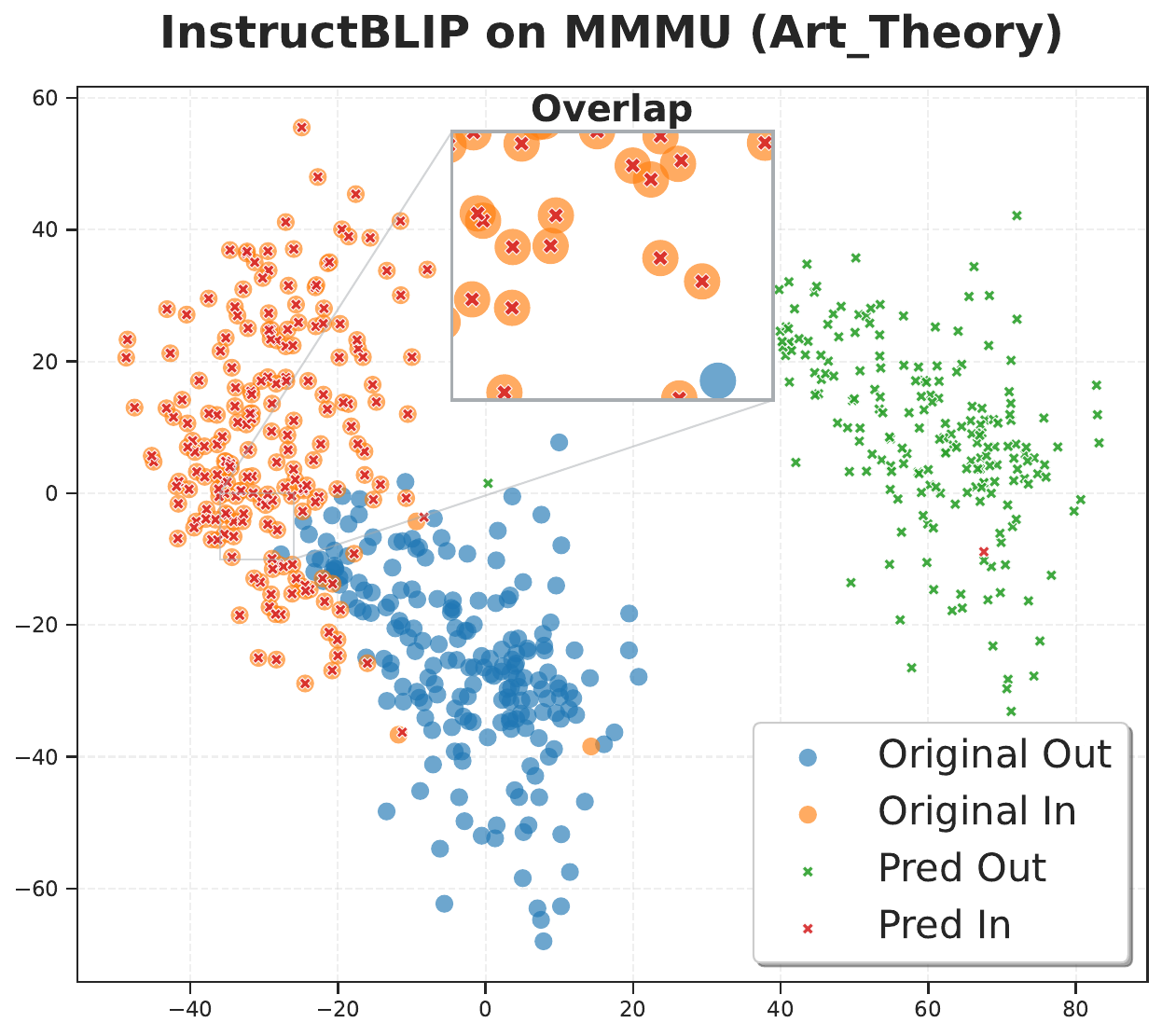}
        \end{subfigure}
        \hfill
        \begin{subfigure}{0.32\textwidth}
            \centering
            \includegraphics[width=\linewidth]{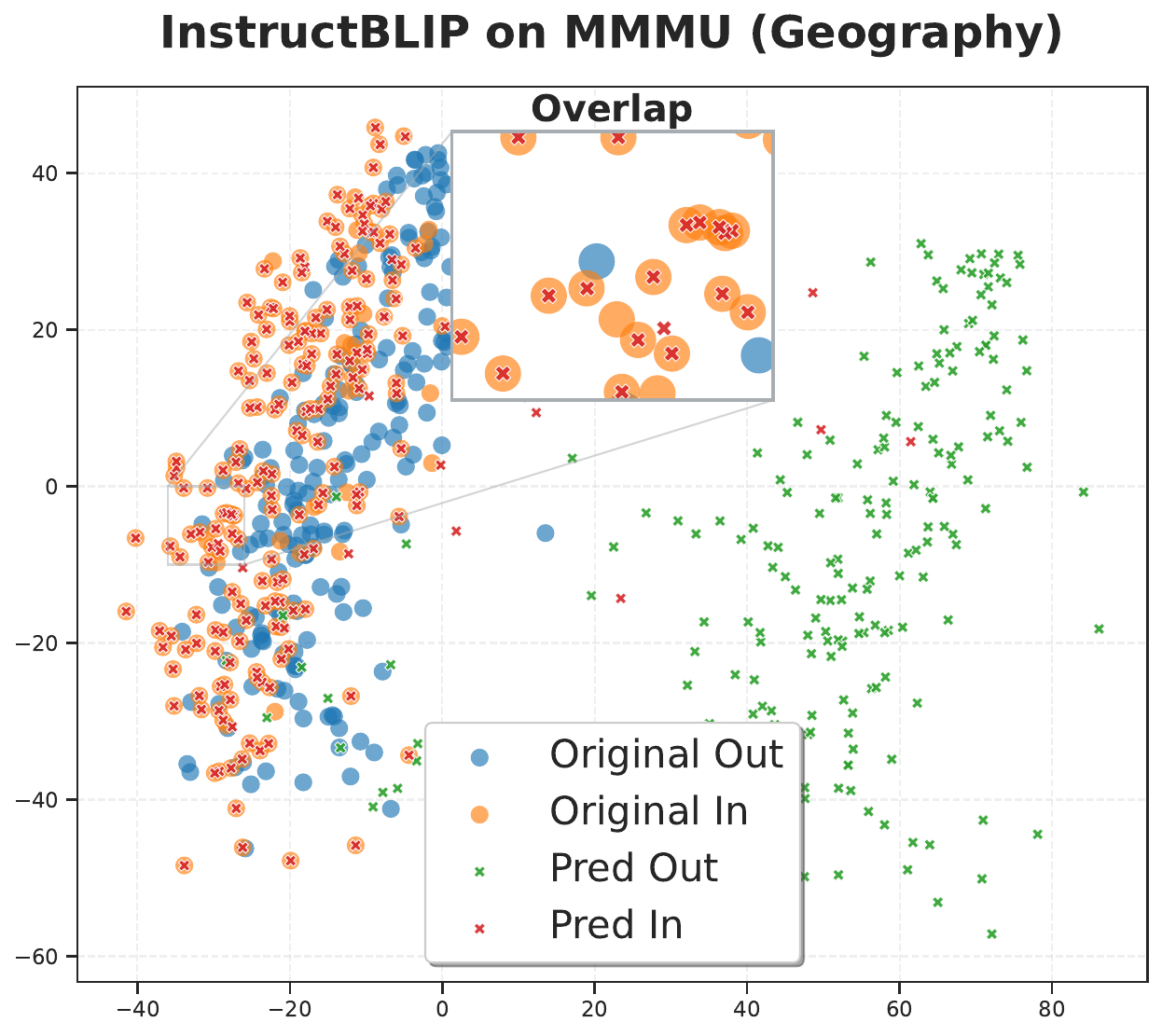}
        \end{subfigure}
        
    \end{minipage}
    }
    \caption{(EQ2) Activation distribution shifts at the 25$^{th}$ layer for in-scope and out-of-scope test samples across different subjects in ScienceQA and MMMU on InstructBLIP-Vicuna-7B.}
    \label{fig:RQ3_InstructBLIP}
\end{figure*}
% \FloatBarrier

\section{(EQ2) Additional Results of Activation Distribution}\label{appendix:activation_distribution}
We report the activation distribution shifts for additional VLMs in Figure~\ref{fig:RQ3_LLava_13b}, Figure~\ref{fig:RQ3_Idefics3}, and Figure~\ref{fig:RQ3_InstructBLIP}, to further validate the generality of our configurable refusal method. Specifically, we visualize the calibrated and original activations for LLaVA-1.5-13B-hf, Idefics3-8B-Llama3, and InstructBLIP-Vicuna-7B across different subject domains. 
% In addition, we report supplementary experimental results on LLaVA-1.5-7B-hf evaluated on MMMU dataset in Figure~\ref{appeddix:MMMU_llava_7b}.

Consistent with the observations in the main paper, calibrated activations for out-of-scope queries are clearly separated from both their original counterparts and in-scope activations, forming compact clusters associated with refusal behavior. Meanwhile, activations corresponding to in-scope queries remain close to their original distributions, indicating that acceptance behavior is largely preserved. 
These results demonstrate that the activation-level separation induced by our method is stable across model architectures and scales, further confirming the effectiveness and robustness of our configurable refusal mechanism.

\section{Refusal Score Calculation}\label{app:refusal_score_calculation}
We calculate a refusal score followed by~\cite{arditi2024refusal,wang2024surgical} in Eq. (\ref{eq:refusal_score}), which quantifies the model’s inclination to generate refusal-related tokens. Formally, at the first decoding step, we compute the log-odds between refusal-related tokens $\mathcal{R}$ (\eg ``Sorry'', ``I'') and all other tokens $\mathcal{V}\setminus \mathcal{R}$ as refusal score. A higher refusal score demonstrates a stronger tendency to refuse. The equation for calculating the refusal score is shown as follows:
\begin{equation}
    \scalebox{0.4}{
        \label{eq:refusal_score}
        \resizebox{0.89\linewidth}{!}{$
            \mathcal{T}(x) = \log \Bigg( \sum_{t \in \mathcal{R}} p_t \Bigg) - \log \Bigg( \sum_{t \in \mathcal{V}\setminus \mathcal{R}} p_t \Bigg).
        $}
    }
\end{equation}

\section{Refusal Score Shift}\label{appendix:refusal_score}
To obtain a more fine-grained understanding of how our method modulates refusal behavior, we examine how our proposed method shifts the underlying tendency of the model to generate refusal-related tokens.
The details of calculating refusal score is introduced in section~\ref{app:refusal_score_calculation} in the appendix.
% Followings~\cite{arditi2024refusal,wang2024surgical}, we adopt the refusal score, which measures the log-odds of generating refusal-related tokens at the first decoding step. 
Unlike the refusal rate, which provides a binary view of whether a refusal occurs, the refusal score captures subtle changes in the refusal tendency of the model.

\begin{figure*}[h]
    \centering
    
    \begin{subfigure}{0.23\textwidth}
        \centering
        \includegraphics[width=\linewidth]{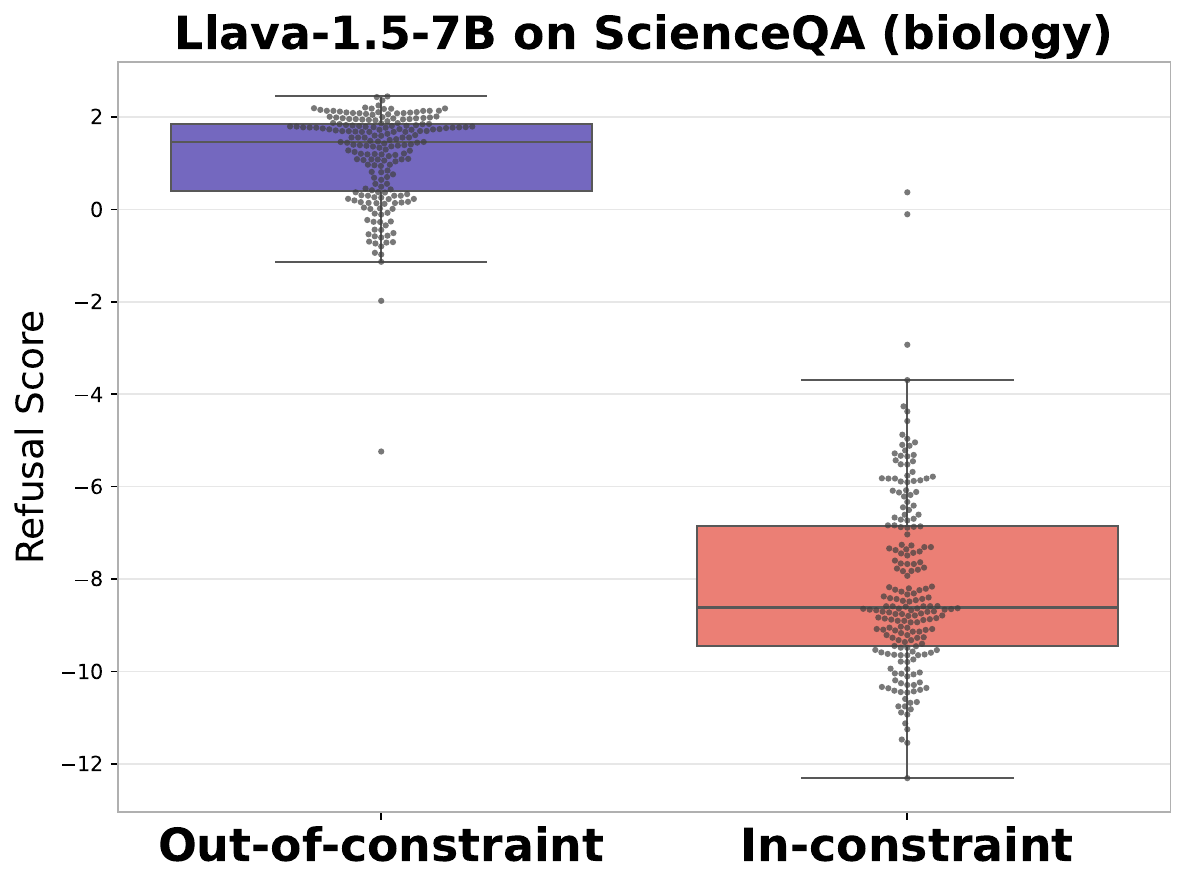}
        % \caption{Layer 10}
        \label{fig:a}
    \end{subfigure}
    \hfill
    \begin{subfigure}{0.23\textwidth}
        \centering
        \includegraphics[width=\linewidth]{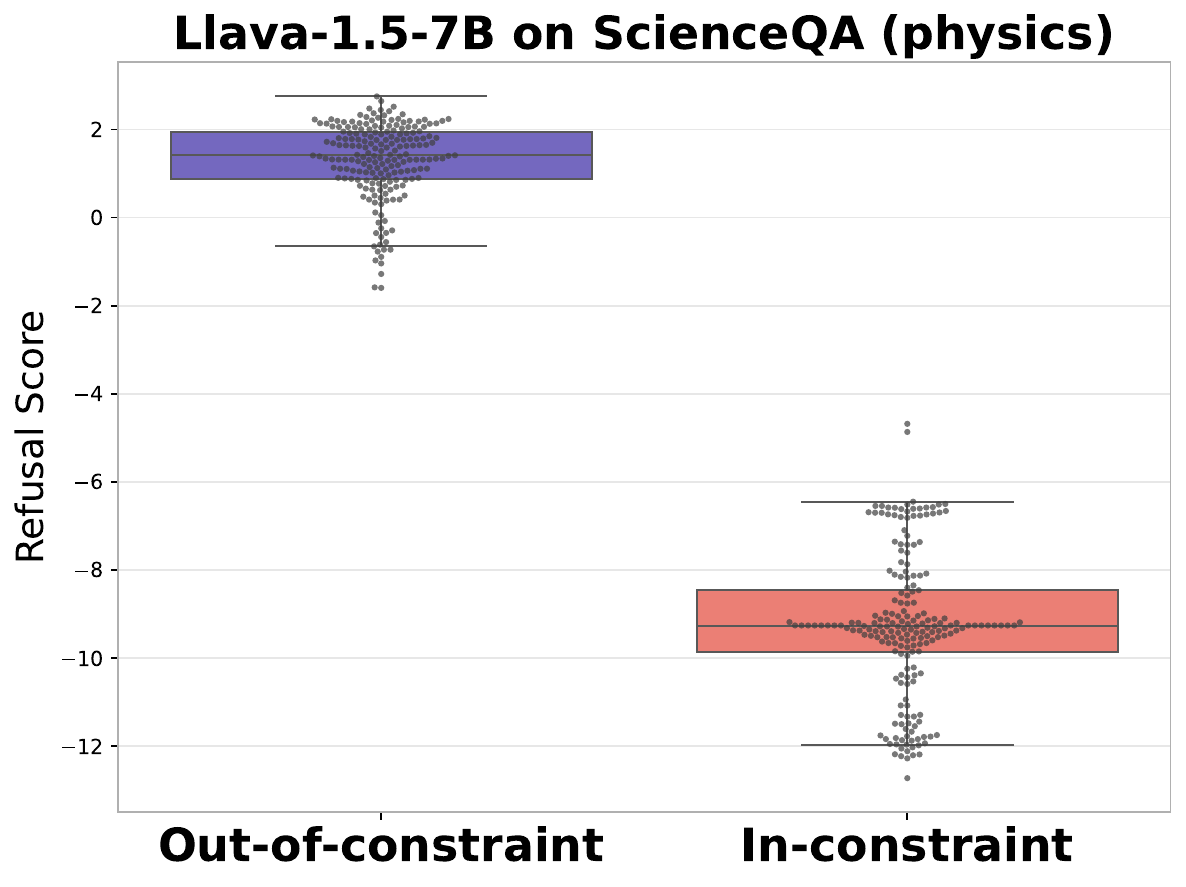}
        % \caption{llava-1.5-7b-hf on ScienceQA (physics)}
        \label{fig:b}
    \end{subfigure}
    \hfill
    \begin{subfigure}{0.23\textwidth}
        \centering
        \includegraphics[width=\linewidth]{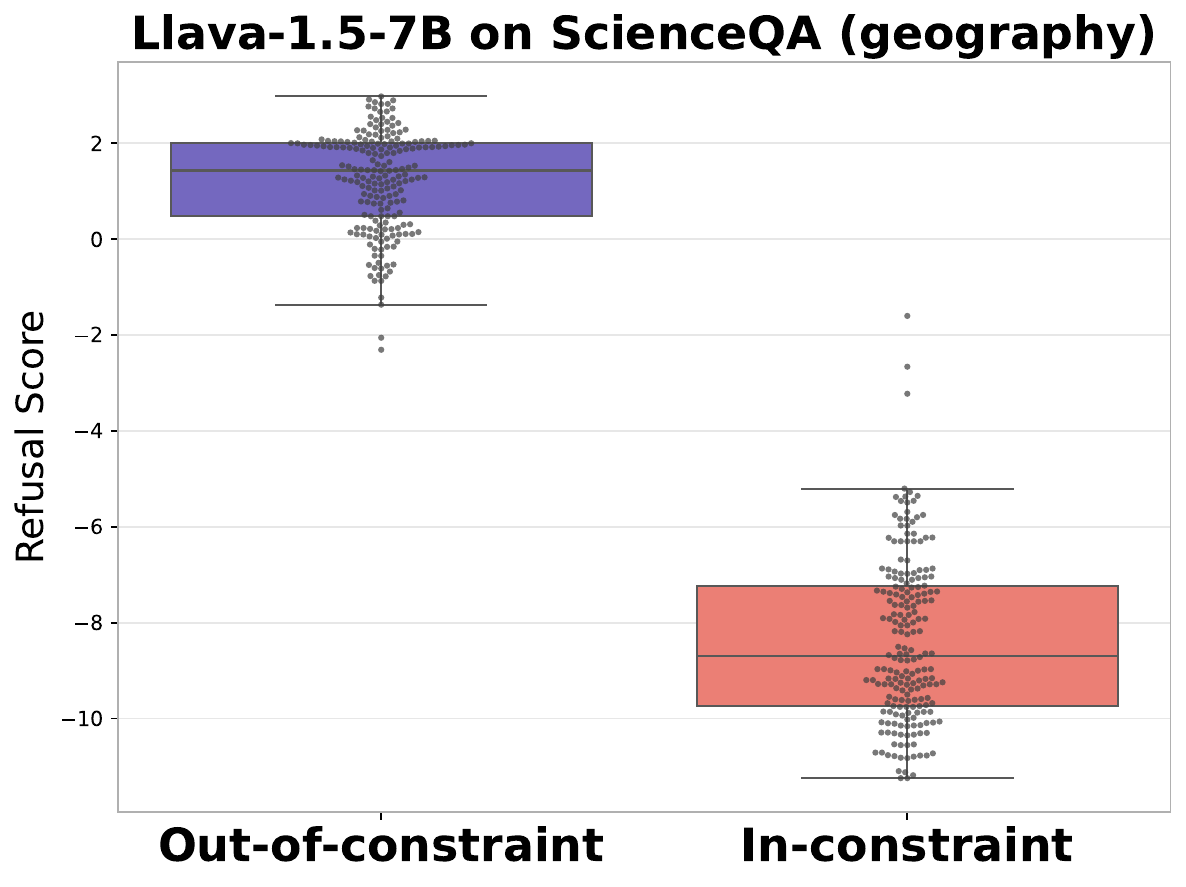}
        \label{fig:c}
    \end{subfigure}
    \hfill
    \begin{subfigure}{0.23\textwidth}
        \centering
        \includegraphics[width=\linewidth]{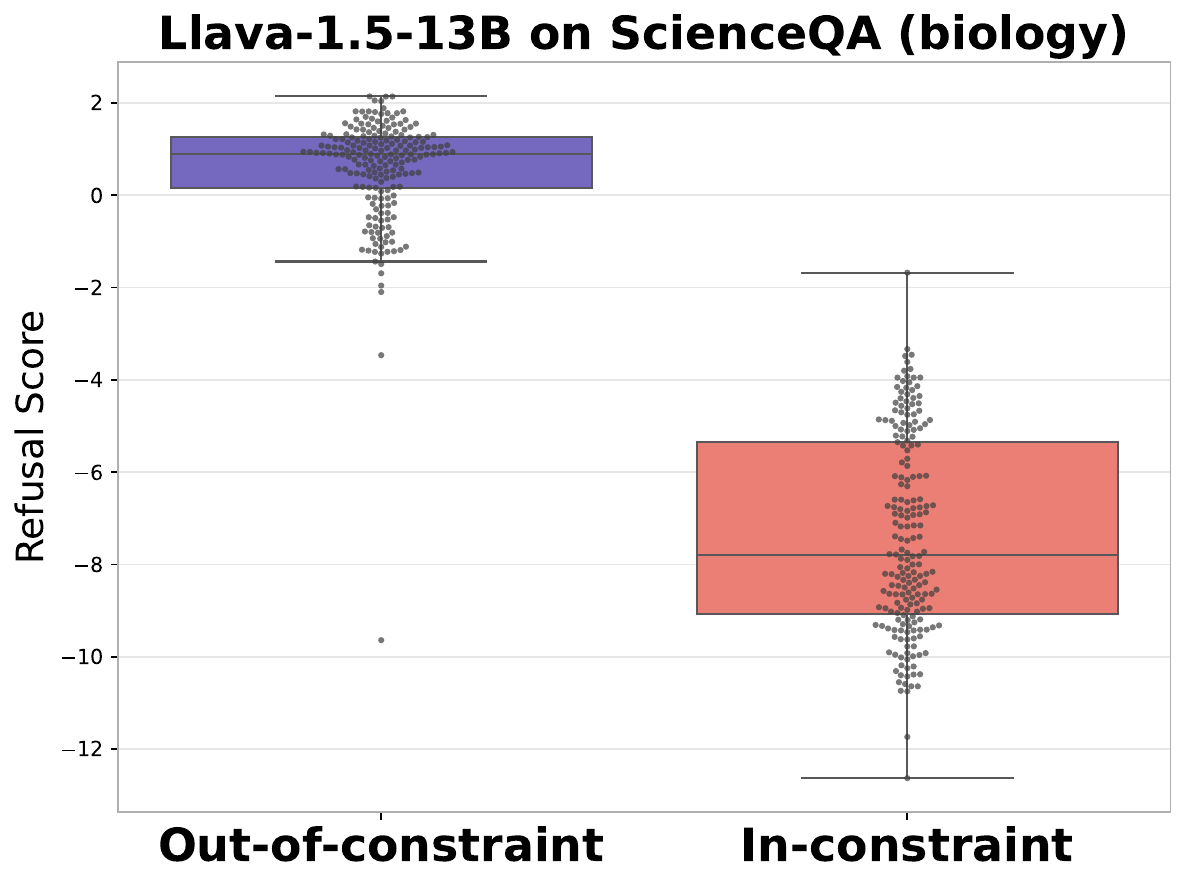}
        \label{fig:c}
    \end{subfigure}
    \hfill
    \begin{subfigure}{0.23\textwidth}
        \centering
        \includegraphics[width=\linewidth]{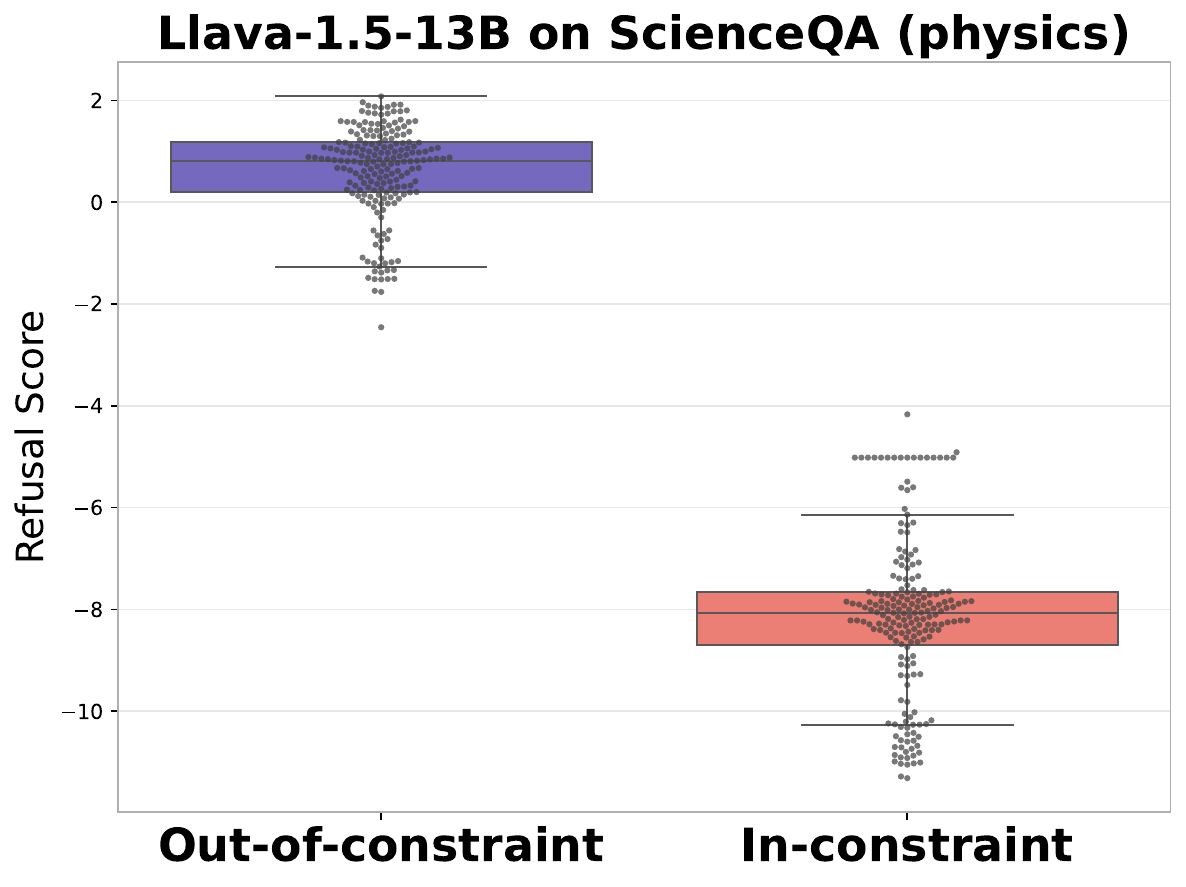}
        \label{fig:c}
    \end{subfigure}
    \hfill
    \begin{subfigure}{0.23\textwidth}
        \centering
        \includegraphics[width=\linewidth]{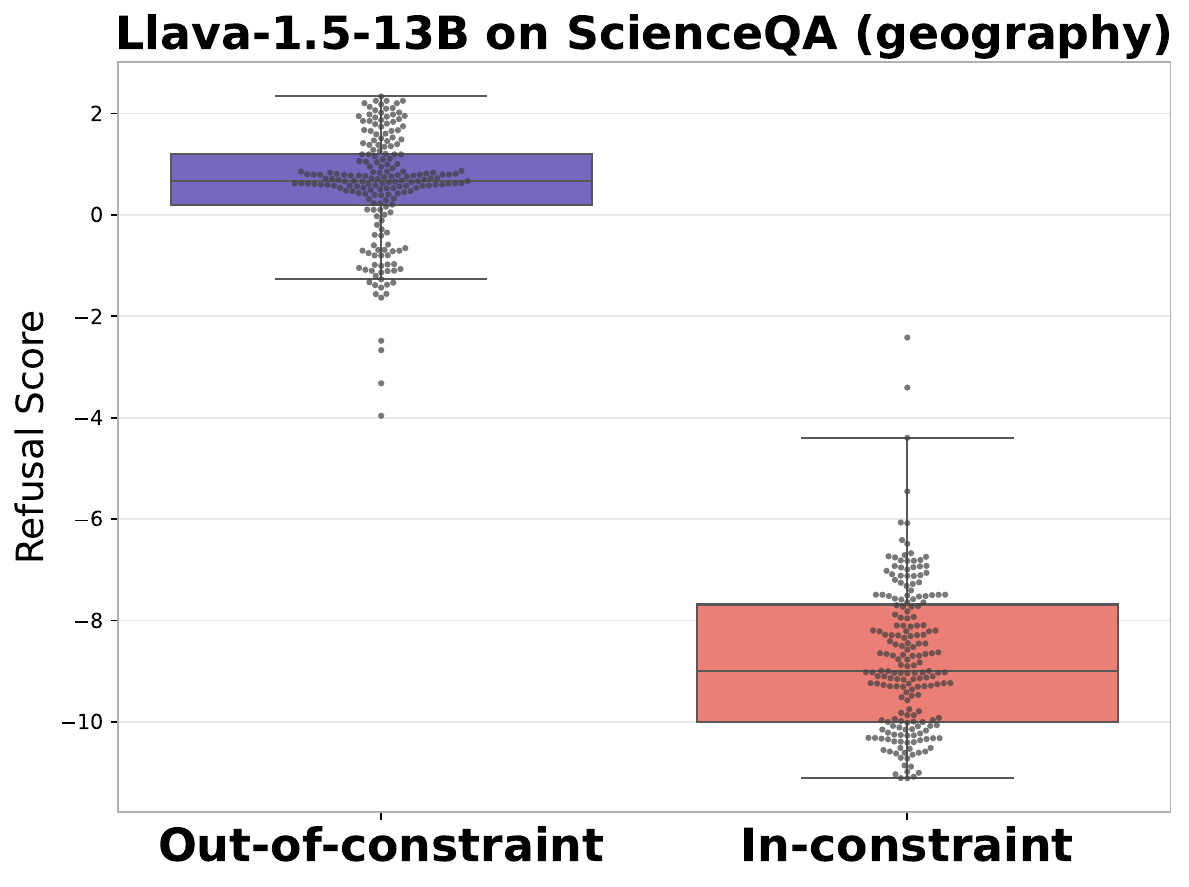}
        \label{fig:c}
    \end{subfigure}
    \hfill
     \begin{subfigure}{0.23\textwidth}
        \centering
        \includegraphics[width=\linewidth]{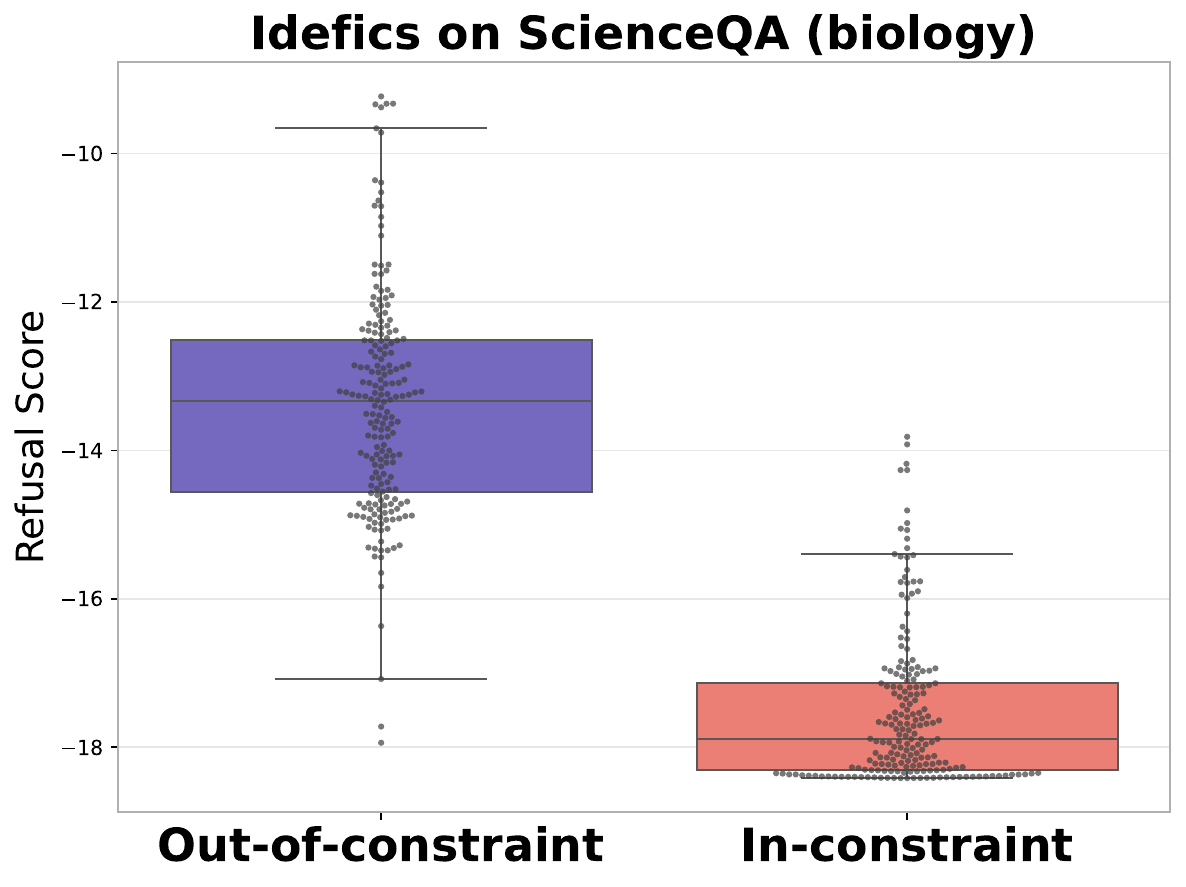}
        \label{fig:a}
    \end{subfigure}
    \hfill
    \begin{subfigure}{0.23\textwidth}
        \centering
        \includegraphics[width=\linewidth]{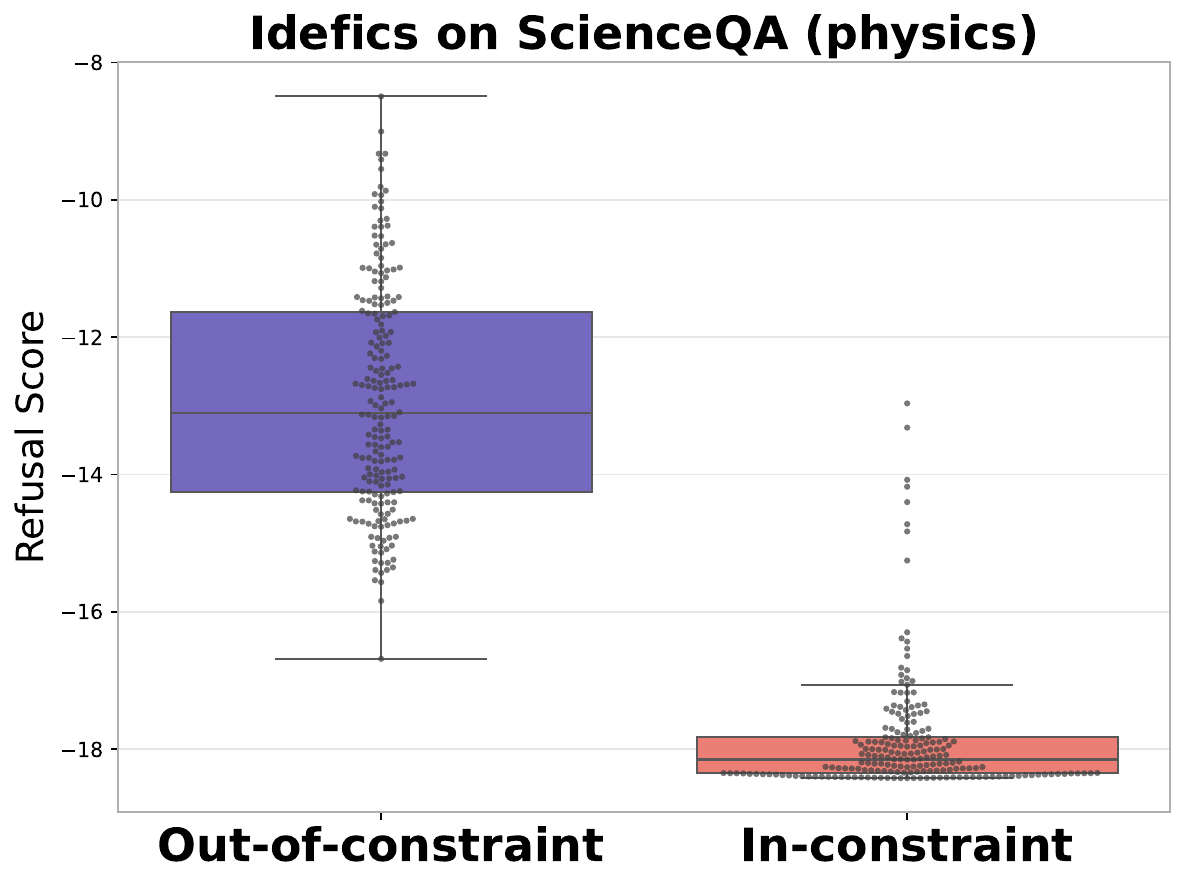}
        \label{fig:b}
    \end{subfigure}
    \hfill
    \begin{subfigure}{0.23\textwidth}
        \centering
        \includegraphics[width=\linewidth]{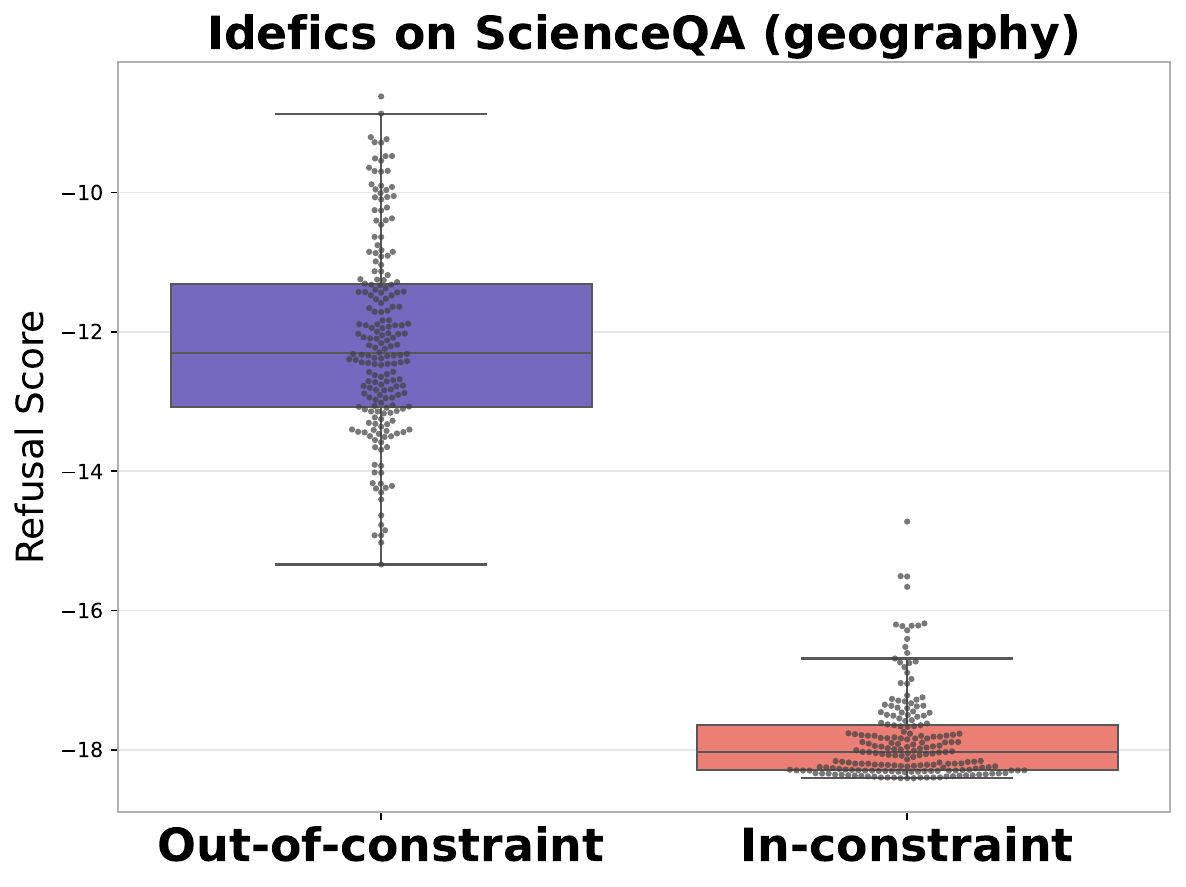}
        \label{fig:c}
    \end{subfigure}
    \hfill
    \begin{subfigure}{0.23\textwidth}
        \centering
        \includegraphics[width=\linewidth]{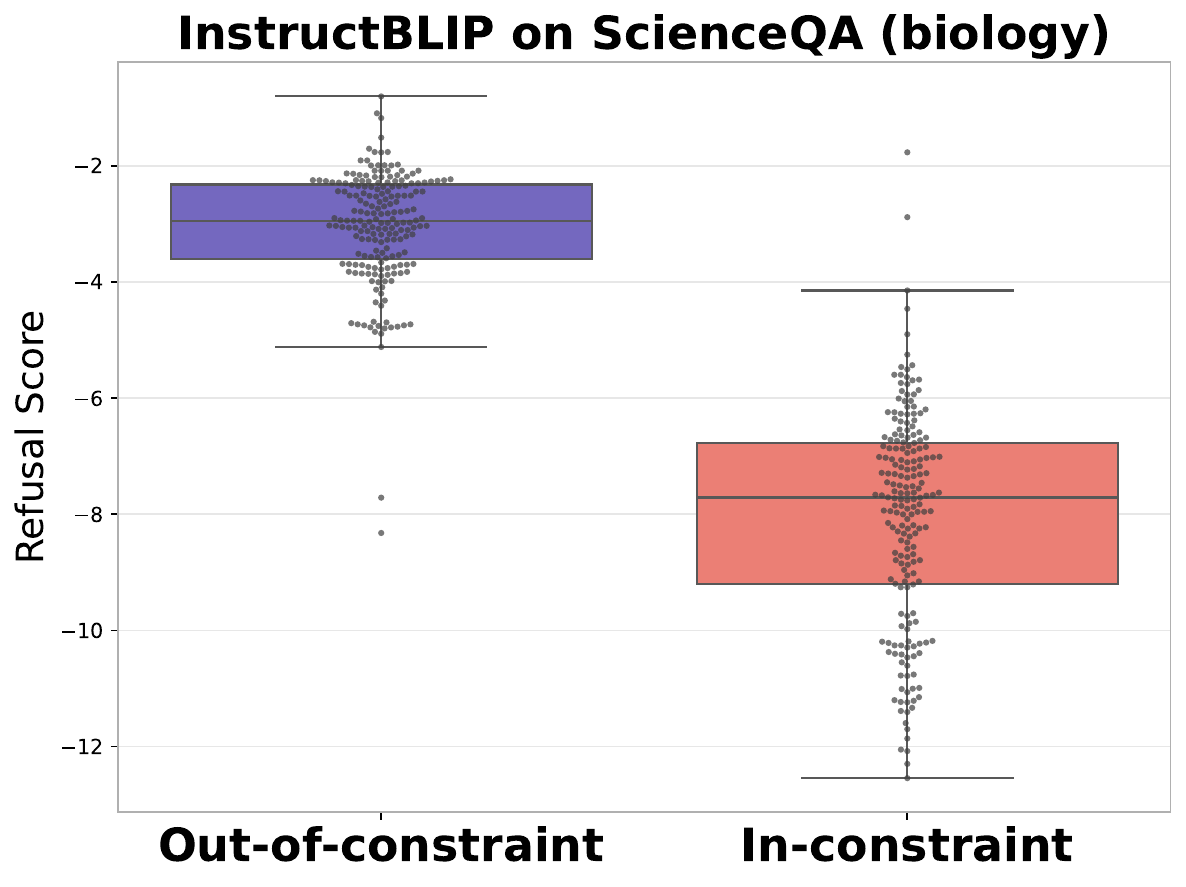}
        \label{fig:c}
    \end{subfigure}
    \hfill
    \begin{subfigure}{0.23\textwidth}
        \centering
        \includegraphics[width=\linewidth]{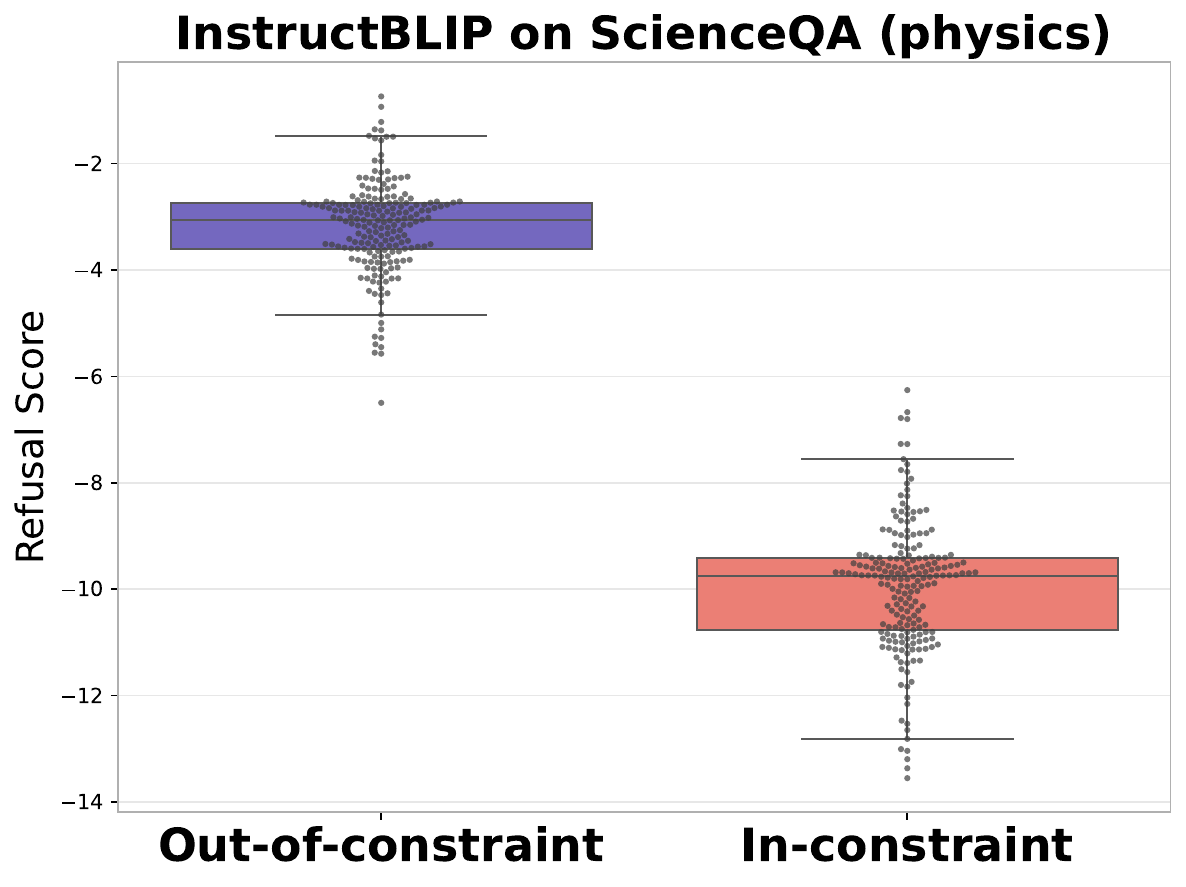}
        \label{fig:c}
    \end{subfigure}
    \hfill
    \begin{subfigure}{0.23\textwidth}
        \centering
        \includegraphics[width=\linewidth]{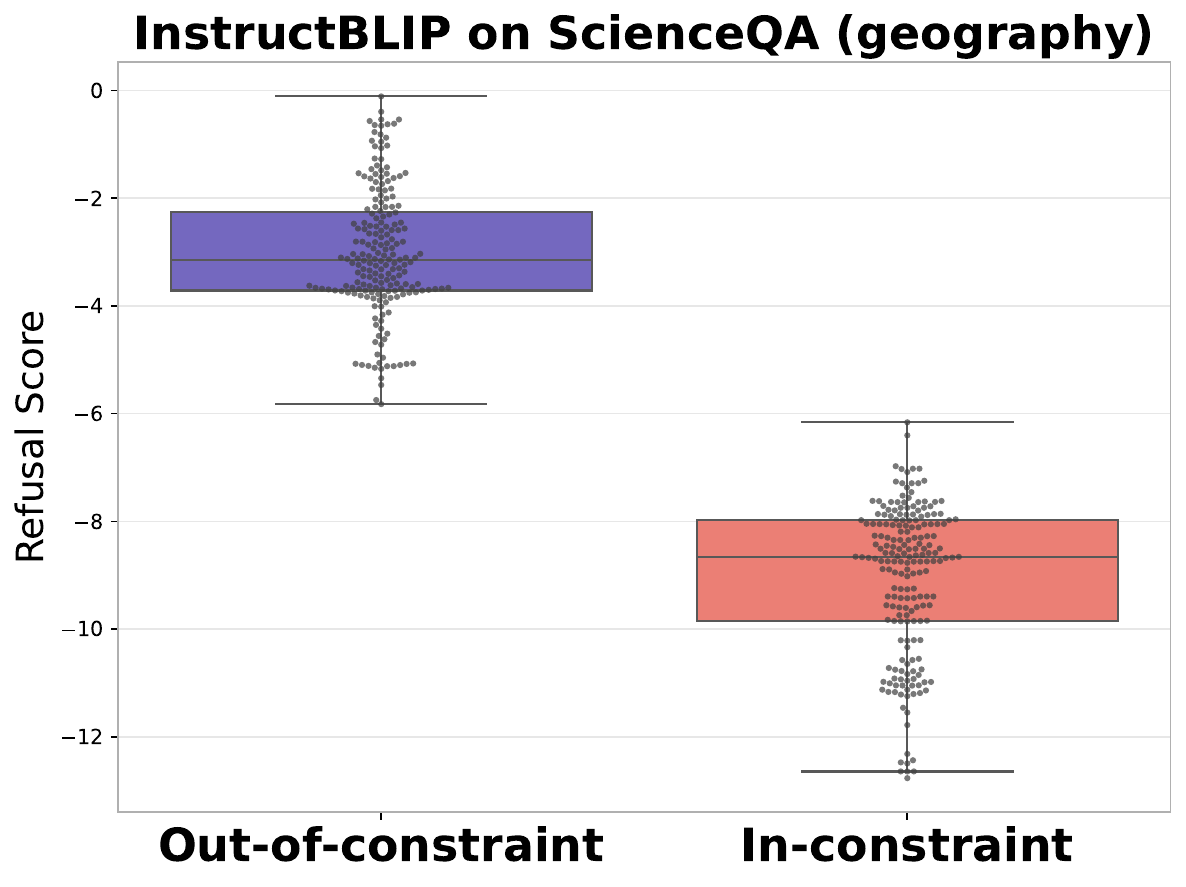}
        \label{fig:c}
    \end{subfigure}
    \vspace{-15pt}
    \caption{Refusal score shift for in-scope and out-of-scope test samples on ScienceQA.}
    \label{fig:RQ4}
\end{figure*}

\begin{figure*}[h]
    \centering
    
    \begin{subfigure}{0.23\textwidth}
        \centering
        \includegraphics[width=\linewidth]{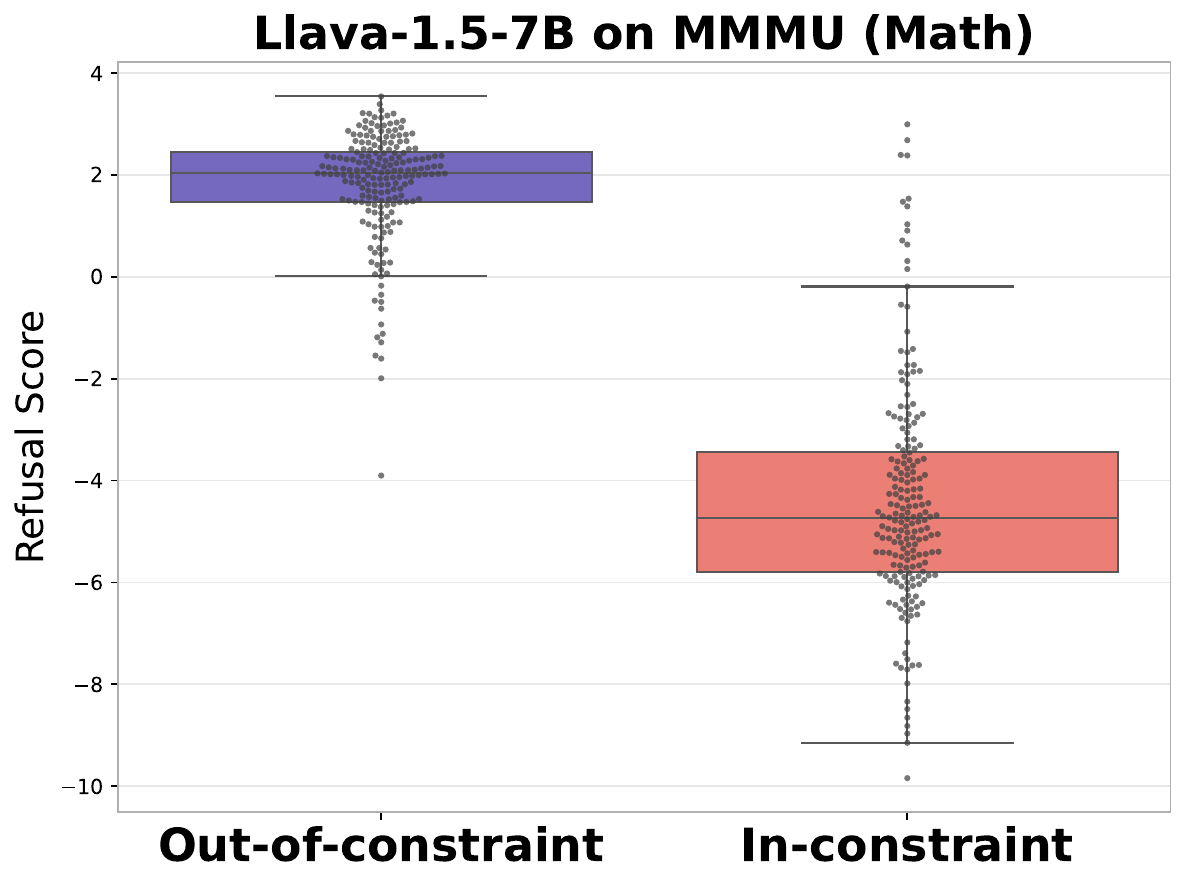}
        % \caption{Layer 10}
        \label{fig:a}
    \end{subfigure}
    \hfill
    \begin{subfigure}{0.23\textwidth}
        \centering
        \includegraphics[width=\linewidth]{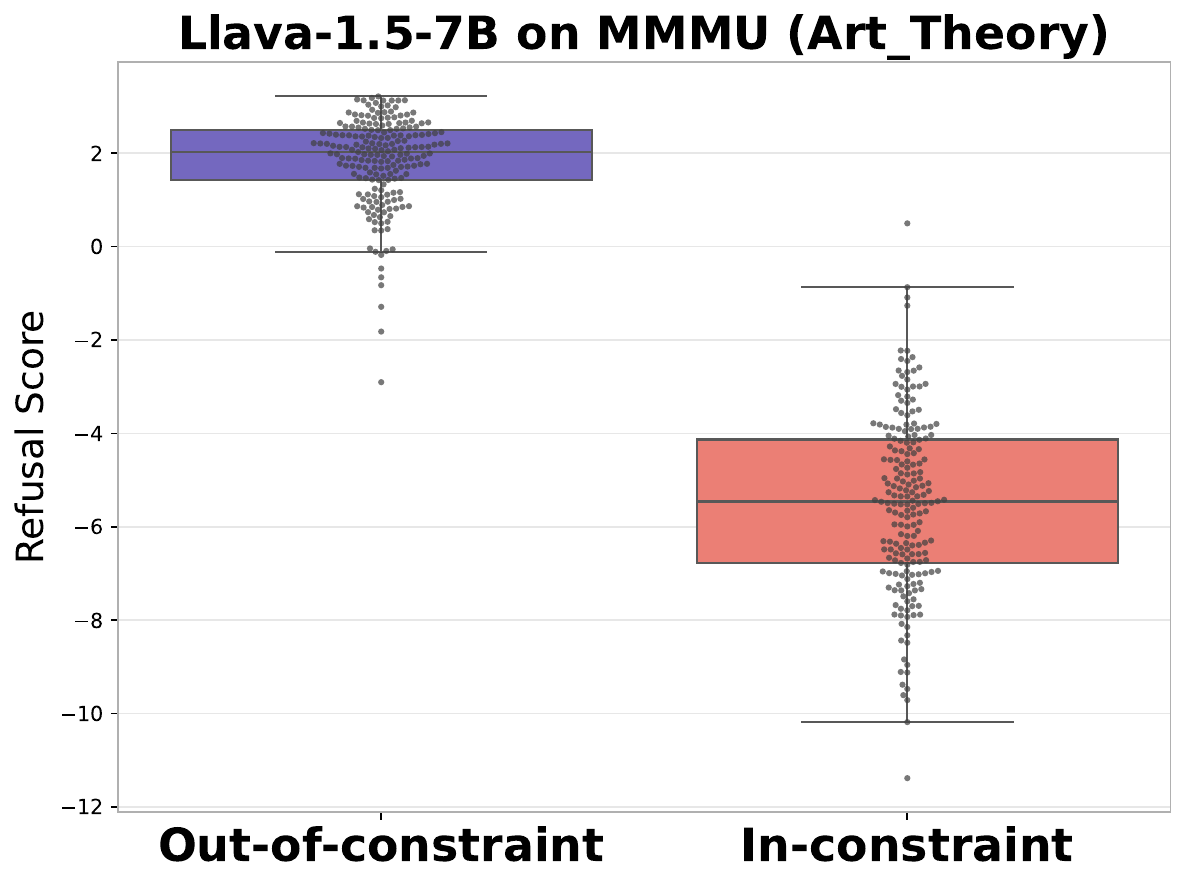}
        % \caption{llava-1.5-7b-hf on ScienceQA (physics)}
        \label{fig:b}
    \end{subfigure}
    \hfill
    \begin{subfigure}{0.23\textwidth}
        \centering
        \includegraphics[width=\linewidth]{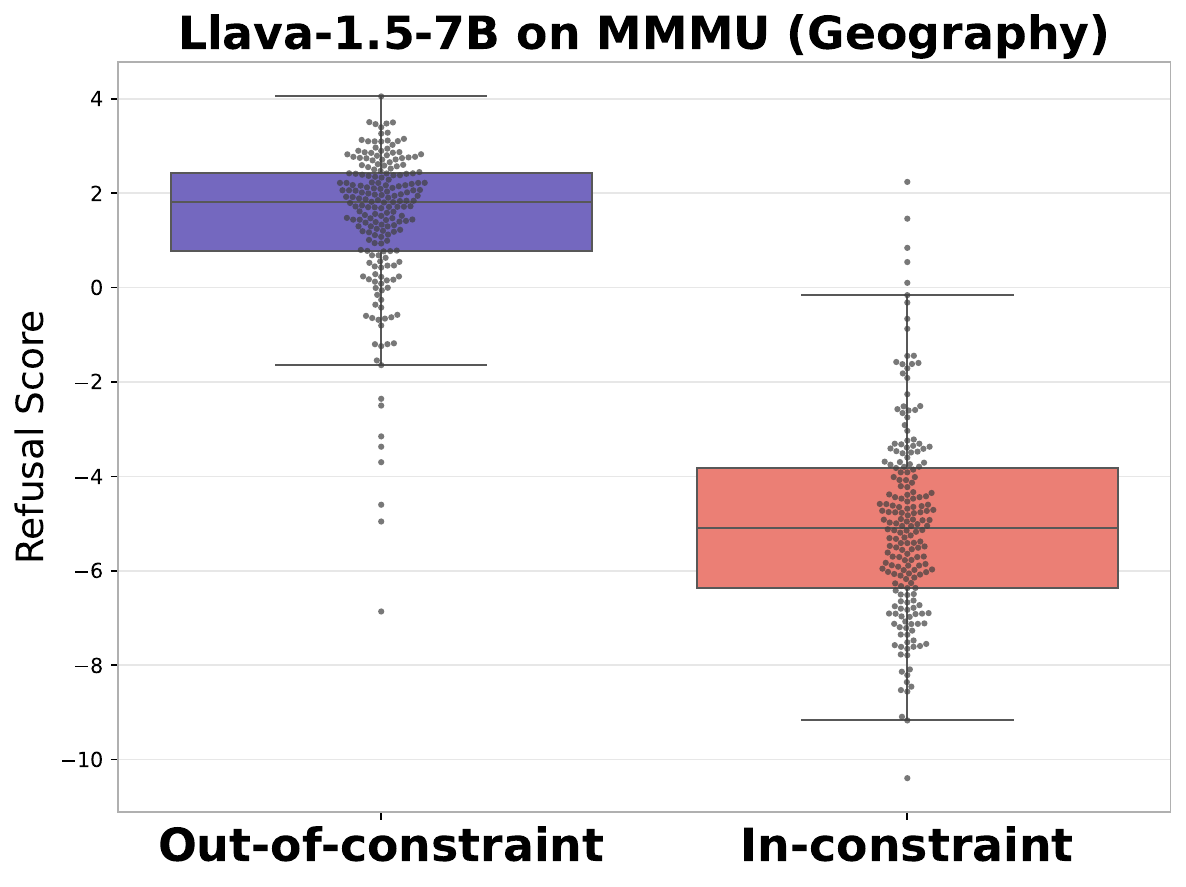}
        \label{fig:c}
    \end{subfigure}
    \hfill
    \begin{subfigure}{0.23\textwidth}
        \centering
        \includegraphics[width=\linewidth]{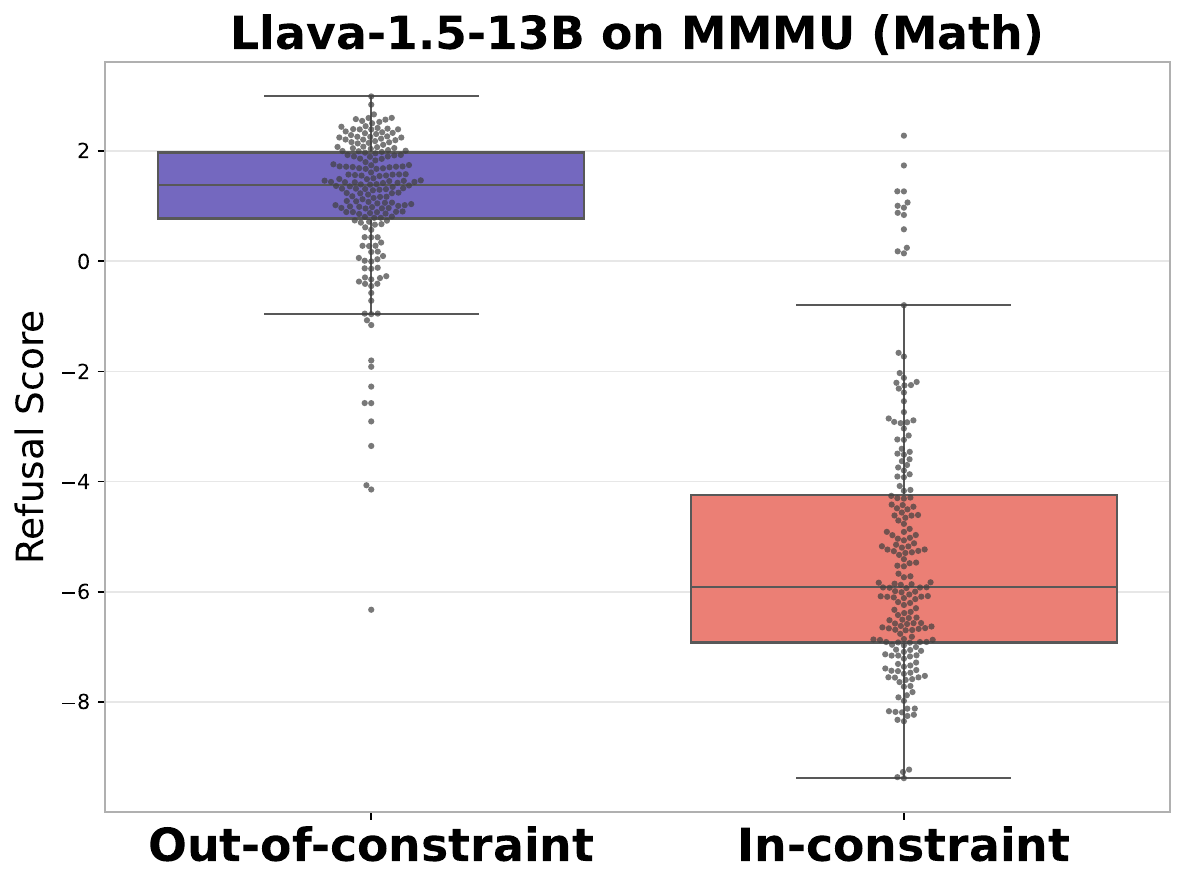}
        \label{fig:c}
    \end{subfigure}
    \hfill
    \begin{subfigure}{0.23\textwidth}
        \centering
        \includegraphics[width=\linewidth]{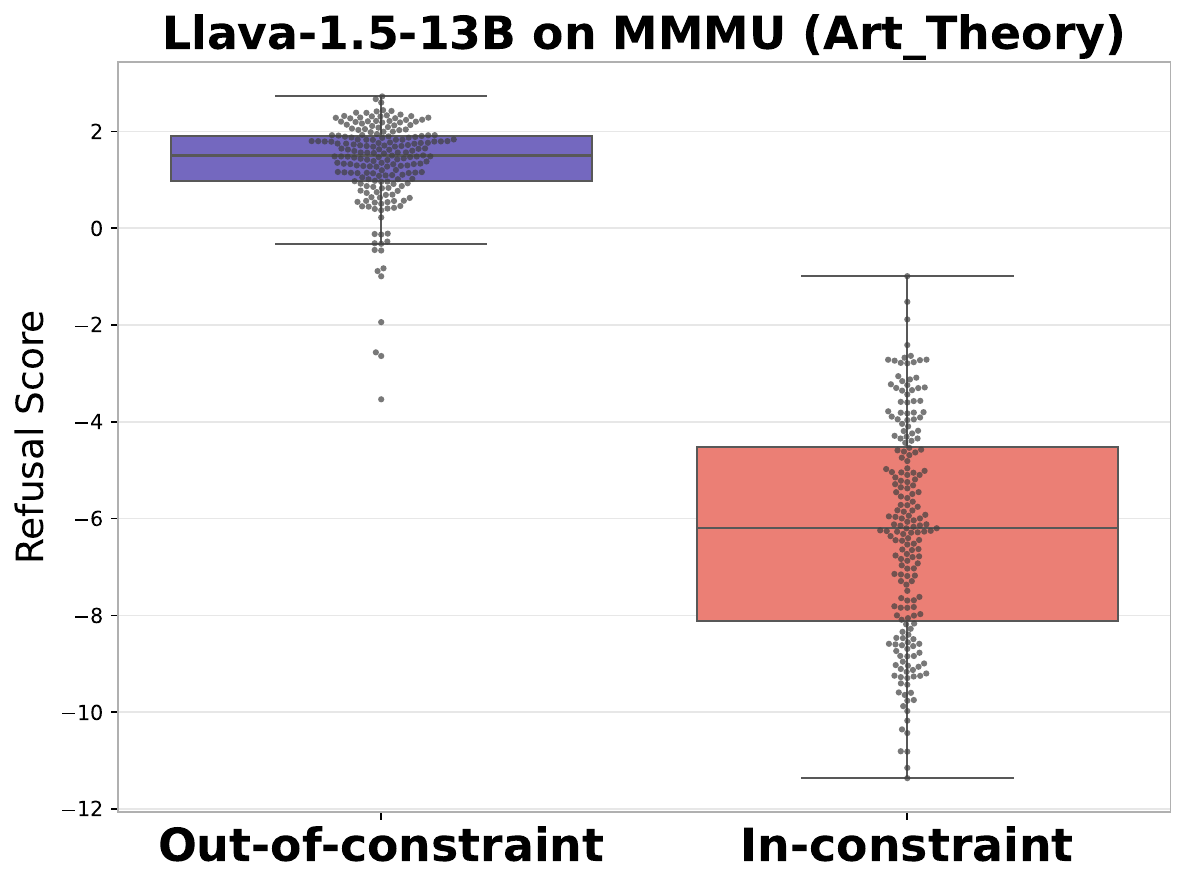}
        \label{fig:c}
    \end{subfigure}
    \hfill
    \begin{subfigure}{0.23\textwidth}
        \centering
        \includegraphics[width=\linewidth]{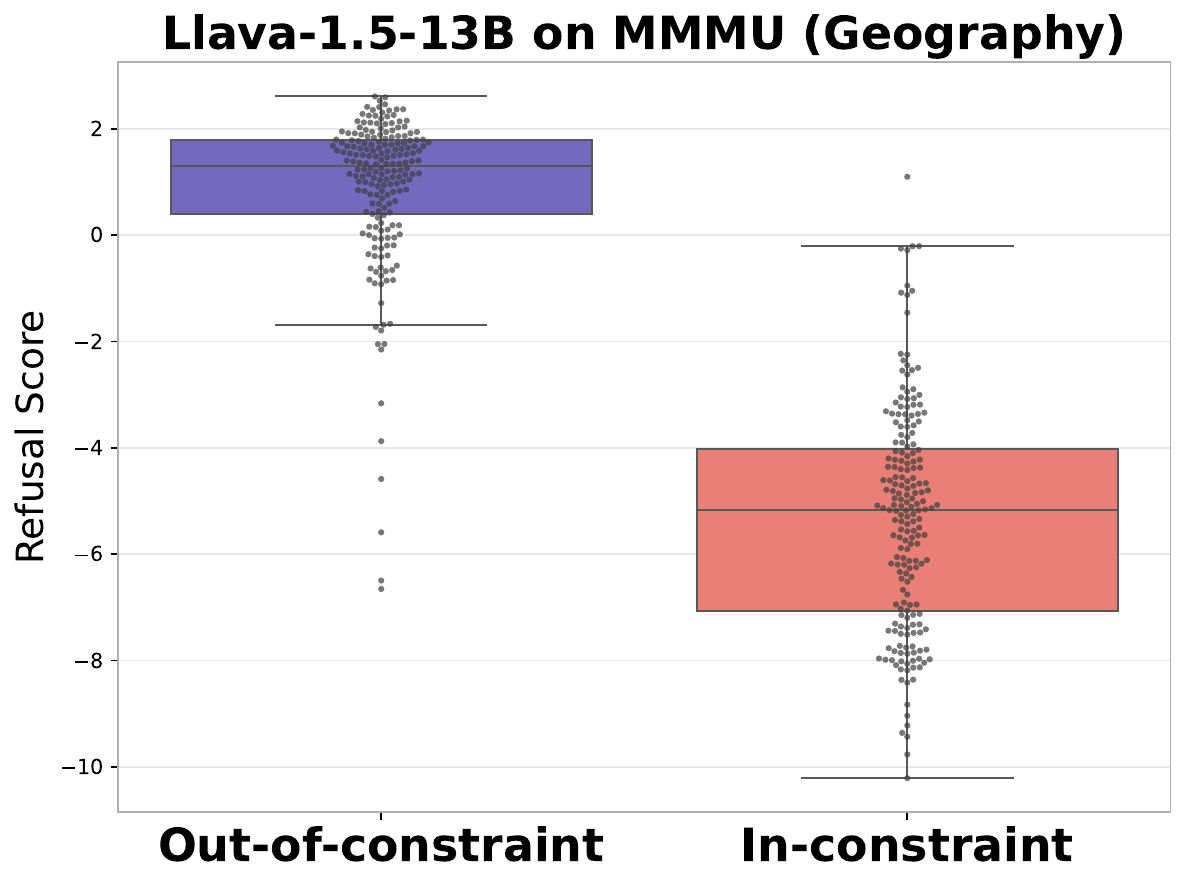}
        \label{fig:c}
    \end{subfigure}
    \hfill
     \begin{subfigure}{0.23\textwidth}
        \centering
        \includegraphics[width=\linewidth]{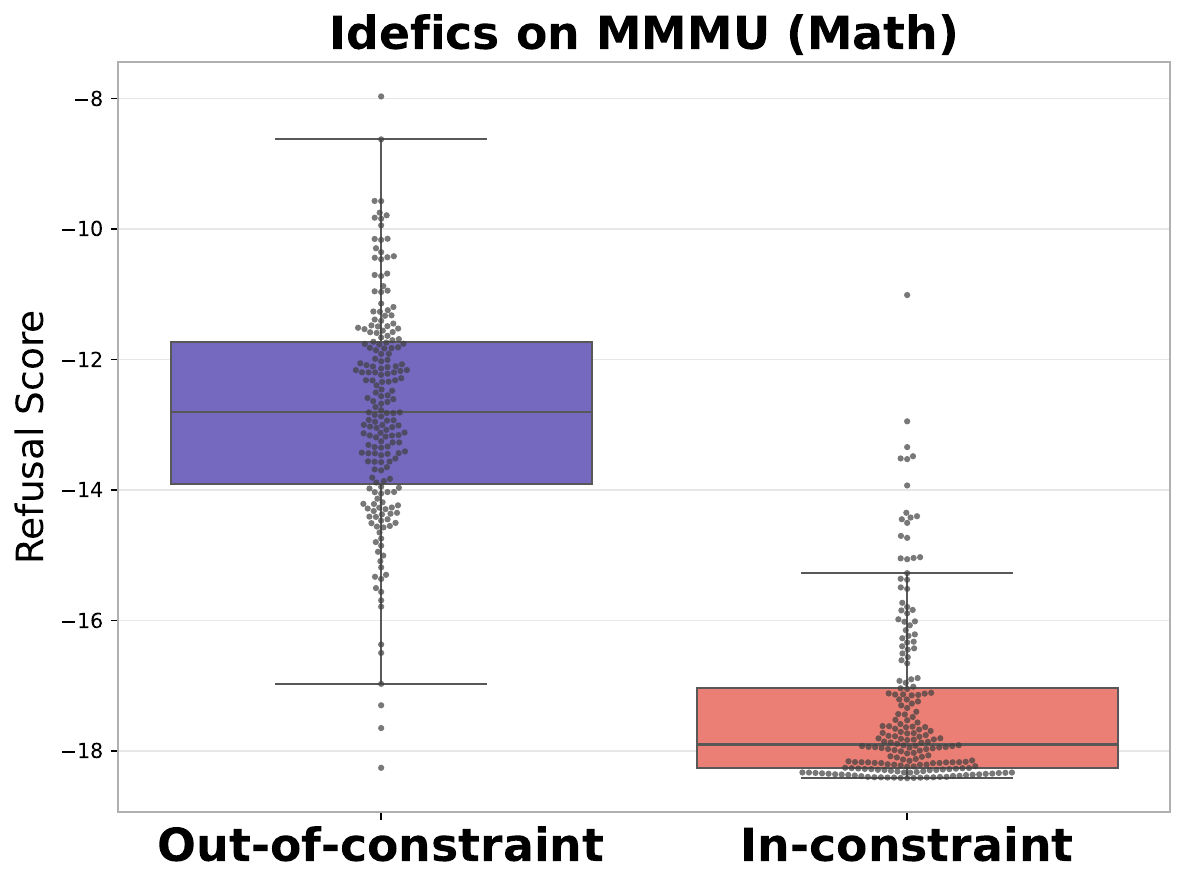}
        \label{fig:a}
    \end{subfigure}
    \hfill
    \begin{subfigure}{0.23\textwidth}
        \centering
        \includegraphics[width=\linewidth]{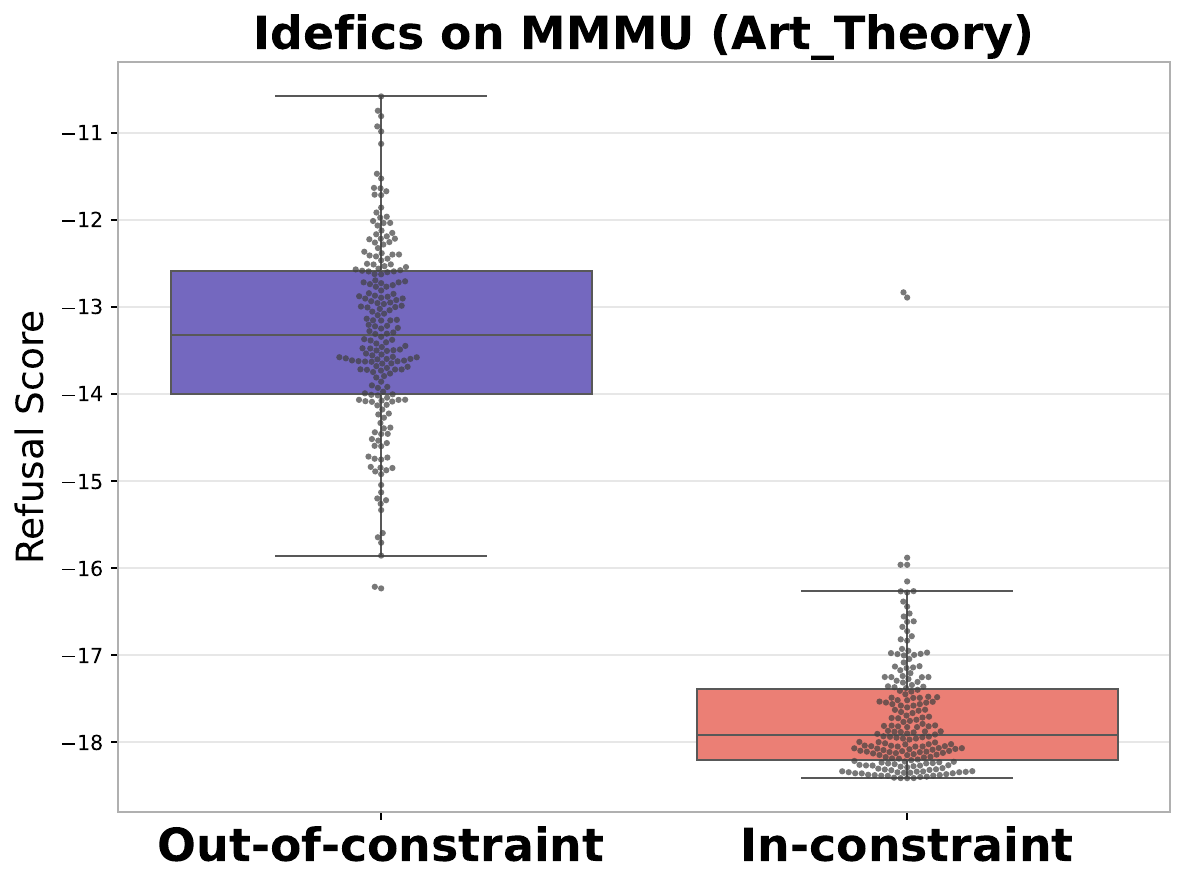}
        \label{fig:b}
    \end{subfigure}
    \hfill
    \begin{subfigure}{0.23\textwidth}
        \centering
        \includegraphics[width=\linewidth]{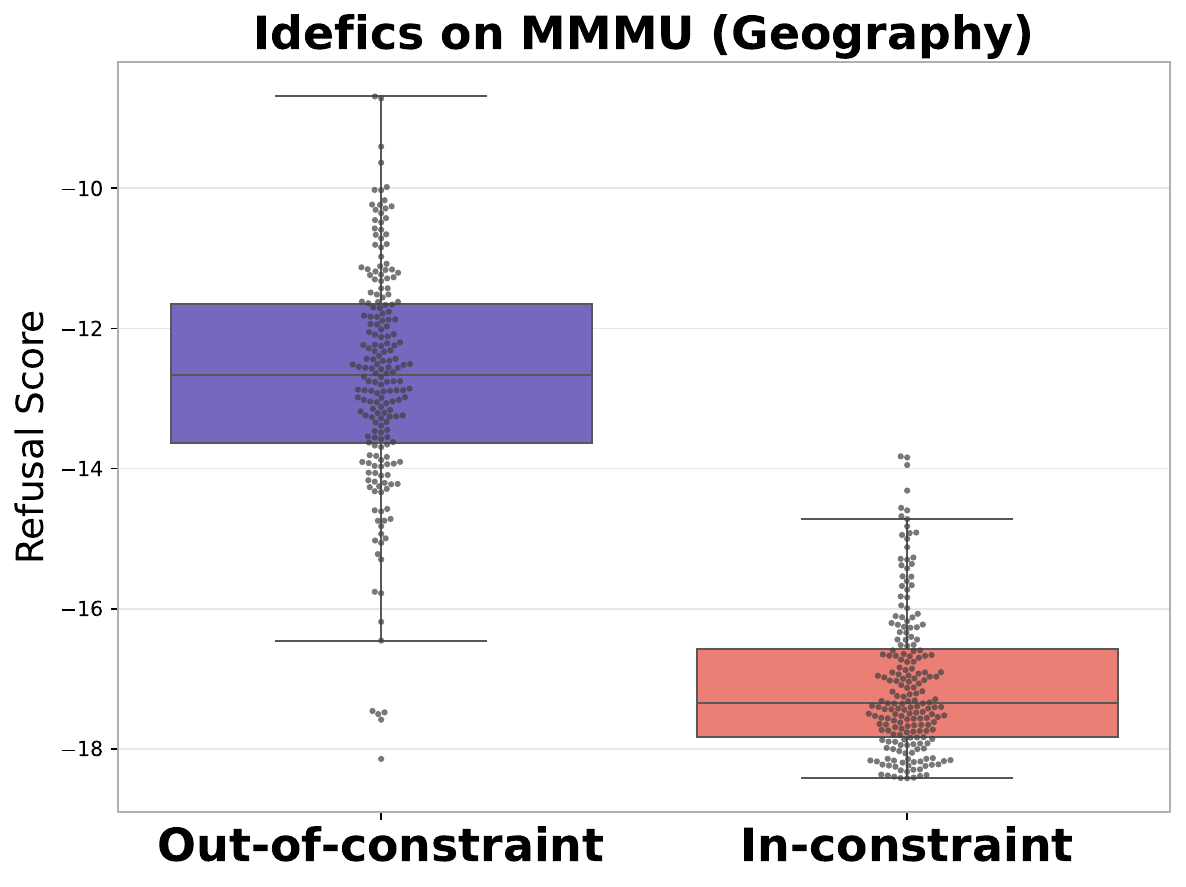}
        \label{fig:c}
    \end{subfigure}
    \hfill
    \begin{subfigure}{0.23\textwidth}
        \centering
        \includegraphics[width=\linewidth]{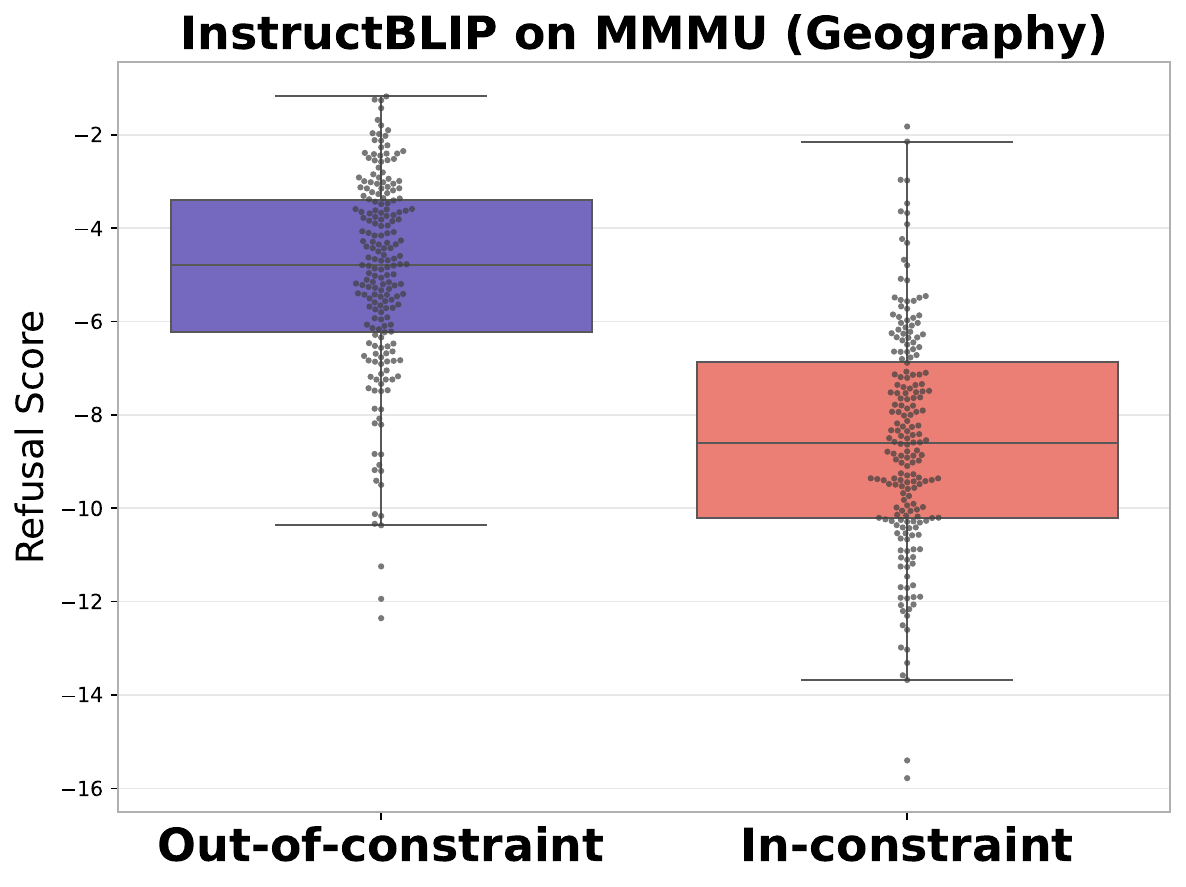}
        \label{fig:c}
    \end{subfigure}
    \hfill
    \begin{subfigure}{0.23\textwidth}
        \centering
        \includegraphics[width=\linewidth]{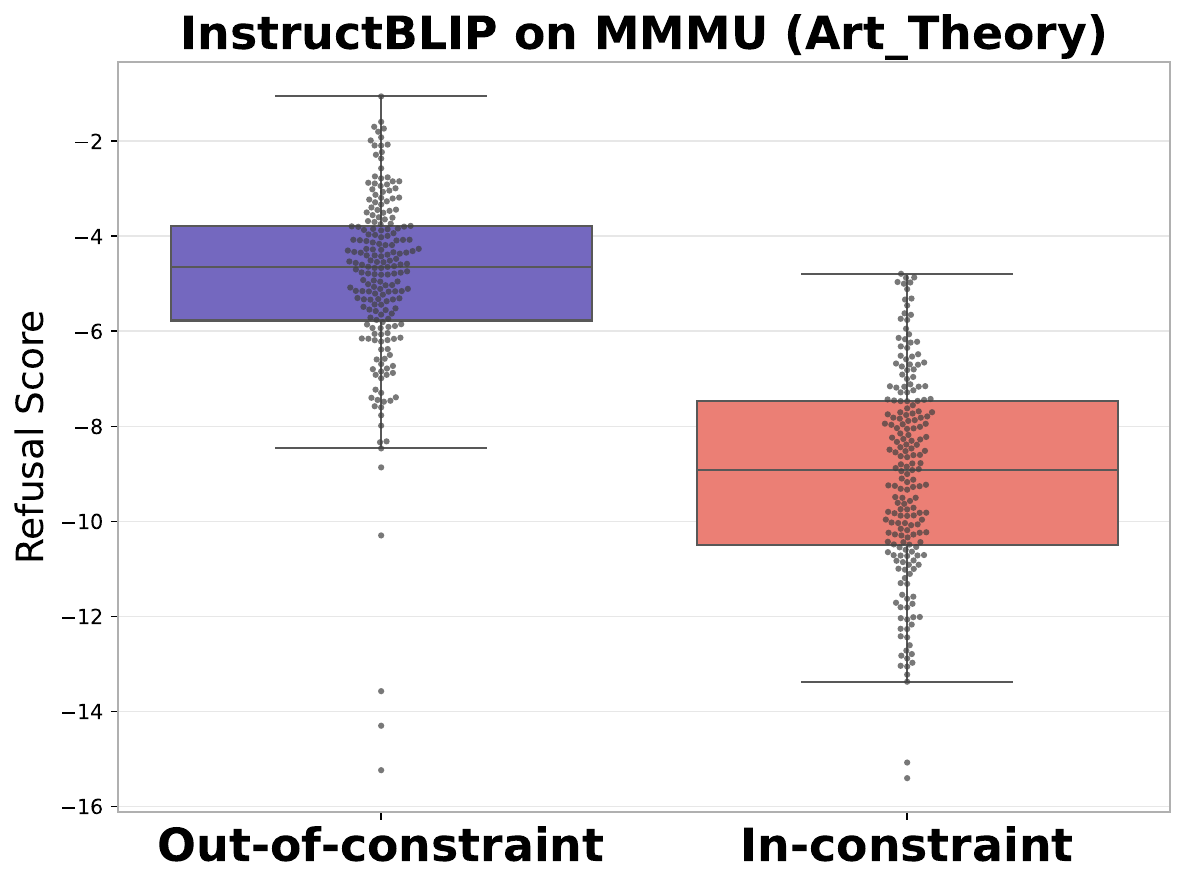}
        \label{fig:c}
    \end{subfigure}
    \hfill
    \begin{subfigure}{0.23\textwidth}
        \centering
        \includegraphics[width=\linewidth]{figures/RQ4/refusal_score_comparison_instructblip-vicuna-7b_MMMU_Geography.pdf}
        \label{fig:c}
    \end{subfigure}

    \caption{Refusal score shift for in-scope and out-of-scope test samples on MMMU.}
    \label{fig:RQ4_MMMU}
\end{figure*}

The results on ScienceQA are shown in Figure~\ref{fig:RQ4}, while the corresponding results on MMMU are presented in Figure~\ref{fig:RQ4_MMMU}. Across all evaluated VLMs and subject domains, out-of-scope test samples consistently exhibit substantially higher refusal scores than in-scope samples. Importantly, in-scope samples remain concentrated in low refusal-score ranges, indicating that our method selectively amplifies refusal behavior for out-of-scope inputs without introducing over-refusal. This consistent pattern across datasets further validates the effectiveness and robustness of CR-VLM.

% The results of ScienceQA are shown in Figure~\ref{fig:RQ4}, and the results for another dataset, MMMU, are presented in the appendix~\ref{appendix:refusal_score} as well.
% Across all VLMs, out-of-scope test samples consistently exhibit substantially higher refusal scores than in-scope samples. Importantly, in-scope samples remain concentrated at low refusal score ranges, confirming that our method selectively amplifies refusal behavior only for out-of-scope inputs without introducing over-refusal.
% Similarly, we also conduct experiments on MMMU dataset across different subjects and the results are reported in Figure~\ref{fig:RQ4_MMMU}.
% Similar to the results on ScienceQA, out-of-scope samples consistently show higher refusal scores than in-scope samples across all evaluated LVLMs. This further validates that our method selectively increases refusal tendency for out-of-scope inputs without introducing over-refusal.

%%%%%%%%%%%%%%%%%% Refusal Score Figures %%%%%%%%%%%%%%%%%%

% \section{Ablation Study}
% \subsection{Over-Refusal Mitigation by Orthogonality}
% \label{appendix:orth}
% We conduct an ablation study by comparing designing loss function with or without the orthogonality regularization term. The results report in Table~\ref{tab:orth} demonstrate that severe over-refusal on in-scope queries across all subjects without the orthogonality loss term. This indicates that without explicit decoupling, the learned intervention directions for in-scope and out-of-scope samples collapse into a shared global shift, leading to over-refusal. 
% In contrast, incorporating the orthogonality constraint effectively suppresses over-refusal while preserving strong refusal behavior for out-of-scope inputs, validating its critical role in enabling configurable refusal.

\section{Additional Ablation Study of Vision Loss}
\label{appendix:vision}
\begin{figure}[h]
    \centering
    \begin{subfigure}{\linewidth}
        \centering
        \includegraphics[width=0.45\linewidth]{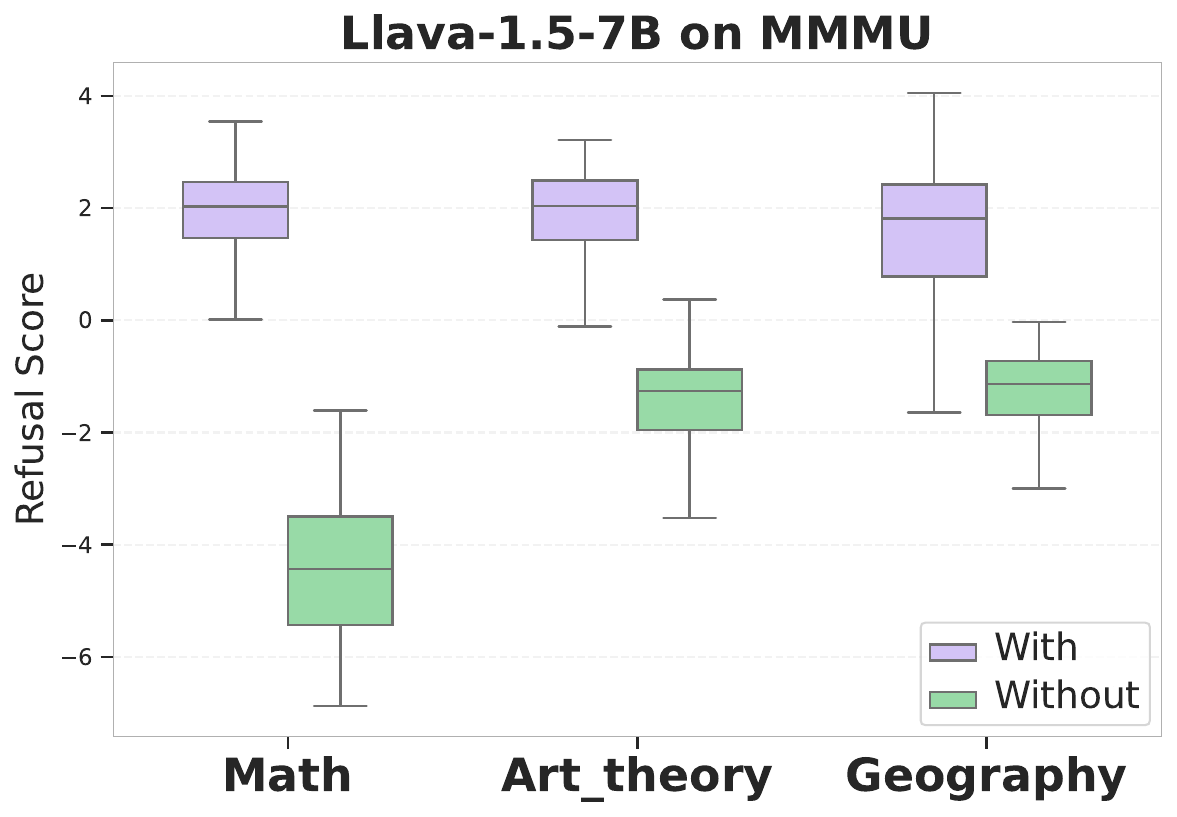}
    \end{subfigure}
    % \vspace{-20pt}
    \caption{(EQ3) Ablation Study of Vision Loss conducted on LLaVA-1.5-7B-hf on MMMU.}
    \label{appendix:mmmu_vision_ablation}
    % \vspace{-10pt}
\end{figure}

To further verify the effectiveness of the vision loss on visual grounding in refusal behavior, we conduct a similar ablation study on the MMMU dataset. Consistent with the main results reported on ScienceQA in the main content, we evaluate refusal behavior using the refusal score while providing the image input with blank textual content to isolate visual contributions by excluding the visual influence.
As shown in Figure~\ref{appendix:mmmu_vision_ablation}, removing the vision loss consistently leads to lower refusal scores across different MMMU subjects. These results suggest that vision loss promotes visual attention of the model when making a refusal behavior.

In addition, we provide a complementary analysis from a representation-level perspective beyond the results shown in section~\ref{sec:vision}. Our goal is to test whether the predicted refusal-related representations preserve visual evidence, rather than being only dominated by textual cues. Concretely, we compute the cosine similarity between the predicted representation produced from multimodal inputs (\ie both vision and text) and a vision-conditioned reference representation (\ie vision-only), and report the averaged similarity across different subjects. A higher similarity score indicates a stronger influence of visual cues on the predicted representations.
The results represented in Table~\ref{tab:vision} demonstrate that the vision loss acts as an effective regularizer, prompting greater reliance on visual cues rather than text-dominated decision patterns.

%%%%%%%%% Vision Table %%%%%%%%%%%%
\begin{table}[h]
    \centering
    \captionsetup{width=\linewidth}
    \caption{Hidden representation alignment with (w) and without (w/o) vision loss.}
    \resizebox{.6\textwidth}{!}{%
    \begin{tabular}{ccccccc}
    \toprule
        \multicolumn{1}{c}{\multirow{2}{*}{\textbf{Dataset}}} & \multicolumn{3}{c}{\textbf{ScienceQA}} & \multicolumn{3}{c}{\textbf{MMMU}} \\
        \cmidrule(lr){2-7}
        & \textbf{Biology} & \textbf{Physics} & \textbf{Geography} & \textbf{Math} & \textbf{Art Theory} & \textbf{Geography} \\
        \cmidrule(lr){1-7}
        w/o & -0.021 & 0.070 & 0.200 & 0.203 & 0.106 & 0.068 \\
        w & \underline{0.019}  & \underline{0.107} &  \underline{0.224}& \underline{0.252} & \underline{0.165} & \underline{0.082} \\
    \bottomrule
    \end{tabular}
    }
    \label{tab:vision}
\end{table}
\vspace{20pt}
% \begin{figure}[h]
% \centering
%     % \setlength{\tabcolsep}{10pt}
%     \scalebox{0.9}
%     {%
%         \begin{tcolorbox}[title=Example prompt used for \textbf{\textit{LLM-as-Judgment}}., label=LLM_as_Judgment_prompt]
%         \textbf{Principles:} Identify whether the response expresses refusal to answer the user’s question. Refusal includes direct rejection, statements of inability (\eg `I cannot see', `I don't have access'), or citing safety/policy restrictions. Regard responses starting with a rejection as refusal as well. If refusal semantics are present, return 1; otherwise return 0. Only output the number 0 or 1.
%         % \newline
%         % \textbf{Example 1:} \newline
%         % Response: I cannot answer. \newline
%         % Answer: 1 \newline
%         % \newline
%         % \textbf{Example 2:} \newline
%         % Response: A scarf is a solid object, typically made of fabric such as wool, silk, or cotton. \newline
%         % Answer: 0 \newline
%     \end{tcolorbox}
%     }% 
%     \caption{The prompts for generating \textbf{\textit{LLM-as-Judgment}} with principles and examples}
%     \label{fig:LLM_as_Judgment_prompt}
% \end{figure}
\begin{figure}[h]
\centering
\begin{tcolorbox}[
    title=Example prompt used for \textbf{\textit{LLM-as-Judgment}}.,
    label=LLM_as_Judgment_prompt,
    width=0.7\linewidth
]
\textbf{Principles:} Identify whether the response expresses refusal to answer the user’s question. Refusal includes direct rejection, statements of inability (\eg `I cannot see', `I don't have access'), or citing safety/policy restrictions. Regard responses starting with a rejection as refusal as well. If refusal semantics are present, return 1; otherwise return 0. Only output the number 0 or 1.
\end{tcolorbox}
\vspace{-10pt}
\caption{The prompts for generating \textbf{\textit{LLM-as-Judgment}} with principles and examples}
\label{fig:LLM_as_Judgment_prompt}
\end{figure}

\section{LLM-as-Judgment}\label{appendix:llm_as_judgment}
For the LLM-as-Judgment framework used to compute the refusal rate and over-refusal rate in this paper, we adopt a strong and popular model, DeepSeek-V3~\cite{liu2024deepseek}, as the evaluator model. The prompt template employed for this evaluation is provided in Figure~\ref{fig:LLM_as_Judgment_prompt}.

\section{Architecture of Calibrated Model}
The architecture of our calibration module for generating activation adjustments used in tunable refusal for VLMs is shown in Table~\ref{tab:arch_model}.
\begin{table}[h]
\centering
\caption{The architecture of the calibrated model.}
\begin{tabular}{cc}
\toprule
\textbf{Input} & Hidden state $x \in \mathbb{R}^{d}$ \\
\midrule
\multirow{3}{*}{\textbf{Flow Network}} & Linear($\text{Input}\_\text{dim}$, 1024) \\
& SiLU $\rightarrow$ Linear(1024, 1024) \\
& SiLU $\rightarrow$ Linear(1024, $\text{Output}\_\text{dim}$) \\
\midrule
\textbf{Gate Network} & Linear, Sigmoid \\
\midrule
\textbf{Residual Update} & $\Delta x$ \\
\midrule
\textbf{Output} & $x_{\text{out}} = x + \alpha\, p\, \Delta x$ \\
\midrule
\textbf{Trainable Parameter} & Scalar $\alpha$ \\
\bottomrule
\end{tabular}
\label{tab:arch_model}
\end{table}

\section{Additional experiment Settings}
We optimize the loss defined in Equation~\ref{eq:overall_loss}, where all loss weights are set to 1 across experiments. We use a batch size of 8, a learning rate of $1\times10^{-4}$, and train for 200 epochs with the Adam optimizer. Activation steering is applied to different layer ranges depending on the backbone: layers 12–32 for LLaVA-1.5-7B-hf, layers 12–40 for LLaVA-1.5-13B-hf, layers 18–32 for InstructBLIP-Vicuna-7B, and layers 20–32 for Idefics3-8B-Llama3. All hyperparameters are selected via preliminary validation and kept fixed across datasets and models. All experiments use a fixed random seed of 42 and identical decoding settings.

\section{Current Limitations and Future Directions}
While our framework demonstrates strong effectiveness in controlling personalized refusal behaviors, it is subject to several main limitations.
First, our method relies on open-source models with accessible hidden states to extract representation activations and compute steering vectors. This restricts its direct applicability to closed-source models (\eg GPT-4), where internal activations are not exposed.
% Second, CR-VLM requires training a lightweight calibration module for each target model, which introduces a one-time computational overhead. However, this overhead is significantly lower than fine-tuning and remains practical when scaling to multiple model architectures \yuchen{Show evidence here, avoid just mentioning `significantly', 'largely' without the number in the entire paper, show the training time difference, tuned model parameter size. I suggest adding a table in the appendix, and refer to that table anytime you mention the efficiency.}
% Third, the current experimental setup considers only one constraint per setting, and evaluating configurable refusal under multiple concurrent constraints is left for future investigation.
Second, our current experiments focus on one constraint at a time to keep the analysis clear and controlled. This is a design choice instead of a limitation of the framework. In practice, multiple constraints can be incorporated together through composition, and a systematic study of such multi-constraint settings is left for future work.
Third, our current evaluation focuses on domain-level configurable refusal, as standardized benchmarks for explicit policy-conditioned scenarios (\eg regulatory constraints) are still lacking. 
% Extending our framework to such settings can be considered in future work once suitable benchmarks become available. However, this does not mean that CR-VLM cannot easily adapt to these benchmarks.
This is mainly a benchmarking constraint rather than a limitation of the framework. Extending CR-VLM to such settings is straightforward, and we plan to explore this direction once suitable benchmarks become available.

\iffalse
\section{Reproducibility}
We release an anonymous code repository containing the implementation of CR-VLM for reviewers.
The anonymous code link is available at: \url{https://anonymous.4open.science/r/Configurable_Refusal_VLM-2E48}.
\fi

%%%%%%%%%%%%%%%%%%%%%%%%%%%%%%%%%%%%%%%%%%%%%%%%%%%%%%%%%%%%%%%%%%%%%%%%%%%%%%%
%%%%%%%%%%%%%%%%%%%%%%%%%%%%%%%%%%%%%%%%%%%%%%%%%%%%%%%%%%%%%%%%%%%%%%%%%%%%%%%

\end{document}